\documentclass{article}
\usepackage[preprint]{neurips_2026}

\usepackage[utf8]{inputenc} 
\usepackage[T1]{fontenc}    
\usepackage{hyperref}       
\usepackage{url}            
\usepackage{booktabs}       
\usepackage{amsfonts}       
\usepackage{nicefrac}       
\usepackage{microtype}      
\usepackage{xcolor}         
\usepackage{graphicx} 
\usepackage{longtable}
\usepackage{multirow}
\usepackage[table]{xcolor}
\usepackage{amsmath}
\usepackage{float}
\usepackage{arydshln}
\usepackage{wrapfig}

\usepackage{titletoc}
\title{BrainNorm: A Foundation Model that knows Normal via Semantic Atlas Pretraining}

\author{%
  Madhumitha Venkatesh \\
  Indian Institute of Technology Hyderabad \\
  \texttt{ai23resch11004@iith.ac.in} \\
  \And
  Shanawaj S Madarkar \\
  Indian Institute of Technology Hyderabad \\
  \texttt{ai25resch14001@iith.ac.in} \\
  \AND
  Konda Reddy Mopuri \\
  Indian Institute of Technology Hyderabad \\
  \texttt{krmopuri@ai.iith.ac.in}
}

\begin{document}

\maketitle

\begin{abstract}
  We introduce \textbf{BrainNorm}, a normative foundation model, trained and tested on $\sim$66,000 T1-weighted structural MRI (T1w sMRI) scans. By leveraging language-image style contrastive pretraining on healthy cohorts across ages, BrainNorm learns a \emph{Semantic Atlas Latent} space (SAL), where each scan is represented as a set of atlas-parcel embeddings. This yields parcel-specific healthy aging template trajectories that support age-consistent template matching and \emph{localized} deviation scoring relative to a subject’s chronological age.
  Across 6 downstream cohorts, BrainNorm demonstrates generalization evaluated across 25 task-setting combinations spanning age estimation, brain-age gap estimation, parcel identification, and single- \& multi-disease classification tasks under direct inference, zero-shot, few-shot \& full-data linear-probe settings.
  The resulting deviation patterns in SAL space enable zero-shot tasks for disease prediction using parcel-wise abnormalities. Fine-tuning on healthy-only cohorts of downstream datasets further improves the performance of various tasks. Across all classification tasks, linear probing on BrainNorm’s frozen embeddings outperforms 9 baselines finetuned under end-to-end supervision. Furthermore, the localized deviations identified by BrainNorm across various neurodegenerative disorders closely align with established neurodegeneration pathology in clinical literature.
\end{abstract}

\section{Introduction}
\label{sec:intro}

\noindent\textit{The pathology of abnormality can be better understood in relation to the normal, 
and so one must first map the boundaries of normal to discover what lies outside.}
Normative modeling makes this principle operational for neuroimaging. Given a large healthy reference cohort, it learns the distribution of typical brain structure and asks a clinically sharper question: \emph{where does this individual fall relative to what is expected?} This framework has long been used in medicine to model ``normal'' variation, most familiarly through pediatric growth charts and centile curves~\cite{Bethlehem2022BrainChartsNature,Schabdach2023BrainGrowthCharts,Cole2012GrowthCharts,Marquand2016NormativeBeyondCaseControl}. Its relevance is particularly strong in neurodegeneration, where large-scale healthy cohorts are more readily available than well-annotated disease cohorts. Many neurological diseases begin years, sometimes decades, before clinical symptoms appear, emerging as subtle and spatially localized departures from otherwise normal aging trajectories~\cite{Young2018,Bethlehem2022BrainChartsNature}. Constructing a reliable normative reference is therefore fundamentally linked to the problem of detecting neurodegenerative anomalies early and at the individual level. Classical approaches relied on probabilistic regression over morphometric features (e.g., Gaussian process and Bayesian normative models), later extended to large multi-site cohorts and lifespan-scale brain charts \cite{Wolfers2018JAMAPsychNormativeSZBD,Wolfers2020PsychMedADHDNormative,Fraza2021WarpedBLR,Rutherford2022NatProtocolsNormativeFramework,Kia2022FederatedHBRLifeCycle,Bethlehem2022BrainChartsNature,Ge2024CentileBrainLancetDH}. However, they remain constrained by hand-crafted regional summaries and limited representation capacity.


Existing approaches to normative inference largely fall into two paradigms, each with structural limitations. The first is \emph{brain-age prediction}, where a model is trained on healthy subjects to estimate chronological age from brain morphology, and the brain-age gap (BAG), difference between predicted and actual age, is interpreted as an anomaly score~\cite{Cole2017DeepLearningBrainAge,Peng2021SFCN,Lee2022NatureAgingBrainAge}. While effective as a global biomarker, this formulation collapses complex spatial variation into a single scalar. The BAG can indicate that a brain appears atypical for its age, but provides little insight into \emph{where} deviations occur or \emph{how} they manifest across regions. Distinct diseases with different regional atrophy patterns can produce the same brain-age gap, making the score non-specific and insensitive to spatial heterogeneity.

The second paradigm employs generative normative models, such as Variational Autoencoders (VAEs)~\cite{Pinaya2019DeepAutoencodersHBM,Kumar2021NormVAE,LawryAguila2022cVAENormativeMICCAI,Pinaya2022TransformersMedIA,Zimmerer2019VAEAnomalyMICCAI} and Diffusion architectures~\cite{Pinaya2022DiffusionMICCAI,Frotscher2025ANDi,puglisi},
to learn latent distributions of healthy anatomy. These models identify anomalies by detecting where voxel-space reconstruction fails. While intuitive, this approach is computationally expensive and "brittle". The resulting residual signal often conflates genuine pathology with noise variability from scanner effects, protocols, or registration errors. Similarly, \emph{conditional/ deformable template matching} techniques~\cite{Lorenzi2019,Fu2025LearnedAgingTemplates,Dalca2019ConditionalDeformableTemplates,Sivera2019MorphChangesAgingAD,Wang2025MorphLDM} replace generative reconstruction with age-conditioned anatomical templates, but the underlying operation remains unchanged, i.e., voxel-wise deformation estimation in image space.  Consequently, both approaches struggle to disentangle true structural deviation from imaging artifacts, often requiring additional downstream classifiers to filter out inherited noise.

\begin{figure*}[t]
  \centering
  \includegraphics[width=\linewidth]{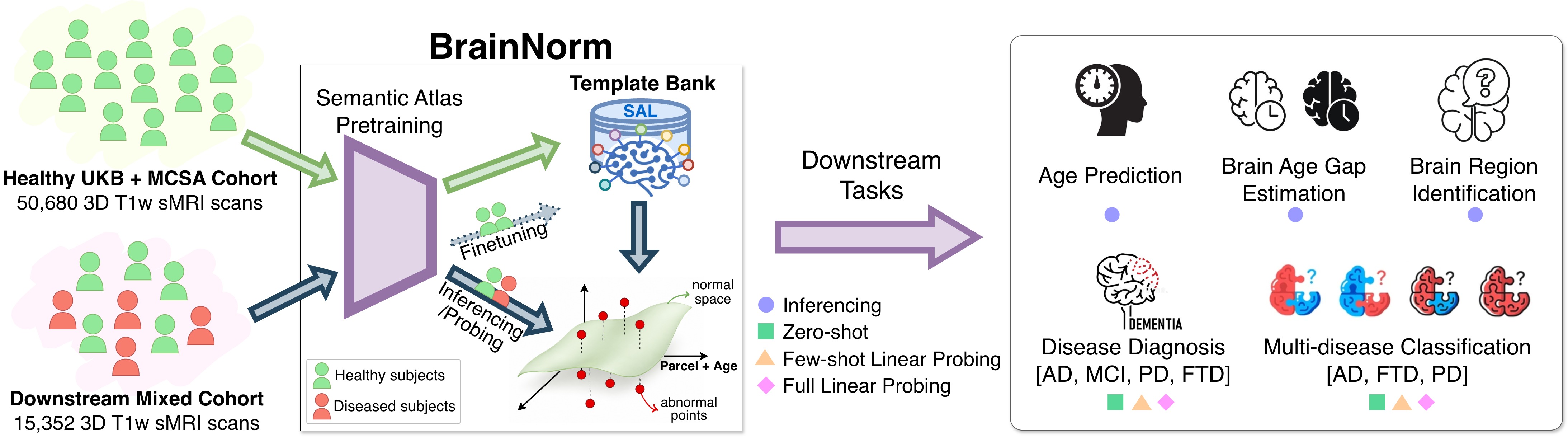}
  \vspace{-1em}
  \caption{BrainNorm Framework: A representation learning technique that models healthy aging as anatomically grounded representations exclusively from healthy subjects. By utilizing frozen embeddings within a Semantic Atlas Latent Space, BrainNorm enables zero-shot, few-shot, and full linear probing for downstream prediction and classification tasks on mixed clinical cohorts. }
  \label{fig:overview}
  \vspace{-0.9em}
\end{figure*}

Taken together, these observations point to a common bottleneck: the challenge lies not merely in detecting deviations, but in the space where the comparison is performed. When normative inference relies on voxel-space residuals, whether from reconstruction or template matching, the resulting anomaly signal becomes entangled with imaging noise that is difficult to isolate genuine atrophy. Normative modeling does not require reproducing voxels; it requires evaluating deviations on representations that are stable across acquisition conditions. Recent advances in representation learning have moved towards prediction in representation space rather than pixel reconstruction~\cite{IJEPA2023,VJEPA2024}. This motivates reframing normative inference from image space to an \emph{atlas-aligned latent space}, where regions are explicitly represented, and deviations can be evaluated per region.

We introduce \textbf{BrainNorm}, a normative foundation model, which addresses these limitations through representation learning via \textit{Semantic Atlas Pretraining} (Fig~\ref{fig:overview}). To the best of our knowledge, this is the first normative foundation model trained and tested on a large-scale cohort of $66,032$ 3D T1w sMRI scans from $49{,}579$ subjects. Departing from traditional voxel-level objectives, our framework learns a \emph{Semantic Atlas Latent Space (SAL)}, an anatomically factorized normative latent space, that is explicitly organized at the parcel level and ordered by age.
Within this space, we evaluate: \textit{how similar is a scan to healthy brains of the same age, and which anatomical parcels deviate from this expectation?}
Without supervision, our learned SAL space directly supports: (i) parcel-wise normative matching against age-consistent healthy templates, (ii) retrieval-based subject age and brain age-gap estimations with a precision of 0.1 year, and (iii) zero-shot normative reference that localizes neurodegenerative anomalies/ deviations across anatomical regions and ages, enabling binary and multi-disease classifications, and parcel-specific interpretability. Further, we demonstrate that BrainNorm achieves superior performance with simple linear probing on pretrained/ healthy-only finetuned representations, compared to end-to-end finetuned baselines.

\section{Related Works} 
\noindent\textbf{Age prediction:}
The idea that brain structure encodes biological aging predates deep learning: BrainAGE \cite{Franke2010KernelAgeNeuroImage} trained kernel models on morphometric features to predict chronological age, using the Brain-Age Gap as a deviation index \cite{Cole2018BrainAgeMortality}. Deep learning then shifted the field toward supervised age regression directly from 3D T1w volumes, first with 3D CNNs \cite{Cole2017DeepLearningBrainAge,Peng2021SFCN,Lee2022NatureAgingBrainAge} and more recently with 3D ViTs \cite{Zhang2025CommsMedicineBAG}. However, optimizing strictly for age prediction can reduce disease sensitivity, revealing a misalignment between age accuracy and abnormality detection \cite{Schulz2025PLOSBioBrainAgeDisease}, motivating normative approaches that condition on age while prioritizing deviation.

\noindent
\textbf{Contrastive Pretraining:}
Contrastive self-supervision has proved effective for learning representations without explicit labels, motivating its adoption in medical imaging~\cite{ouyang2022lne_media,ouyang2022disentangle,zhao2021lssl}. In neuroimaging, recent works such as BrainIAC~\cite{tak2026brainiac}, y-Aware InfoNCE~\cite{dufumier2021yaware}, and AnatCL~\cite{barbano2026anatomical} demonstrate that contrastive self-supervised learning on brain MRI can capture semantically meaningful representations. In parallel, medical vision-language pretraining aligns images with reports or captions (e.g., MedCLIP~\cite{wang-etal-2022-medclip}, MedSigLIP~\cite{sellergren2025medgemma}, 
BioViL-T~\cite{Bannur_2023_CVPR}, 
BiomedCLIP~\cite{zhang2023biomedclip}), enabling retrieval and zero-shot transfer but typically relying on image–report supervision. Finally, CLIP-style objectives have evolved for better scalability. Among them, SigLIP~\cite{Zhai2023SigLIP} replaces softmax-normalized InfoNCE with a sigmoid-based pairwise loss that scores pairs independently and reduces large-batch dependence.
We adopt SigLIP in our framework, which naturally supports multiple positives for a given anchor while retaining the core benefit of contrastive alignment in a shared semantic space.

\begin{figure*}[t]
  \centering
  \includegraphics[width=\linewidth]{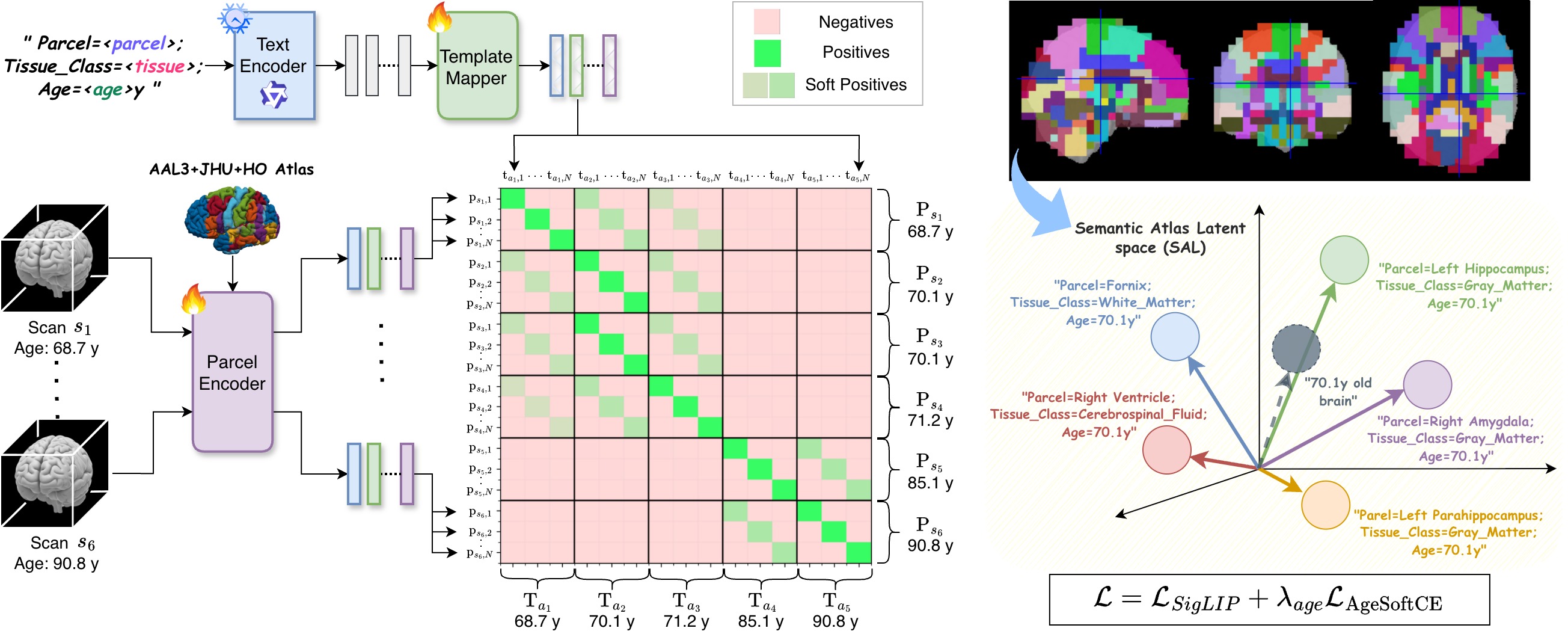}
  \vspace{-1.5em}
  \caption{\textbf{BrainNorm pretraining framework} for a sample batch of scans. BrainNorm encodes a scan as a set of parcel embeddings in a shared semantic atlas latent (SAL) space, rather than compressing it into a single global representation (shown as a single gray vector). These embeddings are aligned to an age-indexed, parcel-specific healthy template bank initialized from structured textual priors and refined through a learnable Template Mapper.}
  \label{fig:framework}
\end{figure*}

\section{BrainNorm}
\label{sec:framework}

BrainNorm learns a \emph{semantic atlas latent space} from healthy cohorts via contrastive pretraining, which is explicitly structured at the anatomical parcel level and ordered by age. 
The central objective of BrainNorm is to construct an age-indexed normative template bank:
$
\{\mathbf{t}_{a,p}\}_{a=A_{min},\dots,A_{max};\;p=1,\dots,N},
$
where each parcel $p$ is associated with a trajectory of embeddings across ages $a$.
These trajectories represent the expected healthy evolution of parcel-specific morphology. BrainNorm enables:
(i) Age-consistent matching by alignment with the appropriate age-conditioned template, and (ii) localized deviation scoring by quantification of parcel-wise abnormality relative to normative aging trajectories.
By explicitly modeling healthy parcel trajectories, BrainNorm transforms normative modeling into a structured representation learning problem: instead of predicting age or disease labels directly, the model learns the geometry of healthy anatomical variation itself.


\subsection{Overview}
\label{sec:motivation}

BrainNorm (Fig~\ref{fig:framework}) operates at parcel-level representations, where each scan $i$ of age $a_i$, is represented by a set of anatomical parcel embeddings $\{\mathbf{p}_{i,p}\}_{p=1}^{N}$ obtained using a dedicated transformer-based Parcel Encoder, adopted from P2P~\cite{madhumitha2026p2p}, 
rather than a single global embedding. The Parcel Encoder uses the atlas to identify the tokens that contain any information about a particular parcel to form a token-set defining the parcel’s context. Each set is processed independently by a shared Transformer encoder followed by generalized mean pooling (GeM, $r=2$) of the self-attended parcel tokens to produce a parcel-specific representation. 
Given these parcel embeddings, the central question becomes: \emph{what geometric structure should a normative latent space exhibit?} BrainNorm's normative reasoning imposes two fundamental constraints in the SAL space:

\noindent
\textbf{(i) Parcel specificity:}
For a parcel embedding $\mathbf{p}_{i,p}$, alignment must be strongest with templates of the same anatomical parcel ($p' = p$), while remaining well separated from other parcels. This ensures anatomical identifiability.

\noindent
\textbf{(ii) Age continuity:}
Healthy aging is continuous. Within a parcel, representations should vary smoothly with age, tolerating small age differences while progressively diverging as the age gap increases. Templates should therefore form ordered trajectories rather than isolated clusters.

To seed this structure, we introduce textual representations as \textit{priors} over the template bank. For each parcel and age, we construct structured texts encoding age and anatomical identity from the brain atlas. We employ Qwen3~\cite{qwen3technicalreport2025} embeddings for their strongly structured latent geometry, which naturally clusters by anatomical semantics while encoding ordinal age signals.

However, text geometry alone does not guarantee consistency with imaging-derived biological aging trajectories, since brain aging is region-specific, i.e., different parcels follow distinct structural trajectories across the lifespan~\cite{susan2003aging,Marjanska2017RegionSpecificAging,Lee2019DistinctBrainRegions}. We therefore introduce a lightweight learnable \emph{template mapper} that ``bends'' the text priors into an age-indexed template bank aligned with parcel embeddings. This provides a principled inductive bias: initialization from a semantically organized language space, followed by alignment using large-scale imaging evidence.

The resulting SAL space exhibits two properties: (i) parcel embeddings cluster by anatomy, and (ii) within each parcel, representations follow smooth age-ordered trajectories. To enforce this geometry, we adopt an age-aware pairwise alignment objective inspired by SigLIP~\cite{Zhai2023SigLIP}. Cross-parcel pairs serve as negatives to preserve anatomical separation, while within-parcel nearby-age neighborhoods form \emph{soft} positives, in which alignment strength decays smoothly with age difference, encouraging continuous progression across age.

\subsection{Semantic Atlas Pretraining}
\subsubsection{Setup and representations}
\label{sec:framework_overview}

\noindent
\textit{Parcel embeddings from 3D scans:}
For each subject $i$ and anatomical parcel $p\in\{1,\dots,N\}$, a Parcel Encoder produces a representation
$\mathbf{z}_{i,p}\in\mathbb{R}^{d'}$, which is projected into a shared SAL space of dimension $D$ and $\ell_2$-normalized:
\begin{equation}
\mathbf{p}_{i,p} = \frac{\mathrm{Proj}(\mathbf{z}_{i,p})}{\|\mathrm{Proj}(\mathbf{z}_{i,p})\|_2}
\in\mathbb{R}^{D},
\qquad
\|\mathbf{p}_{i,p}\|_2=1
\label{eq:image_embed}
\end{equation}

The projection layer $\mathrm{Proj}(\cdot)$ enables alignment between image-derived parcel embeddings and text-derived templates within a common semantic space.

\textit{Text templates as structured priors:}
We discretize age as $a \in \{A_{min},\dots,A_{max}\}$ (with a minimum difference of 0.1 year). For each age $a$ and parcel $p$, we construct age-conditioned text of the form:
\[
\text{\emph{``Parcel=<name>; Tissue\_class=<tissue>; Age=<}$a$\emph{>y''}}
\]
\noindent
These texts are embedded using a frozen Qwen3 text encoder to obtain
$\mathbf{e}^{\mathrm{raw}}_{a,p}\in\mathbb{R}^{d''}$.
A lightweight Template Mapper of three-layer MLP with GeLU non-linearity (denoted $\mathrm{Map}(\cdot)$ and detailed in Appendix~\ref{Asec:template_mapper}), projects these prior embeddings into the shared SAL space:
\begin{equation}
\tilde{\mathbf{t}}_{a,p}
=
\mathrm{Map}(\mathbf{e}^{\mathrm{raw}}_{a,p})
\in \mathbb{R}^{D},
\qquad
\mathbf{t}_{a,p}
=
\frac{\tilde{\mathbf{t}}_{a,p}}
{\|\tilde{\mathbf{t}}_{a,p}\|_2}
\in \mathbb{R}^{D}
\label{eq:template_bank}
\end{equation}

The resulting set $\{\mathbf{t}_{a,p}\}_{a,p=1}^{A,N}$ forms an age-indexed healthy template bank in the SAL space.
Here, $\mathbf{p}_{i,p}$ represents how parcel $p$ \emph{appears} in scan $i$, while
$\mathbf{t}_{a,p}$ represents how parcel $p$ is \emph{expected to appear in a healthy brain} at age $a$.
This enforces each parcel to own its trajectory across ages, rather than sharing a global prototype.

\subsubsection{Training objective}
\label{sec:objective}

\noindent
Training enforces the two structural biases introduced earlier: 
(i) cross-parcel separation (parcel specificity) and
(ii) smooth age ordering within each parcel (age continuity). We implement these using a pairwise SigLIP alignment objective complemented by a within-parcel age-shaping regularizer.

\paragraph{\textbf{Age-aware SigLIP alignment:}}
\label{sec:framework_siglip}



Let $\text{age}_i \in \mathbb{R}$ denote the chronological age of subject $i$.
For each observed parcel embedding $\mathbf{p}_{i,p}$ and candidate parcel template $\mathbf{t}_{a,p'}$, we define a soft target
$y_{i,p}^{a,p'} \in [0,1]$:

\begin{equation}
y_{i,p}^{a,p'}
=
\mathbb{I}[p'=p]\;
\mathbb{I}\!\left(|\text{age}_i - a| \le w \right)\;
\exp\!\left(
-\frac{(\text{age}_i - a)^2}{2\sigma^2}
\right)
\label{eq:soft_targets}
\end{equation}

where $w$ is an age window (in years) restricting positive candidates,
and $\sigma$ controls how rapidly the positive weight decays with age mismatch.
All cross-parcel pairs ($p' \neq p$) act as negatives by construction, enforcing parcel specificity.

\noindent
\textit{Scaled SigLIP logits:}
With a learnable logit scale $\alpha$ and bias $\beta$, compatibility between an observed parcel $\mathbf{p}_{i,p}$
and template $\mathbf{t}_{a,p'}$ is computed using a scaled cosine logit:
\begin{equation}
s_{i,p}^{a,p'}
=
\alpha\,
\mathbf{p}_{i,p}^{\top}
\mathbf{t}_{a,p'}
+
\beta.
\label{eq:compat_logit}
\end{equation}

Since both $\mathbf{p}_{i,p}$ and $\mathbf{t}_{a,p'}$ are unit-normalized, $s_{i,p}^{a,p'}$ supports efficient retrieval over the template bank.

\noindent
\textit{SigLIP loss:}
Using binary cross-entropy on logits, the age-aware SigLIP objective becomes:

\begin{align}
\ell_{\mathrm{bce}}(s,y)
=
-y \log \sigma(s)
-
(1-y)\log(1-\sigma(s)),\\
\mathcal{L}_{\mathrm{SigLIP}}
=
\frac{1}{N_b}
\sum_{(i,p)\in \mathcal{B}}
\frac{1}{A_{\mathcal{B}}N}
\sum_{a=1}^{A_{\mathcal{B}}}
\sum_{p'=1}^{N}
\ell_{\mathrm{bce}}
\!\left(
s_{i,p}^{a,p'},
y_{i,p}^{a,p'}
\right)
\label{eq:siglip_loss}
\end{align}

where $\mathcal{B}$ denotes the set of parcel instances in the batch, ${A_{\mathcal{B}}}$ is the number of unique scan ages in the batch, and $N_b = |\mathcal{B}|$.
This enforces two behaviors simultaneously:
(i) \emph{parcel discrimination}
and (ii) \emph{age separation}, by treating all cross-parcels pairs ($p'\neq p$) and farther ages of within parcel ($p'=p$) as negatives.

\paragraph{\textbf{Within-parcel age shaping via soft cross-entropy.}}
\label{sec:framework_agesoftce}

While SigLIP provides local pairwise supervision, it does not explicitly enforce a smooth unimodal age profile within each parcel. To encourage such structure, we introduce a within-parcel soft age classification regularizer.

Let $\mathcal{A}=\{A_{\min}, A_{\min}+0.1, \dots, A_{\max}\}$ denote the set of discretized ages, with $A=|\mathcal{A}|$. For each sample $i$ and parcel $p$, we form the within-parcel age-logit vector
$
\mathbf{s}^{\mathrm{age}}_{i,p}
=
\big[s_{i,p}^{a,p}\big]_{a\in\mathcal{A}}
\in\mathbb{R}^{A}$.
We define a soft target distribution over ages as
\begin{equation}
q_{i,p}(a)
=
\frac{
\exp\!\left(-\frac{(\mathrm{age}_i-a)^2}{2\sigma_p^2}\right)
}{
\sum\limits_{a'\in\mathcal{A}}
\exp\!\left(-\frac{(\mathrm{age}_i-a')^2}{2\sigma_p^2}\right)
},
\qquad a\in\mathcal{A},
\label{eq:age_soft_target}
\end{equation}
where $\sigma_p=\exp(\theta_p)$ is a learnable parcel-specific standard deviation.
The predicted within-parcel age distribution is:
$
\hat{q}_{i,p}(a)
=
\frac{
\exp\!\left(s_{i,p}^{a,p}\right)
}{
\sum\limits_{a'\in\mathcal{A}}
\exp\!\left(s_{i,p}^{a',p}\right)
},$ where $a\in\mathcal{A}$.
The resulting soft age cross-entropy loss over a batch $\mathcal{B}$ is:
\begin{equation}
\mathcal{L}_{\mathrm{AgeSoftCE}}
=
-\frac{1}{N_b}
\sum_{(i,p)\in\mathcal{B}}
\sum_{a\in\mathcal{A}}
q_{i,p}(a)\log \hat{q}_{i,p}(a).
\label{eq:soft_age_ce}
\end{equation}

\begin{wrapfigure}{r}{0.55\textwidth}
  \centering
\includegraphics[width=0.55\textwidth]{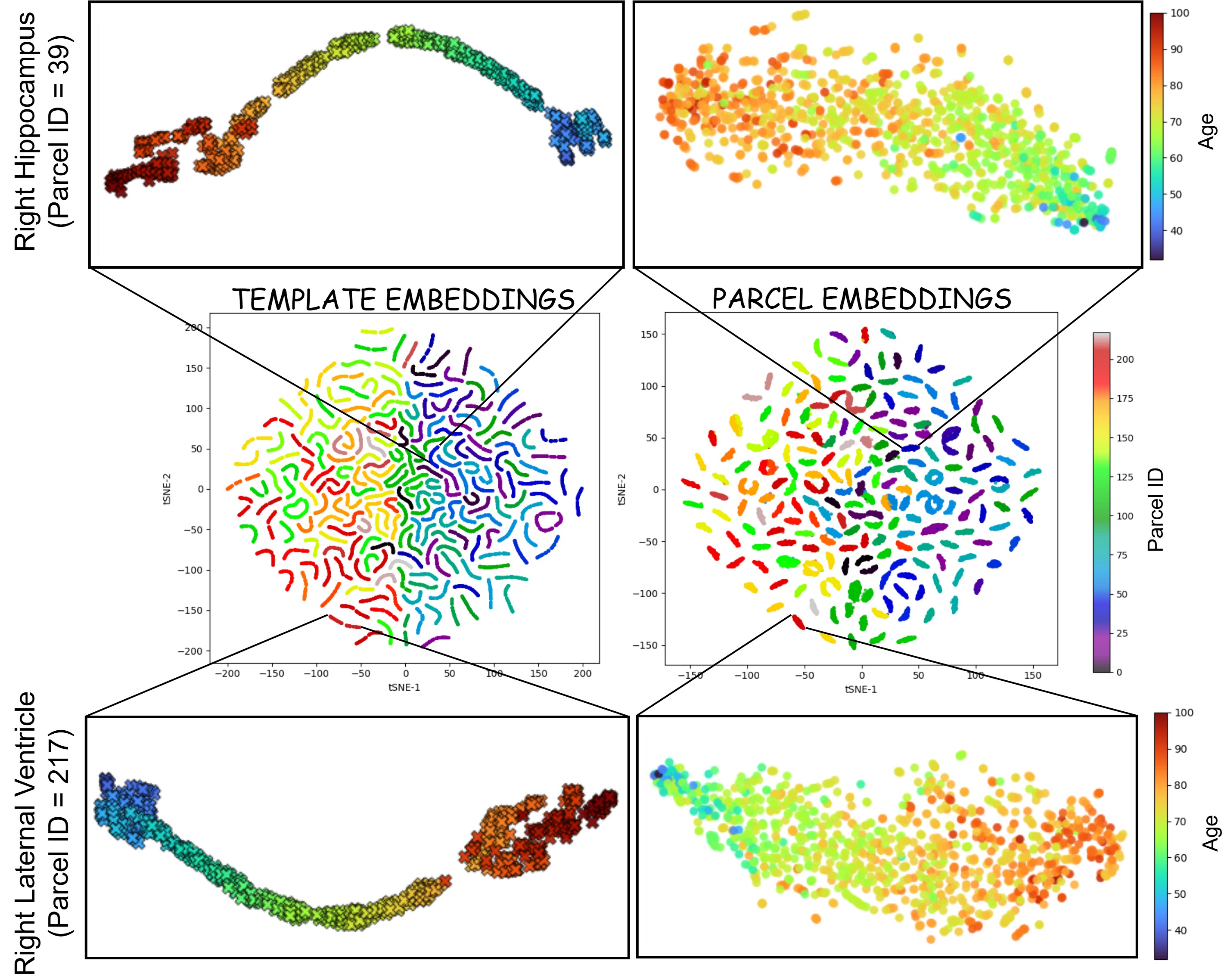}
  \vspace{-1.5em}
  \caption{\textbf{BrainNorm's template and parcel embeddings of healthy samples in the SAL space} are clustered by anatomical parcel and exhibit age-ordered organization (using t-SNE plots). The visualization zooms into two representative parcels, the right hippocampus and lateral ventricle, showing age-wise ordering of parcel embeddings within each anatomical region.}
  \vspace{-3.8em}
  \label{fig:tsne}
\end{wrapfigure}

This loss encourages the within-parcel age logits to form a smooth unimodal profile centered at the subject's age, while allowing parcel-specific variability through $\sigma_p$ in healthy cohorts.


\paragraph{\textbf{Total training objective:}}
\label{sec:framework_total_loss}

The final objective combines global alignment and within-parcel age shaping:
\begin{equation}
\mathcal{L}
=
\mathcal{L}_{\mathrm{SigLIP}}
+
\lambda_{\mathrm{age}}\;\mathcal{L}_{\mathrm{AgeSoftCE}}
\label{eq:total_loss}
\end{equation}
where $\lambda_{\mathrm{age}}$ is a hyperparameter (set to $0.1$).
BrainNorm's pretraining objective yields a structured SAL space. The parcel and template embeddings are visualized in Fig.~\ref{fig:tsne}, showing two sample parcels that are ordered by age and clustered by parcel type.  
The same objective is used for fine-tuning by restricting exclusively to healthy cohorts.
%



\section{Experiments and Results}
\label{sec:exp}
\subsection{Datasets \& Atlas}

We leveraged healthy participants from the \textit{UK Biobank (UKB)}~\cite{Sudlow2015UKBiobank}, a large-scale, publicly available imaging study comprising 48,018 T1-weighted structural MRI scans, for pretraining. Participants with documented neurological disorders were excluded to ensure a clean, healthy cohort. An external healthy cohort of 2,662 scans from the \textit{Mayo Clinic Study of Aging (MCSA)}~\cite{Roberts2008MCSA} (publicly available) was used as an additional held-out evaluation dataset.
For downstream evaluations and tasks, we utilized publicly available mixed clinical cohorts with 15,352 scans spanning multiple neurodegenerative disorders, including \textit{Alzheimer’s disease} (AD), \textit{Mild Cognitive Impairment} with stable subjects (sMCI) and progressors (pMCI), \textit{Frontotemporal dementia} (FTD), and \textit{Parkinson’s disease} (PD). These datasets include the \textit{ADNI} (phases 1,GO,2,3 \& 4)~\cite{Mueller2005ADNI}, \textit{AIBL}~\cite{Fowler2021AIBL}, \textit{MIRIAD}~\cite{Malone2013}, \textit{NIFD}~\cite{Knopman2014FTLDNI}, and \textit{PPMI}~\cite{Marek2011PPMI}. 
BrainNorm was pretrained using 80\% of the UKB cohort and evaluated on the remaining 20\% held-out test set, along with the full held-out MCSA~\cite{Roberts2008MCSA} dataset. For mixed disease cohorts, all downstream tasks were performed using diagnosis-stratified 5-fold subject-wise cross-validation, with performance reported on the 20\% held-out test split within each fold. Each training set further included a 10\% internal validation subset. All splits were performed subject-wise to prevent data leakage across training, validation, and test sets. Cohort details and statistics are provided in Appendix Sec~\ref{Asec:dataset_overview}.
All scans are preprocessed and registered to MNI-152 1mm isotropic space. For preprocessing details, refer to Appendix Sec~\ref{Asec:preprocessing}. 

To enable anatomically grounded parcel-level modeling, we constructed a unified brain atlas combining:
AAL3v1~\cite{TzourioMazoyer2002AAL} atlas for cortical and subcortical gray matter regions,
the JHU white-matter atlas~\cite{Mori2008JHUAtlas} for white matter tracts, and the Harvard–Oxford atlas~\cite{SCR_001476} for ventricular and cerebrospinal fluid regions (details in Appendix Sec~\ref{Asec:atlas}).
This integration yielded $218$ anatomical parcels covering the entire brain volume.

\subsection{Model and Training Details}
\label{sec:train_details}

BrainNorm employs a transformer-based Parcel Encoder with a ViT backbone operating on $8\times8\times8$ voxel patch volumes. 
The encoder consists of 8 layers with 8 attention heads, an embedding dimension of 768, and a forward dimension of 2048. 
Parcel embeddings are projected into a 1024-dimensional SAL space.
The Template Mapper projects frozen Qwen3 text embeddings of dimension 4096 into the same 1024-dimensional atlas space (details in Appendix Sec~\ref{Asec:template_mapper}). 
In total, BrainNorm has been trained with approximately $67$M trainable parameters.
To construct the age-indexed template bank, we generated approximately $200$K structured texts covering 218 parcels across ages 10–100 years at 0.1-year resolution. Their embeddings were extracted using Qwen3-8B~\cite{qwen3technicalreport2025} model and stored offline. 
With $\sim$38,000 training scans of UKB and the SigLIP objective, pretraining involves alignment across $\sim$8M parcel-template pairs. Batch size was set to 32 scans, which corresponds to approximately 7k parcel-template pairs per batch.
Optimization was performed using AdamW with learning rate $1\times10^{-4}$, weight decay $1\times10^{-4}$, and cosine scheduling with 3k warmup steps.
Training was conducted on 4 NVIDIA RTX A6000 GPUs for over 100 hours for 40k steps.


\subsection{Finetuning for Domain Shifts}

BrainNorm is pretrained on a large healthy UKB cohort. 
However, downstream datasets exhibit distributional shifts arising from scanner manufacturer, field strength (1.5T/3T), acquisition protocol, and demographic variation. 
To assess robustness under such shifts, we evaluate under two settings:

\noindent
\textbf{(i) Direct Transfer (No Finetuning):}  
The pretrained UKB checkpoint is directly deployed to downstream datasets for inference and feature extraction.

\noindent
\textbf{(ii) CN-only Finetuning (FT):}  
BrainNorm is finetuned using Eq~\ref{eq:total_loss} exclusively on cognitively normal (CN) subjects from the downstream training split of ADNI, AIBL, MIRIAD, NIFD, PPMI for each cross-validation split. 
Then both the Parcel encoder and the Template Mapper are frozen. 
Further downstream tasks are then performed using these frozen embeddings of both CN and disease subjects, and repeated for 5-folds.

\begin{figure*}[t]
  \centering
 \includegraphics[width=\linewidth]{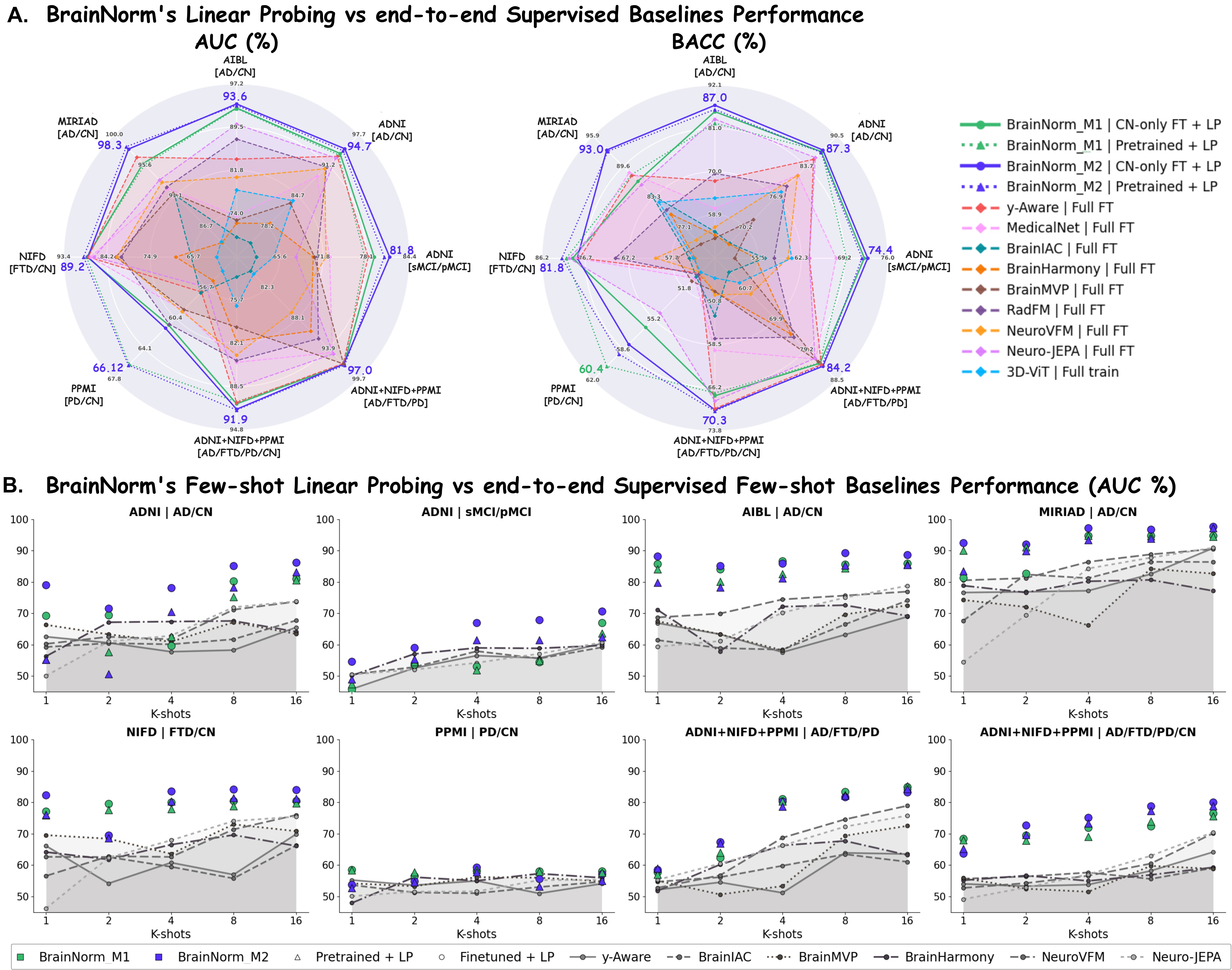}
  
  \caption{\textbf{A. Classification performance} of BrainNorm's full-data linear probing (LP) results of \textit{M1} \& \textit{M2} techniques outperforms all end-to-end supervised (finetuned) baselines. \textbf{B. Few-shot linear probing of BrainNorm} shows superior performance with both \textit{M1} and \textit{M2} representations compared to end-to-end finetuning of pretrained baselines, indicating stronger transferability under small-label budgets. BrainNorm's parcel deviation-based representations (\textit{M2}) shows better results in multiple tasks compared to utilizing only parcel embeddings (\textit{M2}). Plotted values are mean across 5-folds for mixed cohorts.
  }
  \label{fig:Linear_probe_plot}
\end{figure*}

This protocol preserves the normative learning objective while allowing limited cohort-specific calibration. We evaluated BrainNorm's performance against 9 baselines using the same preprocessed scan inputs, fine-tuned/trained under end-to-end task-specific supervision for \emph{(i)} age prediction and \emph{(ii)} disease classification. We finetuned foundation models like NeuroJEPA~\cite{huang2026learning}, NeuroVFM~\cite{kondepudi2026health}, RadFM~\cite{wu2025towards}, 
BrainHarmony~\cite{dong2025brain}, \& BrainIAC~\cite{tak2026brainiac},  and pretrained models like BrainMVP~\cite{rui2025multi}, MedicalNet~\cite{chen2019med3d} \& y-Aware~\cite{dufumier2021yaware}. We further trained the ViT~\cite{Dosovitskiy2020} model from scratch (ViT-B with forward dimension as 2048, having $\sim$$67$M params, i.e., nearly the same capacity as BrainNorm). Across all tasks and cohorts, BrainNorm consistently outperforms the supervised baselines both before and after CN-only finetuning (Fig.~\ref{fig:age_zero}, ~\ref{fig:Linear_probe_plot}).

\begin{table*}[t]
\scriptsize
\centering
\caption{Parcel Identification Performance (Text(region)$\rightarrow$Parcel \& Parcel$\rightarrow$Text(region)) using CN subjects across pretraining datasets and downstream mixed datasets ($mean_{\pm std}$ over 5-folds). Performance improves with finetuning. [FT: Finetuned on CN-only subjects of mixed cohorts]}
\setlength{\tabcolsep}{5pt}
\renewcommand{\arraystretch}{1.2}
\begin{tabular}{llcccccccc}
\hline
\textbf{Type} & \textbf{Metric} & \textbf{Mode} & \textbf{UKB} & \textbf{MCSA} & \textbf{ADNI} & \textbf{AIBL} & \textbf{MIRIAD} & \textbf{NIFD} & \textbf{PPMI} \\
\hline

\multirow{4}{*}{T$\rightarrow$P}
& \multirow{2}{*}{Acc@1} 
& No FT 
& 90.87 & 67.33 & $69.50_{\pm 0.35}$ & $68.71_{\pm 0.65}$ & $73.93_{\pm 1.18}$ & $70.62_{\pm 0.59}$ & $67.18_{\pm 0.85}$ \\
& 
& FT 
& 90.32 & 76.06 & $79.88_{\pm 1.11}$ & $80.01_{\pm 1.25}$ & $83.49_{\pm 2.88}$ & $82.06_{\pm 0.64}$ & $79.47_{\pm 1.09}$ \\
\cline{2-10}
& \multirow{2}{*}{Acc@5} 
& No FT 
& 99.02 & 85.19 & $87.45_{\pm 0.24}$ & $86.20_{\pm 0.48}$ & $90.33_{\pm 0.69}$ & $86.78_{\pm 0.50}$ & $84.93_{\pm 0.58}$ \\
& 
& FT 
& 99.04 & 94.54 & $96.32_{\pm 0.39}$ & $96.01_{\pm 0.50}$ & $97.74_{\pm 0.83}$ & $96.71_{\pm 0.29}$ & $95.59_{\pm 0.56}$ \\
\hline

\multirow{4}{*}{P$\rightarrow$T}
& \multirow{2}{*}{Acc@1} 
& No FT 
& 97.17 & 75.02 & $77.54_{\pm 0.31}$ & $76.64_{\pm 0.83}$ & $81.30_{\pm 1.01}$ & $63.84_{\pm 3.11}$ & $74.84_{\pm 0.60}$ \\
& 
& FT 
& 96.72 & 86.46 & $89.64_{\pm 0.78}$ & $89.89_{\pm 1.18}$ & $93.20_{\pm 1.76}$ & $91.35_{\pm 0.37}$ & $88.71_{\pm 0.54}$ \\
\cline{2-10}
& \multirow{2}{*}{Acc@5} 
& No FT 
& 99.80 & 89.54 & $91.17_{\pm 0.23}$ & $90.27_{\pm 0.43}$ & $93.32_{\pm 0.89}$ & $80.40_{\pm 2.62}$ & $88.72_{\pm 0.34}$ \\
& 
& FT 
& 99.76 & 97.81 & $98.65_{\pm 0.20}$ & $98.64_{\pm 0.30}$ & $99.35_{\pm 0.59}$ & $98.91_{\pm 0.18}$ & $98.13_{\pm 0.27}$ \\
\hline

\end{tabular}
\label{tab:parcel_grouped}
\end{table*}

\subsection{Anatomical Correspondence and Normative Age Inference}

\noindent
\textbf{a. Parcel Identification:}
To assess whether BrainNorm encodes \emph{anatomical identity} in a shared semantic space, we evaluate bidirectional parcel-region retrieval. We ask two questions: (i) given a parcel region, can the model recover its anatomical label? and (ii) given an anatomical label, can the model identify the corresponding parcel in the scan?
Let $\mathbf{p}_{p} \in \mathbb{R}^D$ denote the embedding of parcel $p$ extracted from a scan, and let $\mathbf{t}_{a,q}$ denote the template embedding corresponding to anatomical region $q$ at age bin $a$. Similarity is computed as
$\mathbf{p}_p^\top \mathbf{t}_{a,q}.$
For parcel $\rightarrow$ region retrieval, we rank all templates $\{\mathbf{t}_{a,q}\}_{q=1}^{N}$ given a parcel embedding $\mathbf{p}_p$. 
For region $\rightarrow$ parcel retrieval, we rank all parcel embeddings $\{\mathbf{p}_p\}_{p=1}^{N}$ given a template query $\mathbf{t}_{a,q}$. 
A prediction is considered correct for a ground-truth match within the top-$k$ candidates.
We report Acc@1 and Acc@5 for both in Table~\ref{tab:parcel_grouped}.  
Retrieval accuracy improves consistently after CN-only finetuning, indicating stronger anatomical grounding of the learned representations.

\begin{figure*}[t]
  \centering
 \includegraphics[width=\linewidth]{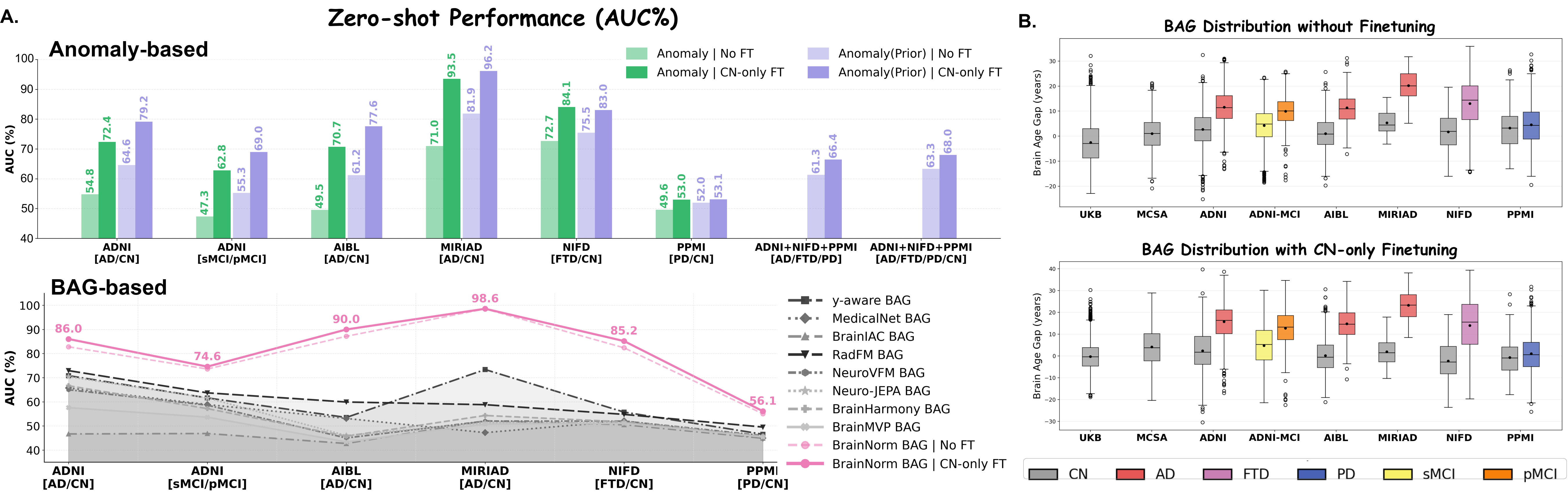}
  \caption{\textbf{A. BrainNorm's zero-shot performance} comparison using anomaly scores and BAG across all classification tasks. These plots illustrate the performance gains of BrainNorm achieved by fine-tuning on CN subjects alone, along with superior performance compared to other supervised baselines across all tasks. Notably, BrainNorm's disease-prior restricted anomaly technique enables zero-shot learning for multi-disease classification. \textbf{B. BAG distribution} improves with finetuning across all datasets and groups.}
  \label{fig:zero_shot}
\end{figure*}

\textbf{b. Retrieval-based Age Estimation:}
Age is inferred via template retrieval across discrete age bins. 
Let $\ell_a$ denote the aggregated similarity between a scan and templates at age bin $a$, computed by averaging parcel-wise similarities.
\begin{wrapfigure}{r}{0.55\textwidth}
  \centering
  \vspace{-1.0em}
\includegraphics[width=0.55\textwidth]{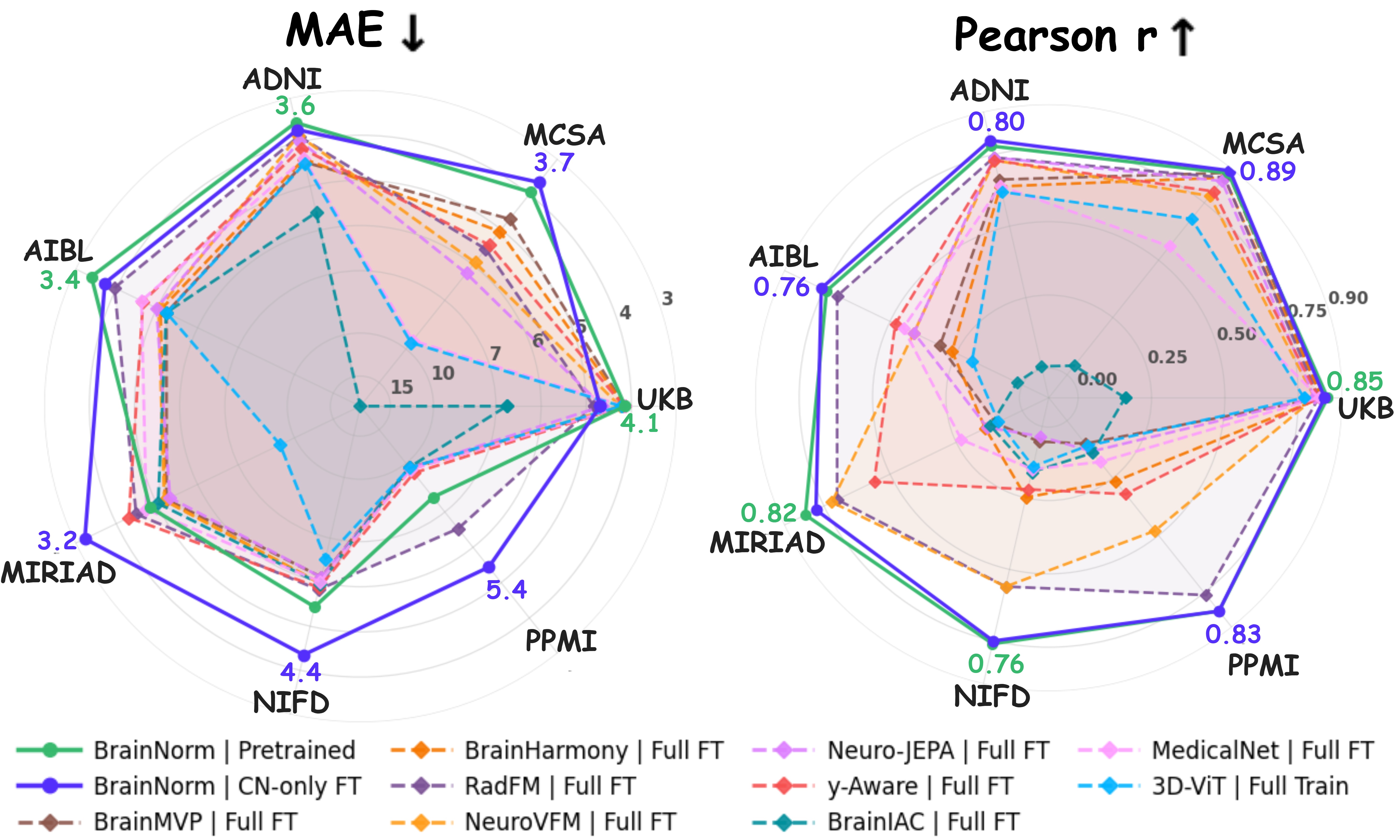}
  \vspace{-1.5em}
  \caption{\textbf{BrainNorm's age estimation} outperforms other supervised baselines' age prediction performance.}
  \vspace{-1.8em}
  \label{fig:age_zero}
\end{wrapfigure}
We construct a temperature-scaled softmax distribution and compute age as:
$
y_{\text{exp}} = \sum_{a\in\mathcal{A}} w_a a,
\quad \text{where},\;\;
w_a = \frac{\exp(\ell_a / \tau)}{\sum_{a'} \exp(\ell_{a'} / \tau)}.
$
\noindent 
We report results using $\tau = 0.5$, which sharpens the age-bin distribution and emphasizes high-similarity bins.
Performance is evaluated using MAE and Pearson correlation. More metrics are in Appendix Table~\ref{tab:age_estimation_methods}. BrainNorm achieves superior age estimation compared to the end-to-end supervised baseline predictions across datasets (Fig.~\ref{fig:age_zero}). Also, BrainNorm demonstrates better generalizability on the held-out MCSA dataset.

\subsection{Neurodegenerative Disease Classification}
\label{sec:classification}

We evaluate disease sensitivity using zero-shot normative deviation scoring and linear probing (LP).













\noindent
\textbf{a. Zero-shot Normative Deviation Scoring:}
Zero-shot anomaly detection is derived directly from normative deviation measures without training a classifier.

\noindent
\textit{(1) Global Anomaly Score:}
Let $s_p$ denote similarity between scan parcel $p$ and its age-matched template $\mathbf{t}_{a,p}$. We define the deviations as top-$K$ anomaly score: $
D_{global} = \frac{1}{K} \sum_{p \in \mathcal{T}_K} (1 - s_p),
$
where $\mathcal{T}_K$ contains the $K$ parcels with the largest deviations (lowest similarity). 

\noindent
\textit{(2) Disease-Prior Restricted Anomaly:}
For disease $d$, top-$K$ deviation aggregation is restricted to identified anatomical parcel sets $\mathcal{P}_d$ from known disease prior knowledge. This produces disease-specific anomaly scores. This enables prior-guided zero-shot disease classification.

\noindent
\textit{(3) BAG-based Deviation:}
As a complementary estimator, we compute parcel-wise age argmax to get subject's age prediction:
$
y_{\text{mean-argmax}} = \frac{1}{|\mathcal{P}|} \sum_{p} a_p^*,
\quad \text{where},\;\;
a_p^* = \arg\max_a s_{i,p}^{a,p}.
$
Then, Brain Age Gap is defined as, $\text{BAG} = y_{\text{mean-argmax}} - y$, where $y$ is chronological age.  
Visualization of BAG distributions across diagnostic groups before and after finetuning are presented in Fig.~\ref{fig:zero_shot}B. CN-only finetuning yields improved group separation across cohorts.
Subjects with larger positive BAG are hypothesized to exhibit accelerated neurodegeneration.

\noindent
The zero-shot results before and after finetuning on healthy subjects are presented in Fig~\ref{fig:zero_shot}A. Notably, the disease-prior-restricted anomaly enables \emph{prior-guided zero-shot multi-disease classification} using BrainNorm. (More details in Appendix Sec~\ref{Asec:zero_Shot})

\noindent
\textbf{b. Linear Probing (few-shot LP and all LP):}
To evaluate the anomaly separability of BrainNorm's representations under supervision, we perform linear probing on frozen embeddings. In particular, we test whether disease-relevant signals can be extracted along: (i) representations from parcel embeddings, and (ii) parcel-wise deviations from age-matched normative templates.



\noindent
\textit{(M1) Parcel Embedding.}
To test whether anatomical parcel representations alone are discriminative, we concatenate parcel embeddings:
$\mathbf{x}_{M1} = concat(\mathbf{p}_1,\dots,\mathbf{p}_N) \in \mathbb{R}^{ND}.$


\noindent
\textit{(M2) Parcel Deviation.}
To capture normative deviation patterns, we compute parcel-wise deviations between scan embeddings and age-matched templates, forming:
$\mathbf{x}_{M2}=concat(\mathbf{p}_1-\mathbf{t}_{a,1},\dots,\mathbf{p}_N-\mathbf{t}_{a,N}) \in \mathbb{R}^{ND}$. KDE plots of UMAP using $\mathbf{x}_{M2}$ show the clusters across the groups in Fig~\ref{fig:viz}B.

Each feature representation is fed to a linear MLP classifier and evaluated under two setups: \textbf{full LP} using all available labeled samples, and \textbf{few-shot LP} using $k$ labeled samples per class ($k \in \{1,2,4,8,16\}$).
Results for full-data linear probing are reported in Fig.~\ref{fig:Linear_probe_plot}A, and few-shot linear probing are reported in Fig.~\ref{fig:Linear_probe_plot}B, 
with mean values across 5-folds. Linear probing with full-data on BrainNorm representations consistently outperforms the end-to-end supervised baselines across all evaluated datasets and tasks, highlighting the advantage of BrainNorm's normative pretraining and CN-only finetuning for identifying disease-relevant representations.

Together, these experiments assess BrainNorm's anatomical faithfulness (parcel identity), normative calibration (age/BAG), disease sensitivity (zero-shot/ linear-probing), and robustness to domain shift (CN-only finetuning). Further experimental results and details with $mean_{\pm std}$ values are provided in Appendix Sec~\ref{Asec:exp}. Refer to Appendix Sec~\ref{Asec:choice_text_encoder} for the choice of text encoder, Sec~\ref{Asec:loss_ablation} for Template Mapper and loss ablations, Sec~\ref{Asec:text_prior_init} for text-prior initialization ablations, and Sec~\ref{Asec:text_query} for prompt-rephrasing experiments demonstrating BrainNorm's flexibility in querying the SAL space. More details of the BrainNorm's embeddings and confounder effects are discussed in Appendix Sec~\ref{Asec:UMAP}.

\section{Interpretability}
\begin{figure*}[!t]
  \centering
 \includegraphics[width=0.98\linewidth]{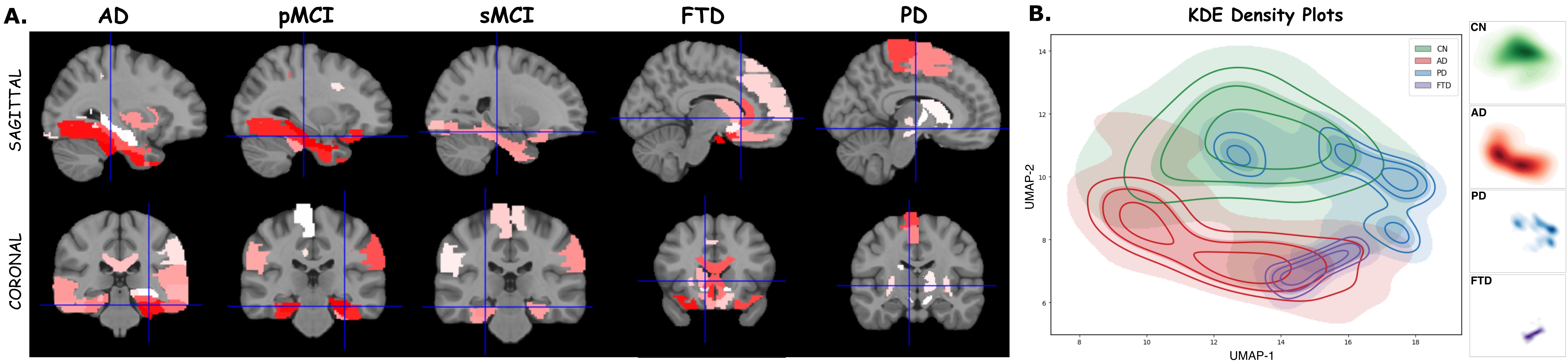}
  \vspace{-0.5em}
  \caption{\textbf{A. Zero-shot identification} and visualization of the top-$k$ anomalous brain parcels across disease cohorts in sagittal and coronal planes. Higher deviation scores from the normative reference are represented by darker shades. \textbf{B. KDE density plot} of UMAP using the parcel residuals ($\mathbf{x}_{M2}$).}
  \label{fig:viz}
\end{figure*}

We visualized the top-$K$ anomalous brain regions in Fig.~\ref{fig:viz} by computing zero-shot parcel-wise anomaly scores ($D_{global}$) relative to age-conditioned templates. For each disease cohort, individual deviation scores are averaged across the test set to generate cohort-level anomaly profile. These top-$K$ parcel regions are then mapped back into volumetric space to visualize spatial distributions of disease-related deviations. Resulting maps align with clinical neurodegeneration literature: specifically, involvement of the hippocampus and parahippocampal gyrus in AD and MCI~\cite{Planche2022,Braak2006}; the frontal/prefrontal cortex and temporal lobes in FTD~\cite{Rascovsky2011,GornoTempini2011}; and the motor cortex and subcortical structures in PD~\cite{Hughes1992,Postuma2015}. Futher scan-level interpretations are discussed in Appendix Sec~\ref{Asec:scan_interpret}

\section{Discussion and Conclusion}
\label{sec:conc}
We presented BrainNorm, the first normative foundation model that learns healthy age-conditioned, region-specific representations within a structured Semantic Atlas Latent space. By leveraging SigLIP-style pretraining, BrainNorm yields representations that outperform fully supervised baselines across multiple tasks.
BrainNorm serves as a unified framework for: age estimation and brain-age gap analysis, brain region identification, and both zero-shot and linear probing for disease classification (e.g., AD, PD, FTD, MCI), without task-specific retraining. 
Our current setup relies on registration to MNI152 space to ensure spatial consistency and enable reliable atlas-based parcel tokenization. The framework is currently built on a single imaging modality (T1w sMRI), and uses an adult brain atlas for defining parcel-level representations. As a result, although the overall approach can be extended across broader age ranges, the present implementation is restricted to adult cohorts due to the atlas choice. Future work will focus on reducing registration dependence, studying multi-atlas combinations at varying levels of granularity, exploring adaptive parcellation techniques, and extending the framework to conditions such as Neurodevelopmental disorders using additional imaging modalities (T2w, fMRI, etc.). 


\section*{Acknowledgement}
This research has been conducted using the UK Biobank resource under Project ID 335526. We gratefully acknowledge the UK Biobank participants and investigators for making this research resource available. We are also grateful for the access to other datasets, including MCSA, ADNI, AIBL, MIRIAD, NIFD, and PPMI.



{\small
\bibliographystyle{unsrt} 
\bibliography{main} 
}






\appendix
\newpage


\clearpage
\appendix

\begin{center}
    \Large\bfseries Appendix
\end{center}

\startcontents[appendix]
\section*{Contents}

\printcontents[appendix]{}{1}{\setcounter{tocdepth}{3}}


\setcounter{section}{0}
\setcounter{figure}{0}
\setcounter{table}{0}
\renewcommand{\thefigure}{A\arabic{figure}}
\renewcommand{\thetable}{A\arabic{table}}

\section{Datasets and Preprocessing Details}
\label{Asec:datasets}

\subsection{Dataset Overview}
\label{Asec:dataset_overview}


\paragraph{\textbf{Downstream Clinical Cohorts.}}
For downstream evaluations and tasks, we utilized publicly available mixed clinical cohorts totalling 15,352 scans, spanning multiple neurodegenerative disorders: \textit{Alzheimer's disease} (AD), \textit{Mild Cognitive Impairment} with stable subjects (sMCI) and progressors (pMCI), \textit{Frontotemporal dementia} (FTD), and \textit{Parkinson's disease} (PD). The constituent datasets are described below.

\begin{itemize}

    \item \textbf{ADNI:} The \textit{Alzheimer's Disease Neuroimaging Initiative} (phases 1, GO, 2, 3, and 4)~\cite{Mueller2005ADNI} is a large multi-center study spanning dozens of North American sites (U.S./Canada), with contributions across Siemens, GE, and Philips scanners at 1.5T and 3T, and protocol families such as MPRAGE. It offers broad diagnostic coverage (CN/MCI/AD) with multi-year follow-ups, yielding substantial inter-site and inter-scanner variability in acquisition parameters and subject demographics. ADNI cohort comprises of 493 AD and 1210 CN subjects (corresponding to 1485 AD scans and 4142 CN scans). MCI subjects are further stratified into \textbf{stable MCI (sMCI)}, for subjects who never progress to dementia, and \textbf{progressive MCI (pMCI)}, for subjects who subsequently convert. The \emph{baseline} is defined as the first MCI diagnosis visit. All scans of sMCI subjects are labeled \texttt{sMCI}, while for pMCI subjects, scans from baseline up to, but \emph{not including} the first dementia diagnosis, are labeled \texttt{pMCI}, with the conversion scan excluded to avoid conflating MCI-stage biomarkers with dementia-stage pathology. The MCI cohort comprises 396 subjects (3915 scans), categorized as sMCI, and 291 subjects (1393 scans) as pMCI.

    \item \textbf{AIBL:} The \textit{Australian Imaging, Biomarkers and Lifestyle} study~\cite{Fowler2021AIBL} is a prospective Australian cohort with imaging conducted primarily at multiple sites in Melbourne and Perth, predominantly on Siemens platforms (1.5T/ 3T). It provides serial T1-weighted scans with rich clinical and biomarker metadata. AIBL cohort includes 104 AD and 500 CN individuals (154 AD scans and 930 CN scans).

    \item \textbf{MIRIAD:} The \textit{Minimal Interval Resonance Imaging in Alzheimer's Disease} dataset~\cite{Malone2013} is a single-center cohort from London (UCL), acquired on a fixed scanner and protocol with dense revisit intervals (weeks to months). The dataset includes 46 AD and 23 CN participants (corresponding to 465 AD scans and 243 CN scans).
        
    \item \textbf{NIFD:} The \textit{Neuroimaging in Frontotemporal Dementia} dataset 
    (also known as FTLDNI)~\cite{Knopman2014FTLDNI} is a multi-site NIH-funded study 
    designed to characterise structural and functional neuroimaging signatures across the 
    FTD spectrum. Data were acquired across three academic medical centres: the University 
    of California San Francisco (UCSF), Mayo Clinic (Rochester), and Massachusetts General 
    Hospital (MGH). T1-weighted MRI scans were acquired at 3T across sites, with each site contributing scanner and protocol-specific variability. The cohort includes 214 FTD patients and 118 cognitively normal controls (605 FTD scans and 357 CN scans).

\item \textbf{PPMI:} The \textit{Parkinson's Progression Markers Initiative}~\cite{Marek2011PPMI} 
is a large-scale, multi-center, international longitudinal study launched by the 
\textit{Michael J. Fox Foundation for Parkinson's Research}, with sites spanning 
North America and Europe. The study was designed to identify progression biomarkers 
for Parkinson's disease across clinical, imaging, and biological modalities. 
Neuroimaging was conducted at both 1.5T and 3T field strengths across multiple 
scanner platforms (Siemens, GE, Philips), introducing meaningful inter-site and 
inter-scanner variability. The cohort includes 
newly diagnosed, untreated PD patients at enrollment alongside healthy controls, 
with longitudinal follow-up visits. The subset used in this study consists of 734 PD patients and 213 controls (1360 PD scans and 303 CN scans).     
The broad geographic and institutional diversity of PPMI provides a rigorous testbed for evaluating the model's generalizability to PD-related neurodegeneration.

\end{itemize}





Detailed per-cohort subject and scan counts, 
stratified by diagnosis, scanner, and field strength, are provided in Table~\ref{tab:cohort_adni_stats}.


\begin{table*}[t]
\caption{Cohort statistics of all datasets used in our experiments.}
\centering
\scriptsize
\setlength{\tabcolsep}{1.5pt}
\renewcommand{\arraystretch}{1}

\begin{tabular}{lccccccc}

\toprule
\textbf{Dataset} & \textbf{\#scans} & \textbf{\#subjects} & \textbf{Age (y)} & \textbf{Gender} & \textbf{Field strength} & \textbf{Manufacturer} \\
 &  &  &  \textbf{(min,max,$\text{mean}_{\pm\text{std}}$)} & \textbf{(M / F)} & \textbf{(1.5T / 3.0T)} & \textbf{(Siemens/GE/Philips)} \\
\midrule

UKB (CN)
& 48{,}018
& 43{,}631
& $(44.0,83.0,64.23_{\pm7.70})$
& 20{,}592 / 23{,}039
& 0 / 43{,}631
& 43{,}631 / 0 / 0 \\

MCSA (CN)
& 2{,}662
& 1{,}611
& $(49.0,89.9,71.83_{\pm9.83})$
& 853 / 758
& 0 / 1{,}611
& 0 / 1{,}611 / 0 \\

\midrule

ADNI   
& 5{,}627 
& 1{,}703 
& $(52.4,93.1,72.93_{\pm7.87})$ 
& 734 / 969 
& 429 / 1{,}411 
& 1{,}063 / 476 / 289 \\

ADNI (MCI)   
& 5{,}308 
& 687 
& $(55.6,91.6,74.18_{\pm7.44})$ 
& 408 / 279 
& 275 / 512 
& 389 / 258 / 134 \\

AIBL   
& 1{,}084 
& 599 
& $(56.3,93.0,73.36_{\pm6.42})$ 
& 255 / 343 
& 107 / 541 
& 599 / 0 / 0 \\

MIRIAD 
& 708 
& 69 
& $(56.0,86.3,69.89_{\pm6.98})$ 
& 31 / 38 
& 69 / 0 
& 0 / 69 / 0 \\

NIFD
& 962
& 332
& $(10.1,87.1,64.18_{\pm8.16})$
& 176 / 156
& 0 / 332
& 317 / 16 / 0 \\

PPMI
& 1{,}663
& 947
& $(31.0,85.0,62.97_{\pm10.01})$
& 576 / 371
& 214 / 737
& 548 / 218 / 171 \\

\bottomrule
\end{tabular}
\label{tab:cohort_adni_stats}
\end{table*}

\subsection{Preprocessing Pipeline}
\label{Asec:preprocessing}

The UKB data considered was already partially preprocessed and registered to MNI space. Thus, only z-score normalization within the brain mask was additionally performed.
All other T1-weighted structural MRI scans of downstream cohorts underwent an identical
preprocessing pipeline, applied uniformly across all cohorts. Standardizing preprocessing in this way is essential to place all volumes within a 
common anatomical reference frame so that the parcel-level atlas maps consistently to the same structures across 
subjects, sites, and scanners.

The pipeline begins with the reorientation of each scan to a canonical 
via \texttt{fslreorient2std} (FSL), 
bias field correction  using
\texttt{N4BiasFieldCorrection}. The resulting 
intensity-homogenized volume is then skull-stripped using 
SynthStrip~\cite{Andrew2022}.
Further, each scan is registered to the MNI152~\cite{Mazziotta2001} 1\,mm isotropic standard template
through affine+nonlinear registration using FSL (\texttt{FLIRT}$\rightarrow$\texttt{FNIRT}).
Voxel intensities are then z-score normalized 
within the brain mask.
Finally, the normalized volumes are padded symmetrically from their native 
MNI grid of $182 \times 218 \times 182$ voxels to $184 \times 224 \times 184$ 
voxels,
to be divisible by the 3D patch size ($8^3$) used in the model architecture.

\subsection{Brain Atlas Construction}
\label{Asec:atlas}

 We constructed a 
unified whole-brain atlas in MNI152 space by integrating three complementary, 
publicly available atlases, each providing optimal coverage for a distinct tissue 
compartment to enable anatomically grounded, parcel-level modelling where each input token or 
feature corresponds to a well-defined neuroanatomical structure. The Atlases utilised are:-

\begin{itemize}

    \item \textbf{AAL3v1}~\cite{TzourioMazoyer2002AAL} was used to define 
    \textbf{cortical and subcortical gray matter} regions. AAL3v1 extends the 
    original AAL atlas with improved subcortical delineations, providing 
    comprehensive bilateral parcellation of cortical gyri and subcortical 
    nuclei (including thalamus, caudate, putamen, pallidum, hippocampus, amygdala, 
    and accumbens), all of which are structures of direct relevance in 
    neurodegenerative disease progression.

    \item \textbf{JHU White-Matter Atlas}~\cite{Mori2008JHUAtlas} was used to 
    define \textbf{white matter tract} regions. The JHU atlas provides labeled 
    regions for major white matter fasciculi including the corticospinal tracts, 
    corpus callosum, superior and inferior longitudinal fasciculi, and 
    uncinate fasciculus, among others. Including white matter as explicit parcels 
    allows the model to capture tract-level structural changes, which are prominent 
    in several disorders represented in cohorts (e.g., FTD, PD).

    \item \textbf{Harvard–Oxford Atlas}~\cite{SCR_001476} was used to define 
    ventricular regions with cerebrospinal fluid (CSF), including the 
    lateral ventricles, third and fourth ventricles, and surrounding CSF spaces. 
    Ventricular enlargement is a reliable, high-sensitivity biomarker of global 
    atrophy in neurodegenerative conditions, and its explicit inclusion as a 
    modeled compartment ensures this signal is captured at the parcel level rather 
    than treated as background.

\end{itemize}

\paragraph{\textbf{Atlas integration.}}  The final integrated atlas yields $218$ anatomical parcels providing complete volumetric coverage of 
the brain, spanning gray matter cortex, subcortical nuclei, white matter tracts, 
and ventricular/CSF spaces. The list of parcels considered from each atlas, along with their label IDs, is reported in Table~\ref{tab:atlas_parcels}.



\section{Experiment details and More Results}
\label{Asec:exp}

\begin{figure*}[t]
  \centering
 \includegraphics[width=0.6\linewidth]{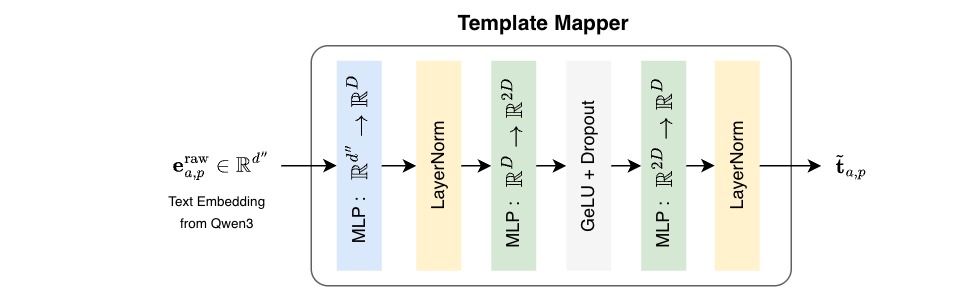}
  \caption{Details of the Template Mapper module.}
  \label{fig:template_mapper}
\end{figure*}


\subsection{Template Mapper details}
\label{Asec:template_mapper}
The template mapper is a 3-layer MLP as presented in Fig~\ref{fig:template_mapper}. The first layer projects the high dimensional text embedding to the SAL's dimension. The later part of the module is inspired from the transformer's Feed Forward network module with GeLU non-linearity. This allows the template embeddings to adapt according to the parcel representations.

\subsection{Ablations}

\subsubsection{Template Mapper \& Loss Ablations}
\label{Asec:loss_ablation}

While the frozen text encoder provides a semantically structured representation with parcel clustering and age ordering, the observed separability and continuity in the visualization alone do not establish that these representations are sufficiently aligned with the parcels' biological aging trajectories. The Template Mapper is introduced to adapt these representations to learn image-derived, parcel-specific aging patterns using the AgeSoftCE loss. We conducted targeted ablations of the Template Mapper and the soft age objective to isolate their respective contributions. These variants are evaluated on parcel identification and age prediction tasks on the UKB test set and the external MCSA cohort.

\begin{table}[t]
\centering
\scriptsize
\caption{\textbf{Ablation of the Template Mapper and AgeSoftCE objective.}
(1) SigLIP + Template Mapper + AgeSoftCE (ours);
(2) SigLIP + Template Mapper, without AgeSoftCE;
(3) SigLIP without Template Mapper and AgeSoftCE}
\label{tab:template_mapper_ablation}

\begin{tabular}{lccccccc}
\toprule

\multirow{2}{*}{\textbf{Dataset}} &
\multirow{2}{*}{\textbf{Ablation}} &
\multicolumn{2}{c}{\textbf{Text$\rightarrow$Parcel}} &
\multicolumn{2}{c}{\textbf{Parcel$\rightarrow$Text}} &
\multirow{2}{*}{\textbf{Age-MAE $\downarrow$}} &
\multirow{2}{*}{\textbf{Pearson-$r$ $\uparrow$}} \\

\cmidrule(lr){3-4}
\cmidrule(lr){5-6}

& &
\textbf{Acc@1 $\uparrow$} &
\textbf{Acc@5 $\uparrow$} &
\textbf{Acc@1 $\uparrow$} &
\textbf{Acc@5 $\uparrow$} &
& \\

\midrule

UKB  & 1 & 90.87 & 99.02 & 97.17 & 99.80 & \textbf{4.14} & \textbf{0.85} \\
UKB  & 2 & \textbf{90.99} & \textbf{99.09} & \textbf{97.99} & \textbf{99.99} & 4.98 & 0.77 \\
UKB  & 3 & 0.48 & 2.34 & 0.32 & 2.05 & 7.74 & 0.44 \\

\midrule

MCSA & 1 & 67.33 & 85.19 & \textbf{75.02} & \textbf{89.54} & \textbf{3.94} & \textbf{0.88} \\
MCSA & 2 & \textbf{70.53} & \textbf{86.60} & 74.19 & 88.12 & 6.78 & 0.84 \\
MCSA & 3 & 0.33 & 2.36 & 0.62 & 2.37 & 8.63 & 0.50 \\

\bottomrule
\end{tabular}
\end{table}

The results in Table~\ref{tab:template_mapper_ablation} clearly separate the roles of the two components. Removing the Template Mapper causes retrieval accuracy to collapse to near-chance levels and substantially degrades age prediction on both datasets, confirming that the frozen text embeddings alone are not aligned with the image-derived parcel representations. Removing only AgeSoftCE largely preserves parcel retrieval performance but worsens age MAE and Pearson correlation, particularly on the external MCSA cohort. This indicates that the soft age objective is specifically responsible for imposing an age-sensitive structure on the learned parcel representations. Thus, the Template Mapper establishes cross-modal parcel alignment, while AgeSoftCE promotes the continuous age structure required for normative modelling.

\subsubsection{Text-Prior Initialization Ablations}
\label{Asec:text_prior_init}

We directly evaluate the role of the text prior by replacing Qwen3 templates with:
\begin{enumerate}
    \item randomly initialized embeddings,
    \item fixed categorical embeddings, and
    \item 218 learnable parcel embeddings modulated by age through FiLM conditioning~\cite{perez2018film}.
\end{enumerate}

\begin{table}[t]
\centering
\scriptsize
\caption{\textbf{Ablation of text-prior initialization strategies.}
We compare Qwen3-based template initialization against randomly initialized embeddings,
fixed categorical embeddings, and learnable parcel embeddings with FiLM-based age conditioning.}
\label{tab:text_prior_ablation}
\begin{tabular}{llcccccc}
\toprule

\multirow{2}{*}{\textbf{Dataset}} &
\multirow{2}{*}{\textbf{Initialization}} &
\multicolumn{2}{c}{\textbf{Text$\rightarrow$Parcel}} &
\multicolumn{2}{c}{\textbf{Parcel$\rightarrow$Text}} &
\multirow{2}{*}{\textbf{Age-MAE $\downarrow$}} &
\multirow{2}{*}{\textbf{Pearson-$r$ $\uparrow$}} \\

\cmidrule(lr){3-4}
\cmidrule(lr){5-6}

& &
\textbf{Acc@1 $\uparrow$} &
\textbf{Acc@5 $\uparrow$} &
\textbf{Acc@1 $\uparrow$} &
\textbf{Acc@5 $\uparrow$} &
& \\

\midrule

UKB  & Random      & 55.93 & 58.34 & 56.02 & 58.19 & 10.32 & 0.68 \\
UKB  & Categorical & \textbf{99.99} & \textbf{99.99} & \textbf{99.98} & \textbf{99.99} & 10.05 & 0.73 \\
UKB  & Learnable   & 91.20 & 99.21 & 96.91 & 97.71 & 7.29 & 0.79 \\
UKB  & Qwen3       & 90.87 & 99.02 & 97.17 & 99.80 & \textbf{4.14} & \textbf{0.85} \\

\midrule

MCSA & Random      & 15.37 & 17.80 & 14.97 & 17.65 & 17.40 & 0.67 \\
MCSA & Categorical & \textbf{94.99} & \textbf{98.17} & \textbf{92.83} & \textbf{97.67} & 17.13 & 0.70 \\
MCSA & Learnable   & 68.52 & 86.33 & 73.24 & 86.53 & 8.90 & 0.78 \\
MCSA & Qwen3       & 67.33 & 85.19 & 75.02 & 89.54 & \textbf{3.66} & \textbf{0.89} \\

\bottomrule
\end{tabular}%
\end{table}

Results on the UKB test set and the external MCSA cohort are reported in Table~\ref{tab:text_prior_ablation}.
These results show that categorical and learnable embeddings can preserve parcel identity, but are substantially weaker for age modelling. In particular, categorical initialization achieves near-perfect parcel retrieval on UKB and remains strong on MCSA, yet exhibits considerably higher age-prediction error. In contrast, Qwen3 achieves the lowest age MAE and highest Pearson correlation on both UKB and MCSA. This suggests that the semantic structure provided by the pretrained text representations yields more effective and generalizable age-conditioned parcel representations than parcel identity alone.

\subsubsection{Text Query Ablations}
\label{Asec:text_query}

BrainNorm's textual inputs provide flexibility in querying the SAL space, rather than constraining the model to a single fixed template. We evaluate four semantically equivalent text-query formulations with different syntax and word order:

\begin{itemize}
    \item \textbf{Template 1 (ours):}
    Parcel=\texttt{<parcel\_name>}; Tissue\_class=\texttt{<tissue>}; Age=\texttt{<a>}y

    \item \textbf{Template 2:}
    The parcel is \texttt{<parcel\_name>}, its tissue class is \texttt{<tissue>},
    and the subject is \texttt{<a>} years old.

    \item \textbf{Template 3:}
    \texttt{<parcel\_name>} of a \texttt{<a>}-year-old subject with
    \texttt{<tissue>} tissue.

    \item \textbf{Template 4:}
    A \texttt{<a>}-year-old subject showing the \texttt{<parcel\_name>},
    which has \texttt{<tissue>} tissue.
\end{itemize}

\begin{table}[t]
\centering
\scriptsize
\caption{\textbf{Robustness to semantic rephrasing of text queries.}
We evaluate four semantically equivalent prompt formulations that vary in syntax and word order.}
\label{tab:prompt_rephrasing}
\begin{tabular}{lccccccc}
\toprule

\multirow{2}{*}{\textbf{Dataset}} &
\multirow{2}{*}{\textbf{Template}} &
\multicolumn{2}{c}{\textbf{Text$\rightarrow$Parcel}} &
\multicolumn{2}{c}{\textbf{Parcel$\rightarrow$Text}} &
\multirow{2}{*}{\textbf{Age-MAE $\downarrow$}} &
\multirow{2}{*}{\textbf{Pearson-$r$ $\uparrow$}} \\

\cmidrule(lr){3-4}
\cmidrule(lr){5-6}

& &
\textbf{Acc@1 $\uparrow$} &
\textbf{Acc@5 $\uparrow$} &
\textbf{Acc@1 $\uparrow$} &
\textbf{Acc@5 $\uparrow$} &
& \\

\midrule

UKB  & 1 & \textbf{90.87} & \textbf{99.02} & \textbf{97.17} & \textbf{99.80} & 4.14 & 0.85 \\
UKB  & 2 & 86.10 & 97.87 & 94.49 & 99.66 & 4.47 & 0.85 \\
UKB  & 3 & 85.96 & 97.89 & 93.43 & 99.58 & 3.99 & 0.85 \\
UKB  & 4 & 80.39 & 96.57 & 93.48 & 99.45 & \textbf{3.91} & 0.85 \\

\midrule

MCSA & 1 & \textbf{67.33} & \textbf{85.19} & \textbf{75.02} & \textbf{89.54} & 3.94 & 0.88 \\
MCSA & 2 & 62.32 & 83.30 & 71.67 & 88.76 & 3.88 & \textbf{0.89} \\
MCSA & 3 & 61.80 & 83.03 & 70.27 & 88.22 & \textbf{3.71} & 0.88 \\
MCSA & 4 & 60.23 & 81.78 & 69.60 & 87.64 & 3.80 & 0.88 \\

\bottomrule
\end{tabular}%
\end{table}

The results are presented in Table~\ref{tab:prompt_rephrasing}. Despite changes in syntax and word order, all prompt formulations retain strong parcel retrieval and comparable age-estimation performance across both the UKB test set and the external MCSA cohort. While retrieval performance decreases moderately for more substantially rephrased prompts, age MAE and Pearson correlation remain highly stable, indicating that the learned SAL representations preserve the underlying age-related structure across semantically equivalent formulations. These results demonstrate that BrainNorm is not restricted to a single fixed textual template and that the SAL space can be queried through multiple equivalent natural-language descriptions without retraining. This flexibility is inherently unavailable to categorical or learnable parcel-ID embeddings, which do not support semantic rephrasing at inference time.

\begin{table*}[t]
\centering
\scriptsize
\setlength{\tabcolsep}{3.2pt}
\caption{
\textbf{Choice of Text Encoder.} Comparison of text encoders for the desired embedding geometry.
The representation should (1) cluster samples by \textbf{parcel identity}
and (2) exhibit a smooth \textbf{age trajectory within each parcel}.
Metrics on the left measure parcel clustering, while metrics on the right
measure age ordering within parcels. Best values are highlighted in bold.
}
\label{tab:text_embedding_geometry}

\begin{tabular}{l|ccc|ccc}
\toprule

& \multicolumn{3}{c}{\textbf{Parcel Clustering}}
& \multicolumn{3}{|c}{\textbf{Age Ordering Within Parcel}} \\

\cmidrule(lr){2-4}
\cmidrule(lr){5-7}

\textbf{Encoder}
& Silhouette $\uparrow$
& Hit@1 $\uparrow$
& NMI $\uparrow$
& Spearman $\uparrow$
& Pair Order Acc $\uparrow$
& kNN Age MAE $\downarrow$ \\

\midrule

\textbf{Qwen3-Embedding-8B}~\cite{qwen3technicalreport2025}
& 0.4637
& 0.9999
& 0.9476
& \textbf{0.9682}
& \textbf{0.9311}
& \textbf{0.5114} \\

MedEmbed-small~\cite{balachandran2024medembed}
& \textbf{0.6667}
& \textbf{1.0000}
& \textbf{0.9962}
& 0.5483
& 0.6922
& 1.9061 \\

MedSigLIP~\cite{sellergren2025medgemma} 
& 0.1224 
& 0.7284 
& 0.8790 
& 0.2906 
& 0.5971 
& 6.0694 \\

BiomedCLIP~\cite{zhang2024biomedclip}
& 0.6133
& 0.9999
& 0.9863
& 0.3816
& 0.6323
& 2.4691 \\

Hashing+SVD
& 0.4440
& 0.9998
& 0.9939
& 0.0024
& 0.5010
& 3.4503 \\

TF-IDF
& 0.2225
& 0.0412
& 0.9928
& 0.0002
& 0.4996
& 4.1861 \\

\bottomrule
\end{tabular}

\end{table*}

\subsubsection{Choice of Text Encoder}
\label{Asec:choice_text_encoder}

\paragraph{Why semantic atlas pretraining instead of direct trajectory regression?}
Although the retrieval-based age estimator resembles non-parametric regression at inference time, BrainNorm differs in the representation over which retrieval is performed. Classical trajectory estimators such as k-NN, kernel regression, Gaussian processes, or brain-chart models typically operate on predefined morphometric summaries or low-dimensional regional statistics. In contrast, BrainNorm learns high-dimensional parcel-level representations directly from structural MRI and aligns them with age-conditioned semantic atlas descriptors. The text encoder is therefore not used to inject unconstrained natural-language knowledge, but to provide a structured semantic coordinate system over parcel identity, tissue class, and age. This encourages prompts from the same anatomical parcel to remain coherent while allowing age-conditioned prompts to form smooth within-parcel trajectories. Consequently, age retrieval, brain age gap estimation, parcel identification, and zero-shot anomaly scoring are all performed in a shared normative representation space rather than through independent per-task regressors.
This is broader than a non-parametric age trajectory estimator.
The retrieval formulation is therefore used as an interpretable inference mechanism over the learned normative representation, rather than as the primary modeling contribution itself.

\paragraph{Selecting a semantic embedding space with parcel-aware age geometry:}
BrainNorm's representation requires a specific geometric structure in the SAL space. 
First, prompts corresponding to the same anatomical parcel should form coherent clusters, ensuring that the embedding preserves parcel identity. 
Second, within each parcel cluster, embeddings corresponding to different ages should follow a smooth and monotonic trajectory that reflects ageing. 
This structure allows the model to interpret age-conditioned prompts as continuous trajectories within each anatomical region.

To identify an embedding model that satisfies these requirements, we evaluate several candidate encoders using two complementary groups of metrics. 

\paragraph{Parcel clustering metrics:}
These metrics evaluate whether prompts corresponding to the same anatomical parcel remain grouped together in the embedding space. 
We first compute the \textit{silhouette score}, which measures how well each sample is separated from samples of other parcels relative to samples from its own parcel. Higher values indicate that embeddings belonging to the same parcel form compact clusters that are clearly separated from other parcels. 
To evaluate local parcel consistency, we compute \textit{nearest-neighbor parcel retrieval (Hit@1)}, defined as the fraction of samples whose nearest neighbor in the embedding space belongs to the same parcel. 
Finally, we measure \textit{normalized mutual information (NMI)} between the ground-truth parcel assignments and clusters obtained by applying $k$-means to the embeddings (with $k$ equal to the number of parcels). NMI evaluates how well the clustering structure induced by the embeddings aligns with the true anatomical parcel labels.

\paragraph{Age trajectory metrics:}
These metrics assess whether embeddings within a parcel exhibit a consistent progression with age. 
For each parcel, we first compute the \textit{Spearman rank correlation} between age and the dominant direction of variation in the embeddings, obtained via principal component analysis. A high correlation indicates that age progression aligns with the principal trajectory of the parcel embeddings. 
We additionally evaluate \textit{pairwise age-order accuracy}, which measures the fraction of sample pairs within a parcel for which the relative ordering of their embedding coordinates matches the chronological ordering of their ages. This metric quantifies whether the embedding space preserves the temporal ordering of samples. 
Finally, we compute the \textit{local nearest-neighbor age error}, defined as the mean absolute difference in age between each sample and its nearest neighbors within the same parcel. Low error values indicate that nearby embeddings correspond to samples with similar ages, reflecting a smooth age trajectory within each parcel.

Table~\ref{tab:text_embedding_geometry} reports the results across several biomedical and SOTA text encoders as well as classical baselines (TF-IDF and Hashing+SVD). These classical methods embeddings do not guarantee smooth age geometry and semantic organization simultaneously unless explicitly regularized. 
While several encoders achieve strong parcel discrimination, \textbf{Qwen3-Embedding-8B} produces the most consistent age-structured geometry. 
In particular, it yields substantially higher parcel-wise age monotonicity and pairwise age-order accuracy while simultaneously achieving the lowest local age prediction error. 
These results indicate that Qwen3 embeddings naturally organize prompts into parcel-specific manifolds where age varies smoothly along the dominant direction. 
Consequently, we adopt Qwen3 as the text encoder for constructing \textit{parcel-conditioned age-conditioned representations} for BrainNorm.

\subsection{Implementation details of SigLIP loss}

In our implementation, SigLIP alignment is performed between parcel embeddings extracted from MRI scans and a bank of age-conditioned textual templates representing anatomical parcels. We perform distributed training across 4 GPUs.

\paragraph{Distributed pair construction:}
Across all GPUs, the union of these ages is gathered using distributed communication. Template coordinates for the corresponding $(p,a)$ unique combinations are then retrieved from the template bank.
To efficiently compute similarities across a large number of parcel templates, the template bank is sharded across GPUs along the parcel dimension. Each GPU retrieves and processes templates for a subset of parcels, computes the mapped template embeddings locally, and then gathers these embeddings across GPUs using distributed communication to obtain all similarities of the template bank in that batch. This strategy avoids replicating the entire template bank on every GPU while still allowing each batch to compute similarities against the full set of parcel–age templates.

\paragraph{Positive pair selection:}
Positive supervision is defined using parcel identity and an age window of $\pm 1$ year around the subject age. For a given parcel embedding, all templates corresponding to the same parcel and whose ages fall within this window are treated as positive pairs. 
When multiple valid templates exist within the age window, their contributions are handled using soft targets rather than selecting a single positive pair. This allows nearby age templates to contribute to the learning signal while preserving the strongest supervision at the exact age when available.

\paragraph{Similarity computation:}
After gathering the template bank across GPUs, cosine similarities between parcel embeddings and all candidate templates are computed in the shared embedding space. These similarities are used to compute the SigLIP binary cross-entropy loss as described in the main paper.


\begin{table*}[t]
\scriptsize
\centering
\caption{Parcel identification on held-out test sets and downstream datasets using only CN subjects (5-fold mean$_{\pm \mathrm{std}}$ where available). PT: pretrained encoder used directly for inference. FT: finetuned only on the CN subset of the corresponding dataset or finetuned on pooled CN data from ADNI+AIBL+MIRIAD+NIFD+PPMI (All).}
\setlength{\tabcolsep}{5.0pt}
\renewcommand{\arraystretch}{1.15}
\begin{tabular}{ccccccc}
\hline
\multirow{2}{*}{\textbf{Dataset}} & \multirow{2}{*}{\textbf{Mode}} & \multirow{2}{*}{\textbf{CN data}}
& \multicolumn{2}{c}{\textbf{Text$\rightarrow$Parcel}} 
& \multicolumn{2}{c}{\textbf{Parcel$\rightarrow$Text}} \\
\cmidrule(lr){4-5}\cmidrule(lr){6-7}
& & 
& \textbf{Acc@1 $\uparrow$} & \textbf{Acc@5 $\uparrow$}
& \textbf{Acc@1 $\uparrow$} & \textbf{Acc@5 $\uparrow$} \\
\hline

\textbf{UKB}
& No FT & UKB
& $90.87$ & $99.02$ & $97.17$ & $99.80$ \\
& FT & UKB
& $90.32$ & $99.04$ & $96.72$ & $99.76$ \\
\hline

\multirow{2}{*}{\textbf{MCSA}}
& No FT & UKB
& $67.33$ & $85.19$ & $75.02$ & $89.54$ \\
& FT & All
& $76.06$ & $94.54$ & $86.46$ & $97.81$ \\
\hline

\multirow{3}{*}{\textbf{ADNI}}
& No FT & UKB
& $69.50_{\pm 0.35}$ & $87.45_{\pm 0.24}$
& $77.54_{\pm 0.31}$ & $91.17_{\pm 0.23}$ \\
& FT & ADNI
& $76.64_{\pm 3.01}$ & $94.75_{\pm 1.48}$ 
& $86.95_{\pm 2.18}$ & $97.76_{\pm 0.87}$\\
& FT & All
& $79.88_{\pm 1.11}$ & $96.32_{\pm 0.39}$
& $89.64_{\pm 0.78}$ & $98.65_{\pm 0.20}$ \\

\hline

\multirow{3}{*}{\textbf{AIBL}}
& No FT & UKB
& $68.71_{\pm 0.65}$ & $86.20_{\pm 0.48}$
& $76.64_{\pm 0.83}$ & $90.27_{\pm 0.43}$ \\
& FT & AIBL
& $77.57_{\pm 1.54}$ & $94.29_{\pm 0.69}$ 
& $85.53_{\pm 0.91}$ & $97.04_{\pm 0.35}$\\
& FT & All
& $80.01_{\pm 1.25}$ & $96.01_{\pm 0.50}$
& $89.89_{\pm 1.18}$ & $98.64_{\pm 0.30}$ \\

\hline

\multirow{3}{*}{\textbf{MIRIAD}}
& No FT & UKB
& $73.93_{\pm 1.18}$ & $90.33_{\pm 0.69}$
& $81.30_{\pm 1.01}$ & $93.32_{\pm 0.89}$ \\
& FT & MIRIAD
& $76.35_{\pm 2.43}$ & $92.19_{\pm1.22}$ 
& $83.56_{\pm 1.83}$ & $94.92_{\pm 1.10}$\\
& FT & All
& $83.49_{\pm 2.88}$ & $97.74_{\pm 0.83}$
& $93.20_{\pm 1.76}$ & $99.35_{\pm 0.59}$ \\

\hline

\multirow{3}{*}{\textbf{NIFD}}
& No FT & UKB
& $70.62_{\pm 0.59}$ & $86.78_{\pm 0.50}$
& $63.84_{\pm 3.11}$ & $80.40_{\pm 2.62}$ \\
& FT & NIFD
& $75.71_{\pm 0.39}$ & $92.39_{\pm 0.26}$ 
& $83.47_{\pm 0.41}$ & $95.10_{\pm 0.27}$\\
& FT & All
& $82.06_{\pm 0.64}$ & $96.71_{\pm 0.29}$
& $91.35_{\pm 0.37}$ & $98.91_{\pm 0.18}$ \\

\hline

\multirow{3}{*}{\textbf{PPMI}}
& No FT & UKB
& $67.18_{\pm 0.85}$ & $84.93_{\pm 0.58}$
& $74.84_{\pm 0.60}$ & $88.72_{\pm 0.34}$ \\
& FT & PPMI
& $71.56_{\pm 0.82}$ & $90.01_{\pm 0.53}$ 
& $78.69_{\pm 0.65}$ & $92.85_{\pm 0.57}$\\
& FT & All
& $79.47_{\pm 1.09}$ & $95.59_{\pm 0.56}$
& $88.71_{\pm 0.54}$ & $98.13_{\pm 0.27}$ \\

\hline
\end{tabular}
\label{tab:parcel_identification_split}
\end{table*}

\begin{table}[t]
\centering
\scriptsize
\caption{Disease-specific prior parcel IDs used for region-restricted analysis.}
\label{tab:disease_priors}
\begin{tabular}{ll}
\toprule
\textbf{Class} & \textbf{Parcel IDs} \\
\midrule
\multirow{1}{*}{AD/MCI} & \multirow{1}{*}{\begin{tabular}{l}
38, 39, 40, 41, 36, 37, 68, 69, 66, 67, 62, 63, 84, 85, 18, 19, 216, 217 \end{tabular}} \\
&\\
\multirow{2}{*}{PD} & \multirow{2}{*}{\begin{tabular}{l}
156, 157, 158, 159, 74, 75, 72, 73, 76, 77, 154, 155, 160, 161, 162, 163, 164,\\ 165, 0, 1, 14, 15, 58, 59, 172, 173 \end{tabular}} \\
&\\
&\\
\multirow{3}{*}{FTD} & \multirow{3}{*}{\begin{tabular}{l}
2, 3, 4, 5, 6, 7, 8, 9, 10, 11, 82, 83, 86, 87, 80, 81, 84, 85, 32, 33, 146, 147,\\ 148, 149, 150, 151, 24, 25, 26, 27, 28, 29, 30, 31, 20, 21, 152, 153 \end{tabular}} \\
&\\
&\\
\bottomrule
\end{tabular}
\end{table}

\subsection{Details of supervised age prediction tasks for prior works}
For age prediction, pretrained checkpoints from BrainIAC, MedicalNet, and y-Aware models are used to initialize their respective backbone architectures. The original task-specific heads are replaced with regression layers designed to output continuous age estimates, after which the models are fine-tuned on the target dataset (CN-only) for chronological age prediction. For 3D-ViT, the model is trained from scratch. Prior to training, target age values are normalized using the mean and standard deviation of the training set to stabilize optimization. The models then predict normalized age values, and Smooth L1 loss is applied between predicted and normalized targets to provide robustness against outliers while maintaining sensitivity to smaller errors. During inference, predicted normalized ages are denormalized back to the original age scale for evaluation using metrics such as MAE, RMSE, Pearson correlation, and $R^2$. This approach allows all three pretrained architectures to adapt their learned feature spaces while preserving the benefits of prior pretraining.

Table~\ref{tab:age_estimation_methods} provides the results of age prediction for BrainNorm and baselines across all datasets. The results are reported as $mean_{\pm std}$ across all folds of downstream mixed cohorts.

\subsection{Parcel Identification}
The detailed results of both Text$\rightarrow$Parcel and Parcel$\rightarrow$Text is presented in Table~\ref{tab:parcel_identification_split} with $mean_{\pm std}$ values across all 5-folds. Across all datasets, the performance improves with finetuning on more healthy scans.

\begin{figure*}[t]
  \centering
 \includegraphics[width=\linewidth]{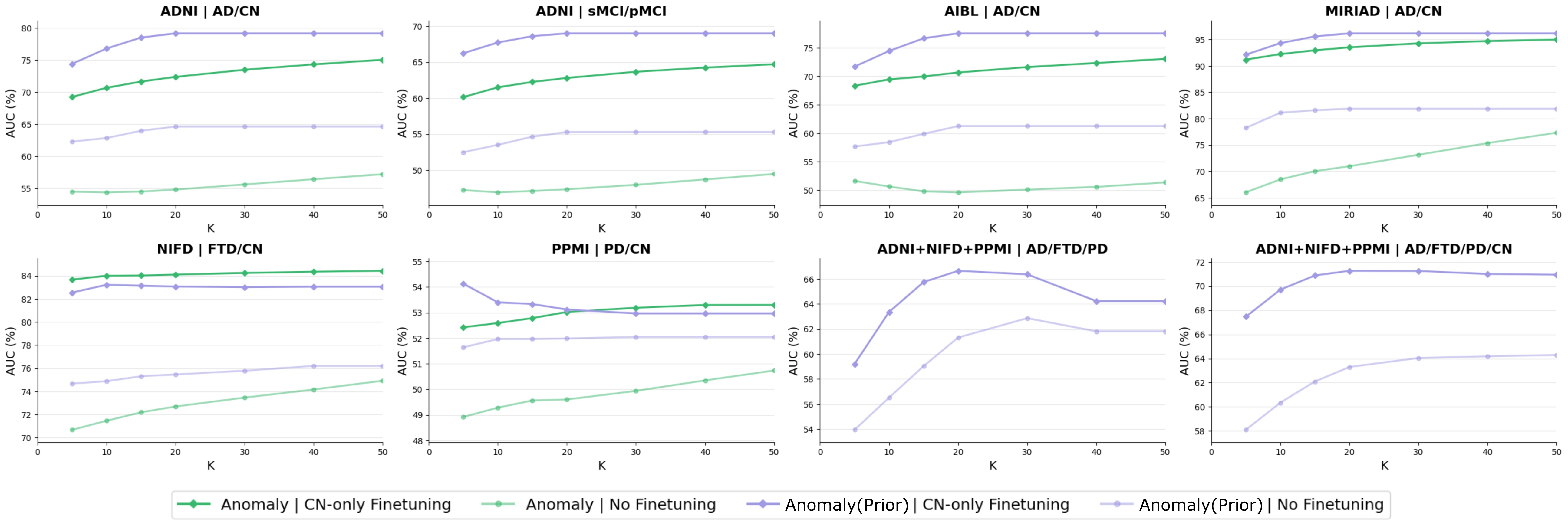}
  \caption{Zero-shot performance with increasing top-K parcels considered for both with and without fine-tuning on the full dataset.}
  \label{fig:Zero_shot}
\end{figure*}


\subsection{Zero-shot Prediction Details}
\label{Asec:zero_Shot}

\paragraph{Anomaly-based:}
The zero-shot experiments that utilized priors for anomaly prediction incorporated only those anatomical regions known to be affected according to the neurodegeneration literature~\cite{Planche2022,Braak2006,Rascovsky2011,GornoTempini2011,Hughes1992,Postuma2015}. We identified the most relevant parcels associated with each diagnosis, as listed in Table~\ref{tab:disease_priors}. These parcel IDs can be mapped to their corresponding anatomical regions using Table~\ref{tab:atlas_parcels}.

For each scan, let $\mathbf{s} = [s_1,\dots,s_P]^\top \in \mathbb{R}^{P}$,
denote the parcel-wise similarity vector between the scan embeddings and the age-matched normative template, where higher similarity indicates stronger agreement with the normative reference and lower similarity indicates greater deviation. The global anomaly score and disease-prior restricted anomaly scores are then computed (Sec~\ref{sec:classification}a) from this similarity profile using $K=20$. The results are reported in Table~\ref{tab:zero_shot_before_after}. A sweep on increasing K values are also recorded and plotted in Fig~\ref{fig:Zero_shot}

For binary zero-shot diagnosis, these anomaly scores are used directly as continuous diagnostic variables. In particular, we do not train any task-specific classifier on top of the frozen representations. Instead, the scalar anomaly score itself is treated as the ranking variable, and performance is evaluated using the area under the ROC curve (AUC). 

For multi-disease zero-shot evaluation, we compute one class-specific anomaly score for each disease using its corresponding anatomical prior subset. These scores are not summed across diseases. Rather, each disease hypothesis produces its own score, which is interpreted as class-specific diagnostic evidence. The cognitively normal class is represented using the negative global anomaly score,
\[
\ell_{\text{CN}} = -D_{\text{global}},
\]
while each disease class $d$ is represented by its own prior-restricted anomaly score,
\[
\ell_d = D_d.
\]
These class-specific scores are converted into probabilities through a softmax over class logits,
\[
P(y=d \mid x) = \frac{\exp(\ell_d)}{\sum_c \exp(\ell_c)}.
\]
Thus, multi-disease zero-shot prediction is based on relative comparison across disease-specific anomaly scores rather than on any learned classifier.

Overall, this protocol evaluates whether BrainNorm's frozen age-conditioned normative representation, together with anatomically grounded disease priors, is sufficient to separate controls from disease and to distinguish among multiple disease classes without explicit supervised training.
Detailed results of zero-shot prediction for all methods across all tasks and datasets are provided in Table~\ref {tab:zero_shot_before_after}.

\paragraph{BAG-based:}
BrainNorm also enables zero-shot disease prediction through Brain Age Gap (BAG). 
For each scan $x_i$ with chronological age $y_i$, we first obtain the retrieval-based predicted age $\hat{y}_i$ from the age-conditioned normative template space and compute:
\begin{equation}
\mathrm{BAG}_i = \hat{y}_i - y_i .
\end{equation}
We use $\mathrm{BAG}_i$ directly as a disease-agnostic anomaly score, where larger positive values indicate that the brain appears older than expected under the learned normative ageing trajectory. 
For CN-vs-disease prediction, a scan is classified as abnormal when:
\begin{equation}
\mathrm{BAG}_i \geq \gamma .
\end{equation}
We report threshold-free AUROC performance for all tasks. The values are reported as $mean_{\pm std}$ across all 5-folds in Table~\ref{tab:zero_shot_before_after}. Similarly for all baselines works, we used the same finetuned checkpoint for age prediction to calculate $\mathrm{BAG}$ for comparison.


\subsection{Traditional Normative Model baseline comparison}

BrainNorm characterizes abnormality as deviation from parcel-specific, age-conditioned healthy reference templates. We assess the abnormalities captured by BrainNorm in two complementary settings:
\begin{enumerate}
    \item \textbf{subject-level anomaly scores}, evaluated through zero-shot disease classification (Sec.~4.5a); and
    \item \textbf{parcel-level deviation representations}, evaluated by linearly probing the frozen representations (Sec.~4.5b, BrainNorm\_M2).
\end{enumerate}

We compare both settings against Hierarchical Bayesian Regression (HBR), using conventional morphometric features extracted with FreeSurfer, including regional volume, cortical surface area, and cortical thickness, following the traditional normative-modeling formulation of Rutherford et al.~\cite{rutherford2022normative}. HBR is pretrained on UK Biobank and evaluated across AD/CN and sMCI/pMCI from ADNI, AD/CN from AIBL \& MIRIAD, PD/CN from PPMI, FTD/CN from NIFD.

For zero-shot disease classification, we consider: (i) Direct transfer of the UKB-trained normative model to the target cohort; and (ii) CN-only target-cohort adaptation using only healthy control subjects, without using disease labels.
For linear probing without fine-tuning, we probe BrainNorm's frozen parcel-deviation representations (BrainNorm\_M2) and compare them against a linear probe trained on HBR's concatenated regional morphometric deviation scores. All results are reported as the mean over 5-fold cross-validation.

\begin{table*}[t]
\centering
\scriptsize
\caption{\textbf{Zero-shot disease classification using subject-level anomaly scores.}
Comparison between HBR and BrainNorm under direct transfer from UKB and target-cohort adaptation using only healthy controls. Results are reported as AUC (\%). Best and second-best results for each task may be highlighted in bold and italics, respectively.}
\label{tab:hbr_brainnorm_zeroshot}
\begin{tabular}{lcccc}
\toprule
\textbf{Task} &
\textbf{HBR} &
\textbf{HBR} &
\textbf{BrainNorm} &
\textbf{BrainNorm} \\
&
\textbf{Direct Transfer} &
\textbf{CN-only Adaptation} &
\textbf{Direct Transfer} &
\textbf{CN-only Fine-tuning} \\
\midrule

AD/CN [ADNI]      & \textbf{75.20} & 72.01 & 54.78 & \textit{72.37} \\
AD/CN [MIRIAD] & \textit{90.81} & 88.97 & 71.00 & \textbf{93.51} \\
AD/CN [AIBL]  & \textbf{71.04} & \textit{72.51} & 49.55 & 70.70 \\
sMCI/pMCI [ADNI]  & 58.81 & \textit{59.89} & 47.33 & \textbf{62.79} \\
PD/CN [PPMI]      & \textit{51.86} & 49.96 & 49.60 & \textbf{53.02} \\
FTD/CN [NIFD]     & 67.85 & \textit{75.21} & 72.69 & \textbf{84.09} \\

\bottomrule
\end{tabular}%
\end{table*}

\begin{table*}[t]
\centering
\scriptsize
\caption{\textbf{Linear probing of frozen normative-deviation representations.}
No CN-only adaptation or downstream representation fine-tuning is performed. BrainNorm$_{\mathrm{M2}}$ uses frozen parcel-deviation representations, while HBR uses concatenated regional morphometric deviation scores. $\Delta$ denotes BrainNorm minus HBR. Results are reported in \%.}
\label{tab:hbr_brainnorm_linearprobe}
\begin{tabular}{lcccccc}
\toprule
\textbf{Task} &
\textbf{HBR BAcc} &
\textbf{HBR AUC} &
\textbf{BrainNorm BAcc} &
\textbf{BrainNorm AUC} &
\textbf{$\Delta$ BAcc} &
\textbf{$\Delta$ AUC} \\
\midrule

AD/CN [ADNI]      & 82.29 & 90.25 & \textbf{86.85} & \textbf{94.27} & +4.56 & +4.02 \\
AD/CN [AIBL]  & 83.53 & 91.06 & \textbf{85.91} & \textbf{93.39} & +2.38 & +2.33 \\
AD/CN [MIRIAD] & 88.47 & 94.57 & \textbf{92.72} & \textbf{98.26} & +4.25 & +3.69 \\
sMCI/pMCI [ADNI]  & 67.47 & 74.55 & \textbf{71.96} & \textbf{80.75} & +4.49 & +6.20 \\
PD/CN  [PPMI]     & 55.87 & 60.42 & \textbf{59.08} & \textbf{66.12} & +3.21 & +5.70 \\
FTD/CN [NIFD]     & 70.14 & 75.42 & \textbf{81.82} & \textbf{89.15} & +11.68 & +13.73 \\
AD/PD/FTD   & 80.85 & 92.86 & \textbf{83.98} & \textbf{97.01} & +3.13 & +4.15 \\
AD/PD/FTD/CN & 68.69 & 86.64 & \textbf{70.32} & \textbf{91.91} & +1.63 & +5.27 \\

\bottomrule
\end{tabular}%
\end{table*}

\begin{figure*}[t]
  \centering
 \includegraphics[width=\linewidth]{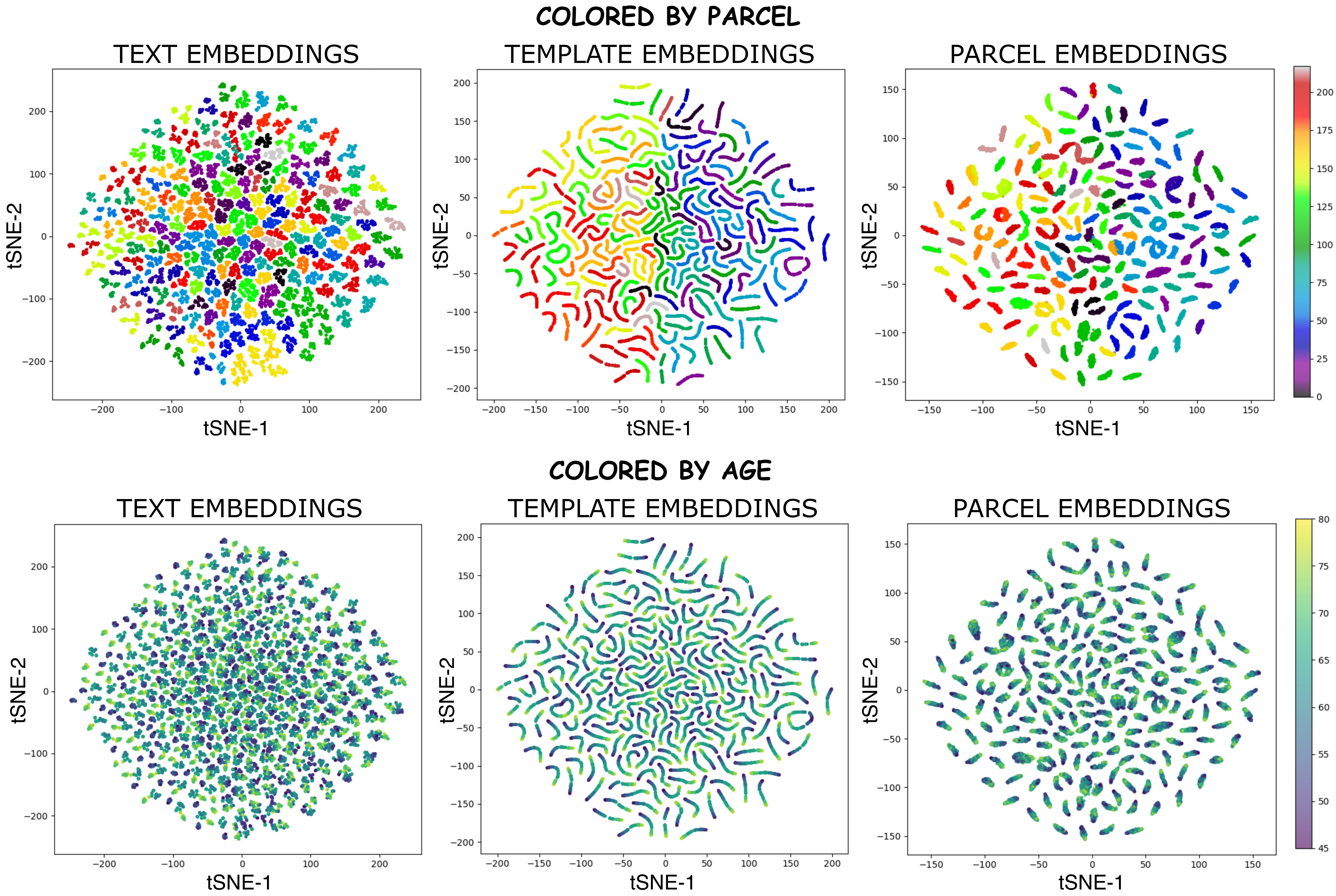}
  \vspace{-1.5em}
  \caption{t-SNE visualizations of text priors ($\mathbf{e}^{\mathrm{raw}}_{a,p}$), learned templates ($\mathbf{t}_{a,p}$), and parcel embeddings ($\mathbf{p}_{i,p}$) on held-out UKB~\cite{Sudlow2015UKBiobank} test data. The learned SAL space exhibits visible anatomical clustering and ribbon-like derived structure of age-ordering within each parcel.}
  \label{fig:pt_embed}
\end{figure*}

\begin{figure*}[t]
  \centering
 \includegraphics[width=0.8\linewidth]{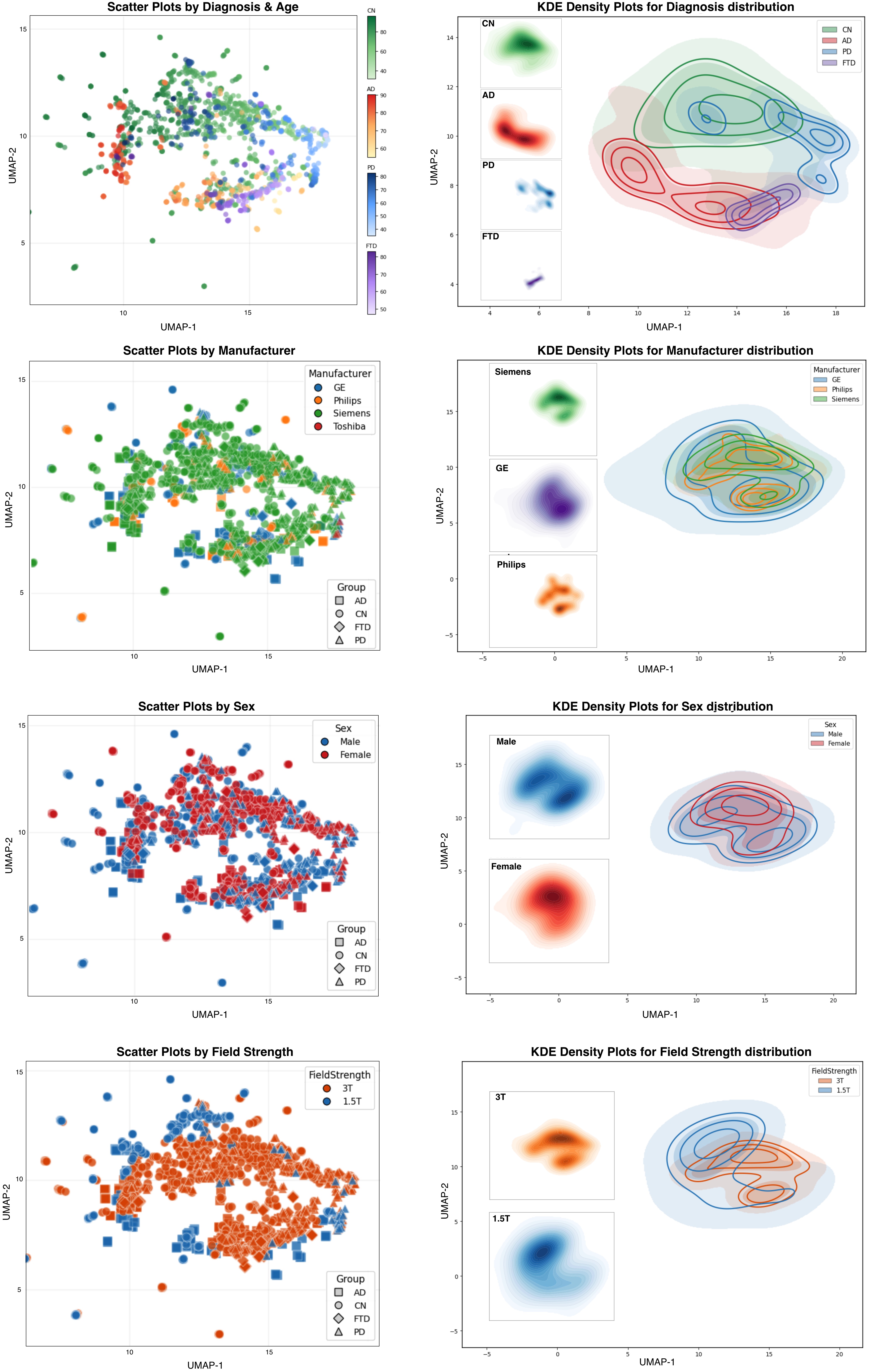}
  \caption{UMAP scatter plots and KDE density maps visualize the distribution of scan-level residual embeddings across neurodegenerative cohorts. The diagnosis-based scatter plot is colored by clinical group with an age gradient, revealing an apparent age-related organization in the embedding space, where relatively younger scans tend to lie toward the right and older scans toward the left. The diagnosis-specific KDE maps further show structured group-wise density patterns, suggesting that the residual embedding space captures disease-associated deviations from the normative representation. In contrast, the manufacturer-, sex-, and field-strength-based scatter plots and KDE maps show that scans from different vendors, sex categories, and field strengths are broadly dispersed across the embedding space rather than forming distinct confounder-specific clusters. This suggests that the observed diagnostic organization is unlikely to be primarily driven by scanner manufacturer, sex, or field-strength effects, supporting the interpretation that the learned SAL space captures biologically meaningful anomaly deviations with limited evidence of these confounding effects.} 
  \label{fig:umap}
\end{figure*}

\paragraph{Inference.}
These results reveal important distinctions between the two normative-modelling pipelines. 
\begin{itemize}
    \item HBR provides a strong classical normative baseline and, under direct cross-cohort transfer, achieves higher subject-level anomaly-score AUC than unadapted BrainNorm on several tasks. Following healthy-only target-cohort fine-tuning, however, BrainNorm improves substantially and achieves stronger anomaly-score performance on four of the six evaluated tasks. We therefore do not position BrainNorm as superior in zero-shot AUC. Rather, BrainNorm provides a 3D MRI foundation-model formulation in which disease-label-free inference can be performed from deviations relative to learned parcel-wise, age-conditioned healthy reference representations.
    \item HBR changes comparatively little following healthy-only target-cohort adaptation, whereas BrainNorm improves markedly on AD/CN, sMCI/pMCI, and FTD/CN. This suggests that healthy-only fine-tuning can align BrainNorm's learned parcel-wise healthy references with the target cohort, thereby improving the quantification of deviations without requiring disease supervision. In contrast, HBR operates on standardized scalar morphometric phenotypes and exhibits more limited gains from cohort adaptation in these experiments.
    \item BrainNorm further represents abnormality through high-dimensional parcel-deviation representations rather than solely through scalar shifts in individual morphometric measures. Consequently, a simple global aggregation of parcel abnormalities may not fully exploit the information encoded in these representations. The consistently stronger linear-probe performance indicates that BrainNorm's frozen parcel-deviation representations retain more discriminative disease-related information for the evaluated downstream tasks than HBR's concatenated scalar morphometric deviations.
    \item Together with the strong performance obtained by linearly probing BrainNorm's frozen representations without CN-only fine-tuning, these results indicate that disease-relevant anomaly cues are already encoded in the parcel-deviation representations before target-cohort adaptation, even when they are not fully reflected by a simple aggregated subject-level anomaly score. 
\end{itemize}

BrainNorm should therefore be assessed through both its subject-level anomaly scoring and its parcel-level deviation representations, which capture complementary aspects of the learned normative space.

\subsection{Formulation of Linear Probing}
\label{sec:supp_zero_shot_linear_probe}





To test whether disease-related signals are linearly extractable from the frozen representation, we train lightweight supervised heads on top of three feature constructions.

\noindent
\textit{(M1) Parcel Embedding.}
To test whether anatomical parcel representations alone are discriminative, we concatenate parcel embeddings:
$\mathbf{x}_{M1} = concat(\mathbf{p}_1,\dots,\mathbf{p}_N) \in \mathbb{R}^{ND}.$

\noindent
\textit{(M2) Parcel Deviation.}
To capture normative deviation patterns, we compute parcel-wise deviations between scan embeddings and age-matched templates, forming:
$\mathbf{x}_{M2}=concat(\mathbf{p}_1-\mathbf{t}_{a,1},\dots,\mathbf{p}_N-\mathbf{t}_{a,N}) \in \mathbb{R}^{ND}$.

\noindent
\textit{(M3) Parcel similarity vector.}
We directly use the age-matched parcel-template cosine similarities
$
\mathbf{x}_{M3}
=
[s_1,\dots,s_P]^\top
\in \mathbb{R}^{P},
$
where $s=\mathbf{p}_{a,p}\cdot\mathbf{t}_{a,p}$. 

\paragraph{Classifier architecture.}
Each feature vector $\mathbf{x}$ is fed to a single-layer MLP. For binary tasks, the MLP outputs logits for two classes; for multiclass tasks, it outputs $C$ logits followed by softmax at evaluation time.
Detailed results of linear-probing prediction and few-shot prediction for all three methods across all tasks and datasets are provided in Table~\ref {tab:fulltrain_linearprobe_binary_admci_baselines_brainnorm},~\ref {tab:fulltrain_linearprobe_binary_pdftd_baselines_brainnorm},~\ref {tab:fulltrain_linearprobe_multi_baselines_brainnorm}, and Table~\ref{tab:fewshot_brainnorm_binary},~\ref{tab:fewshot_brainnorm_multiclass} respectively.

\subsection{BrainNorm's embeddings visualizations}
\label{Asec:UMAP}

Fig.~\ref{fig:pt_embed} visualizes the learned Semantic Atlas Latent (SAL) space, where BrainNorm's template and parcel embeddings organize according to both anatomical identity and age. Embeddings from the same parcel form coherent clusters, while within each cluster, age-conditioned embeddings follow a smooth ordering. This suggests that the learned space preserves parcel-specific identity while capturing normative age-related variation.

The Fig~\ref{fig:umap} visualizes the two-dimensional UMAP projection of residual whole-brain embeddings computed after subtracting age-matched normative template representations. The scatter plot shows individual subjects, with colours indicating diagnostic groups and age gradients within each group. The inset density panels provide group-specific density structure, while the combined density map shows the relative spatial organization and overlap between diagnostic groups.
The residual embedding visualization shows that age-matched normative deviations organize subjects into clusters while also preserving an apparent age-related ordering within the embedding space. The plots suggests that CN subjects largely occupy the upper residual space, whereas AD shifts toward the lower-left/lower-central region, PD extends toward the upper-right with broader dispersion, and FTD forms a more compact lower-right cluster. The density plots further summarize these class-wise regions, showing high-density zones with some overlap between disease groups. This indicates that the learned residual embeddings capture both diagnostic deviation patterns and age-dependent variation, suggesting that the model retains meaningful information about normal aging while also separating disease-associated departures from the normative trajectory. More importantly, the effect of confounders analysis suggests that the embeddings organization is unlikely to be driven by variations in manufacturer/ sex/ field strength.

\subsection{Scan-level Interpretability}
\label{Asec:scan_interpret}

\begin{figure*}[t]
  \centering
 \includegraphics[width=\linewidth]{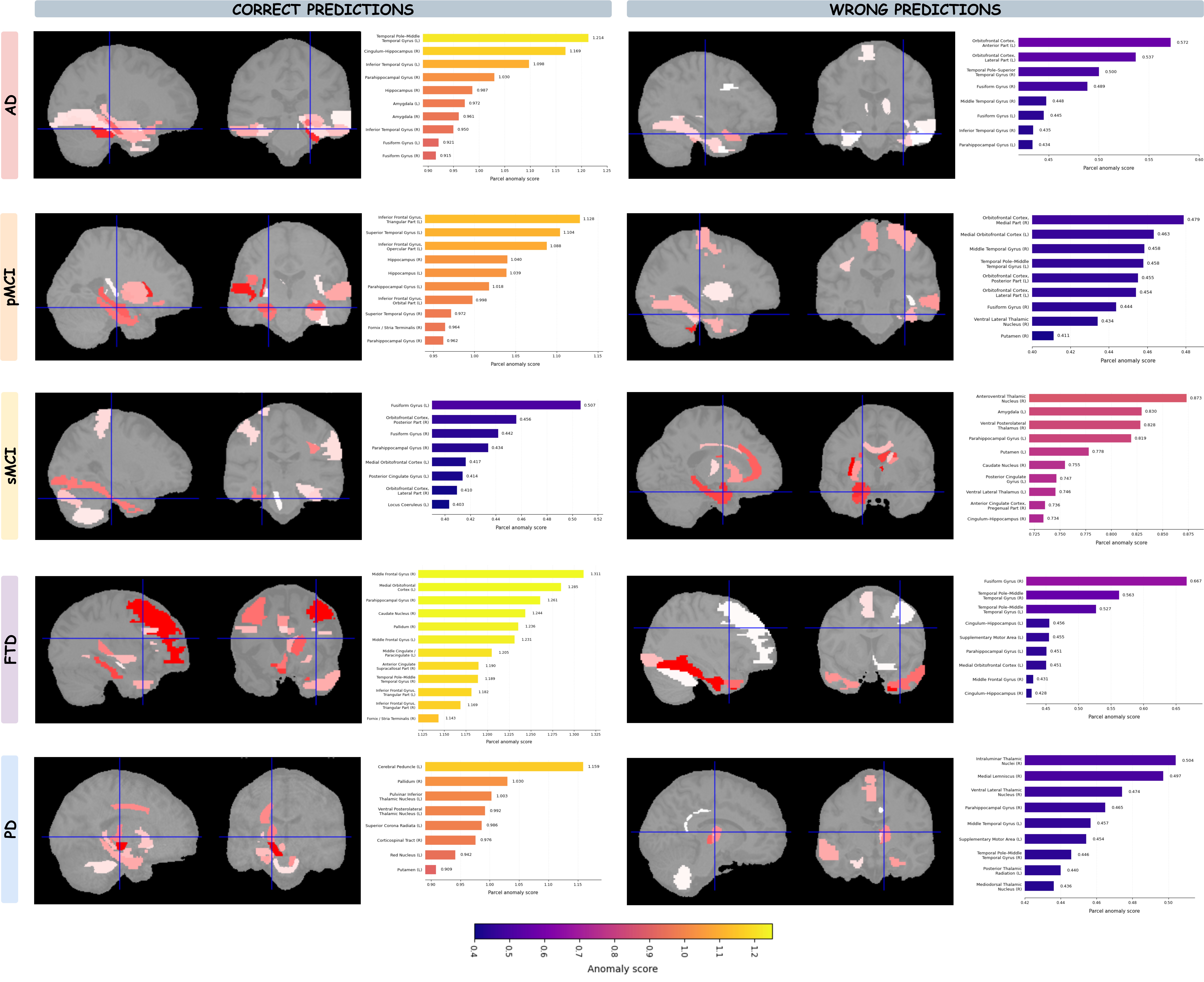}
  \caption{\textbf{Scan-level interpretability across neurodegenerative disease.} Representative correct and incorrect predictions are shown for AD, pMCI, sMCI, FTD, and PD. Correct predictions correspond to AD, pMCI, sMCI, FTD, and PD subjects assigned to their respective ground-truth classes, whereas wrong predictions correspond to AD, FTD, and PD subjects misclassified as CN, pMCI subjects misclassified as sMCI, and sMCI subjects misclassified as pMCI. For each subject, the spatial map shows the highest-ranking parcel-wise deviations from the age-matched normative representation, with the image colors normalized independently within that subject to emphasize the spatial distribution of its regional abnormalities. The adjacent bar plot reports the corresponding raw parcel anomaly scores, with bar colors mapped using a common anomaly-score scale across all scans to enable direct comparison of deviation magnitude between subjects. Across correct predictions, the highlighted parcels show strong disease-relevant deviations. In contrast, wrong predictions generally retain anatomically relevant regional patterns but exhibit anomaly magnitudes that are shifted in the direction of the predicted class: AD, FTD, PD, and pMCI false negatives show comparatively weaker deviations, whereas sMCI subjects misclassified as pMCI show comparatively stronger deviations.}
  \label{fig:scan_interpret}
\end{figure*}

To examine whether the learned normative representation provides interpretable evidence at the individual-subject level, we analyze representative correct and incorrect predictions for AD, FTD, PD, sMCI, and pMCI. For each scan, parcel-wise anomaly is quantified as the deviation from its age-matched normative representation, and the highest-ranking disease-relevant parcels are projected back onto the image. The corresponding parcel anomaly scores are shown using a shared color scale across subjects, such that both the spatial localization and absolute magnitude of normative deviation can be compared directly (Fig.~\ref{fig:scan_interpret}).

Correct predictions exhibit distinct and anatomically coherent deviation patterns across diseases. Correctly identified AD demonstrates prominent temporal and limbic abnormalities, including the temporal pole, hippocampus, parahippocampal gyrus, amygdala, inferior temporal cortex, and fusiform gyri. Similarly, the correctly classified pMCI subject exhibits strong hippocampal, parahippocampal, and temporal-related deviations, consistent with a more pronounced progression-like pattern. In contrast, the correctly identified sMCI subject shows substantially weaker regional deviations. FTD is characterized by strong frontal, orbitofrontal, cingulate, temporal, and striatal deviations, while the correctly identified PD subject shows a more motor-system-dominant pattern involving the cerebral peduncle, pallidum, putamen, thalamic nuclei, corona radiata, corticospinal tract, and red nucleus. Thus, the normative anomaly representation does not collapse different disorders onto a common abnormality score, but preserves heterogeneous regional patterns at the individual-scan level.

The failure cases further clarify the model's decision mechanism. AD and FTD subjects predicted as CN retain partially disease-consistent temporal or frontotemporal abnormalities, respectively, but their anomaly magnitudes are markedly reduced relative to the correctly classified counterparts. A similar behavior is observed for PD, where the false-negative scan contains thalamic and motor-related abnormalities but lacks the stronger and more complete basal-ganglia–motor pathway pattern observed in the correct prediction. The MCI errors show the complementary behavior: a pMCI subject classified as sMCI exhibits only weak frontal-temporal deviations, whereas an sMCI subject classified as pMCI shows substantially elevated limbic, cingulate, and thalamic abnormalities. These examples suggest that errors arise not because the model attends to anatomically unrelated regions, but because the strength and completeness of disease-relevant normative deviations place the subject on the opposite side of the decision boundary.

Together, these results demonstrate that the learned parcel-wise normative space supports subject-specific interpretation at two complementary levels: it localizes where an individual deviates from the expected healthy representation and quantifies how strongly those regions deviate. This provides an interpretable explanation for both successful and failed predictions. Future work can explore reducing these wrong predictions by improving the alignment of the learned normative representation, enabling more accurate capture of disease-relevant deviations at the scan and parcel levels.

\begin{table*}[h!]
\scriptsize
\centering
\caption{BrainNorm's Age estimation on held-out test sets and downstream datasets using only CN subjects vs Prior works age prediction with full supervision. FT denotes finetuning on the CN subset of the corresponding dataset or pooled CN data from ADNI+AIBL+MIRIAD+NIFD+PPMI (All).  Values are reported as mean$_{\pm \mathrm{std}}$ over 5 folds where available.}
\resizebox{0.9\textwidth}{!}{
\begin{tabular}{cllccccc}
\hline
\textbf{Dataset} & \textbf{Method} & \textbf{Mode} & \textbf{CN Train/FT data}
& \textbf{MAE $\downarrow$}
& \textbf{RMSE $\downarrow$}
& \textbf{Pearson $r \uparrow$}
& \textbf{$R^2 \uparrow$} \\
\hline

\multirow{5}{*}{\textbf{UKB}}
& 3D-ViT & Full Train & UKB & $4.23$ & $5.65$ & $0.77$ & $0.58$ \\
& BrainIAC & Full FT & UKB & $6.76$ & $8.30$ & $0.16$ & $0.16$ \\
& BrainMVP & Full FT & UKB & $4.15$ & $\underline{5.21}$ & $0.83$ & $\mathbf{0.79}$ \\
& BrainHarmony & Full FT & UKB & $\underline{4.15}$ & $\mathbf{5.20}$ & $0.83$ & $0.76$ \\
& RadFM & Full FT & UKB & $4.83$ & $5.62$ & $0.82$ & $\underline{0.77}$ \\
& NeuroVFM & Full FT & UKB & $4.25$ & $5.11$ & $0.81$ & $0.72$ \\
& Neuro-JEPA & Full FT & UKB & $4.64$ & $5.58$ & $0.81$ & $0.62$ \\
& MedicalNet & Full FT & UKB & $4.16$ & $5.27$ & $0.82$ & $0.53$ \\
& y-Aware & Full FT & UKB & $4.17$ & $5.23$ & $0.82$ & $0.54$ \\
\cdashline{2-8}
& BrainNorm & No FT & UKB & $\mathbf{4.14}$ & $5.22$ & $\mathbf{0.85}$ & $0.54$ \\
& BrainNorm & CN-only FT & All & $4.69$ & $5.77$ & $\underline{0.84}$ & $0.44$ \\
\hline

\multirow{5}{*}{\textbf{MCSA}}
& 3D-ViT & Infer & UKB & $9.06$ & $10.68$ & $0.68$ & $0.02$ \\
& BrainIAC & Infer & UKB & $18.49$ & $15.83$ & $0.04$ & $0.01$ \\
& BrainMVP & Infer & UKB & $4.70$ & $5.77$ & $0.88$ & $0.66$ \\
& BrainHarmony & Infer & UKB & $5.05$ & $6.16$ & $0.86$ & $0.61$ \\
& RadFM & Infer & UKB & $5.55$ & $6.81$ & $0.86$ & $0.52$ \\
& NeuroVFM & Infer & UKB & $5.93$ & $7.21$ & $0.78$ & $0.46$ \\
& Neuro-JEPA & Infer & UKB & $6.24$ & $7.46$ & $0.85$ & $0.43$ \\
& MedicalNet & Infer & UKB & $8.75$ & $10.42$ & $0.56$ & $0.10$ \\
& y-Aware & Infer & UKB & $5.43$ & $6.78$ & $0.80$ & $0.53$ \\
\cdashline{2-8}
& BrainNorm & No FT & UKB & $\underline{3.94}$ & $\underline{4.90}$ & $\underline{0.88}$ & $\underline{0.75}$ \\
& BrainNorm & CN-only FT & All & $\mathbf{3.66}$ & $\mathbf{4.61}$ & $\mathbf{0.89}$ & $\mathbf{0.78}$ \\
\hline

\multirow{6}{*}{\textbf{ADNI}}
& 3D-ViT & Full Train & ADNI & $4.48_{\pm 0.16}$ & $5.65_{\pm 0.23}$ & $0.62_{\pm 0.03}$ & $0.38_{\pm 0.03}$ \\
& BrainIAC & Full FT & ADNI & $5.60_{\pm 0.06}$ & $7.11_{\pm 0.12}$ & $0.01_{\pm 0.13}$ & $0.00_{\pm 0.01}$ \\
& BrainMVP & Full FT & ADNI & $4.46_{\pm 0.13}$ & $5.70_{\pm 0.16}$ & $0.66_{\pm 0.04}$ & $0.37_{\pm 0.04}$ \\
& BrainHarmony & Full FT & ADNI & $4.38_{\pm 0.17}$ & $5.52_{\pm 0.17}$ & $0.64_{\pm 0.04}$ & $0.40_{\pm 0.05}$ \\
& RadFM & Full FT & ADNI & $3.87_{\pm 0.24}$ & $4.71_{\pm 0.22}$ & $0.74_{\pm 0.01}$ & $0.60_{\pm 0.02}$ \\
& NeuroVFM & Full FT & ADNI & $3.85_{\pm 0.03}$ & $4.94_{\pm 0.03}$ & $0.73_{\pm 0.00}$ & $0.52_{\pm 0.01}$ \\
& Neuro-JEPA & Full FT & ADNI & $3.98_{\pm 0.06}$ & $4.98_{\pm 0.09}$ & $0.74_{\pm 0.01}$ & $\underline{0.60_{\pm 0.01}}$ \\
& MedicalNet & Full FT & ADNI & $4.38_{\pm 0.09}$ & $5.54_{\pm 0.17}$ & $0.64_{\pm 0.01}$ & $0.40_{\pm 0.02}$ \\
& y-Aware & Full FT & ADNI & $4.14_{\pm 0.13}$ & $5.25_{\pm 0.16}$ & $0.73_{\pm 0.02}$ & $0.46_{\pm 0.02}$ \\
\cdashline{2-8}
& BrainNorm & No FT & UKB & $\mathbf{3.56_{\pm 0.13}}$ & $\mathbf{4.49_{\pm 0.19}}$ & $0.78_{\pm 0.03}$ & $\mathbf{0.61_{\pm 0.05}}$ \\
& BrainNorm & CN-only FT & ADNI & $3.88_{\pm 0.29}$ & $4.81_{\pm 0.41}$ & $\mathbf{0.80_{\pm 0.03}}$ & $0.54_{\pm 0.10}$ \\
& BrainNorm & CN-only FT & All & $\underline{3.72_{\pm 0.20}}$ & $\underline{4.66_{\pm 0.30}}$ & $\underline{0.80_{\pm 0.03}}$ & $0.57_{\pm 0.07}$ \\

\hline

\multirow{6}{*}{\textbf{AIBL}}
& 3D-ViT & Full Train & AIBL & $5.24_{\pm 0.37}$ & $6.44_{\pm 0.54}$ & $0.19_{\pm 0.13}$ & $0.01_{\pm 0.01}$ \\
& BrainIAC & Full FT & AIBL & $5.23_{\pm 0.38}$ & $6.43_{\pm 0.54}$ & $0.02_{\pm 0.06}$ & $0.01_{\pm 0.01}$ \\
& BrainMVP & Full FT & AIBL & $5.23_{\pm 0.07}$ & $6.44_{\pm 0.08}$ & $0.31_{\pm 0.01}$ & $0.01_{\pm 0.03}$ \\
& BrainHarmony & Full FT & AIBL & $5.10_{\pm 0.38}$ & $6.26_{\pm 0.56}$ & $0.27_{\pm 0.10}$ & $0.04_{\pm 0.02}$ \\
& RadFM & Full FT & AIBL & $3.96_{\pm 0.44}$ & $4.94_{\pm 0.48}$ & $0.70_{\pm 0.08}$ & $0.50_{\pm 0.11}$ \\
& NeuroVFM & Full FT & AIBL & $5.04_{\pm 0.06}$ & $6.20_{\pm 0.06}$ & $0.41_{\pm 0.01}$ & $0.07_{\pm 0.02}$ \\
& Neuro-JEPA & Full FT & AIBL & $4.99_{\pm 0.13}$ & $6.11_{\pm 0.16}$ & $0.41_{\pm 0.07}$ & $0.10_{\pm 0.05}$ \\
& MedicalNet & Full FT & AIBL & $4.62_{\pm 0.33}$ & $5.76_{\pm 0.41}$ & $0.45_{\pm 0.07}$ & $0.19_{\pm 0.07}$ \\
& y-Aware & Full FT & AIBL & $4.66_{\pm 0.15}$ & $5.68_{\pm 0.21}$ & $0.48_{\pm 0.08}$ & $0.21_{\pm 0.09}$ \\
\cdashline{2-8}
& BrainNorm & No FT & UKB & $\underline{3.50_{\pm 0.26}}$ & $\underline{4.42_{\pm 0.45}}$ & $\underline{0.74_{\pm 0.02}}$ & $\underline{0.52_{\pm 0.04}}$ \\
& BrainNorm & CN-only FT & AIBL & $\mathbf{3.39_{\pm 0.26}}$ & $\mathbf{4.31_{\pm 0.45}}$ & $\mathbf{0.76_{\pm 0.03}}$ & $\mathbf{0.55_{\pm 0.05}}$ \\
& BrainNorm & CN-only FT & All & $3.72_{\pm 0.20}$ & $4.67_{\pm 0.35}$ & $0.76_{\pm 0.02}$ & $0.47_{\pm 0.05}$ \\

\hline

\multirow{6}{*}{\textbf{MIRIAD}}
& 3D-ViT & Full Train & MIRIAD & $8.15_{\pm 24.43}$ & $9.03_{\pm 24.04}$ & $0.09_{\pm 0.75}$ & $0.01_{\pm 4.18}$ \\
& BrainIAC & Full FT & MIRIAD & $5.03_{\pm 1.28}$ & $6.48_{\pm 1.83}$ & $0.12_{\pm 0.59}$ & $0.08_{\pm 0.33}$ \\
& BrainMVP & Full FT & MIRIAD & $5.24_{\pm 0.04}$ & $7.05_{\pm 0.10}$ & $0.10_{\pm 0.09}$ & $0.01_{\pm 0.03}$ \\
& BrainHarmony & Full FT & MIRIAD & $5.20_{\pm 1.32}$ & $6.67_{\pm 1.73}$ & $0.14_{\pm 0.23}$ & $0.01_{\pm 0.24}$ \\
& RadFM & Full FT & MIRIAD & $4.49_{\pm 1.07}$ & $5.71_{\pm 1.37}$ & $0.70_{\pm 0.08}$ & $0.40_{\pm 0.29}$ \\
& NeuroVFM & Full FT & MIRIAD & $5.18_{\pm 0.17}$ & $6.87_{\pm 0.23}$ & $0.72_{\pm 0.10}$ & $0.01_{\pm 0.07}$ \\
& Neuro-JEPA & Full FT & MIRIAD & $5.32_{\pm 0.05}$ & $6.92_{\pm 0.07}$ & $0.13_{\pm 0.14}$ & $0.02_{\pm 0.02}$ \\
& MedicalNet & Full FT & MIRIAD & $4.74_{\pm 1.36}$ & $5.84_{\pm 1.84}$ & $0.23_{\pm 0.57}$ & $0.13_{\pm 0.29}$ \\
& y-Aware & Full FT & MIRIAD & $4.31_{\pm 1.06}$ & $5.39_{\pm 1.15}$ & $0.56_{\pm 0.33}$ & $0.20_{\pm 0.43}$ \\
\cdashline{2-8}
& BrainNorm & No FT & UKB & $4.83_{\pm 1.30}$ & $5.31_{\pm 1.42}$ & $\mathbf{0.82_{\pm 0.17}}$ & $0.29_{\pm 0.40}$ \\
& BrainNorm & CN-only FT & MIRIAD & $\underline{3.46_{\pm 1.56}}$ & $\underline{3.89_{\pm 1.61}}$ & $\underline{0.80_{\pm 0.19}}$ & $\underline{0.52_{\pm 0.30}}$ \\
& BrainNorm & CN-only FT & All & $\mathbf{3.23_{\pm 1.18}}$ & $\mathbf{3.78_{\pm 1.27}}$ & $0.78_{\pm 0.24}$ & $\mathbf{0.52_{\pm 0.28}}$ \\

\hline

\multirow{6}{*}{\textbf{NIFD}}
& 3D-ViT & Full Train & NIFD & $6.53_{\pm 0.60}$ & $8.08_{\pm 0.98}$ & $0.14_{\pm 0.13}$ & $0.01_{\pm 0.04}$ \\
& BrainIAC & Full FT & NIFD & $5.97_{\pm 0.72}$ & $7.71_{\pm 0.87}$ & $0.16_{\pm 0.13}$ & $0.01_{\pm 0.02}$ \\
& BrainMVP & Full FT & NIFD & $6.12_{\pm 0.05}$ & $7.77_{\pm 0.01}$ & $0.05_{\pm 0.03}$ & $0.01_{\pm 0.02}$ \\
& BrainHarmony & Full FT & NIFD & $5.97_{\pm 0.71}$ & $7.72_{\pm 0.89}$ & $0.25_{\pm 0.15}$ & $0.01_{\pm 0.01}$ \\
& RadFM & Full FT & NIFD & $5.83_{\pm 0.77}$ & $6.95_{\pm 0.83}$ & $0.56_{\pm 0.18}$ & $0.18_{\pm 0.18}$ \\
& NeuroVFM & Full FT & NIFD & $5.98_{\pm 0.02}$ & $7.71_{\pm 0.07}$ & $0.56_{\pm 0.02}$ & $0.01_{\pm 0.02}$ \\
& Neuro-JEPA & Full FT & NIFD & $6.11_{\pm 0.06}$ & $7.82_{\pm 0.07}$ & $0.03_{\pm 0.07}$ & $0.02_{\pm 0.02}$ \\
& MedicalNet & Full FT & NIFD & $6.01_{\pm 0.64}$ & $7.76_{\pm 0.87}$ & $0.15_{\pm 0.09}$ & $0.01_{\pm 0.01}$ \\
& y-Aware & Full FT & NIFD & $5.88_{\pm 0.75}$ & $7.67_{\pm 0.91}$ & $0.22_{\pm 0.09}$ & $0.02_{\pm 0.02}$ \\
\cdashline{2-8}
& BrainNorm & No FT & UKB & $5.44_{\pm 0.42}$ & $6.75_{\pm 0.62}$ & $\underline{0.76_{\pm 0.08}}$ & $0.20_{\pm 0.25}$ \\
& BrainNorm & CN-only FT & NIFD & $\mathbf{4.33_{\pm 0.40}}$ & $\underline{5.51_{\pm 0.56}}$ & $\mathbf{0.77_{\pm 0.08}}$ & $\underline{0.47_{\pm 0.15}}$ \\
& BrainNorm & CN-only FT & All & $\underline{4.35_{\pm 0.62}}$ & $\mathbf{5.44_{\pm 0.67}}$ & $0.75_{\pm 0.08}$ & $\mathbf{0.49_{\pm 0.14}}$ \\

\hline

\multirow{6}{*}{\textbf{PPMI}}
& 3D-ViT & Full Train & PPMI & $9.33_{\pm 0.91}$ & $11.93_{\pm 1.23}$ & $0.11_{\pm 0.24}$ & $0.01_{\pm 0.04}$ \\
& BrainIAC & Full FT & PPMI & $9.39_{\pm 0.68}$ & $11.92_{\pm 0.81}$ & $0.14_{\pm 0.07}$ & $0.00_{\pm 0.01}$ \\
& BrainMVP & Full FT & PPMI & $9.34_{\pm 0.06}$ & $11.63_{\pm 0.04}$ & $0.10_{\pm 0.01}$ & $0.01_{\pm 0.04}$ \\
& BrainHarmony & Full FT & PPMI & $9.18_{\pm 0.75}$ & $11.63_{\pm 0.99}$ & $0.26_{\pm 0.20}$ & $0.03_{\pm 0.01}$ \\
& RadFM & Full FT & PPMI & $6.53_{\pm 0.59}$ & $9.94_{\pm 0.87}$ & $0.76_{\pm 0.03}$ & $0.43_{\pm 0.11}$ \\
& NeuroVFM & Full FT & PPMI & $9.23_{\pm 0.24}$ & $11.72_{\pm 0.46}$ & $0.48_{\pm 0.07}$ & $0.02_{\pm 0.08}$ \\
& Neuro-JEPA & Full FT & PPMI & $9.26_{\pm 0.13}$ & $11.60_{\pm 0.20}$ & $0.13_{\pm 0.13}$ & $0.02_{\pm 0.03}$ \\
& MedicalNet & Full FT & PPMI & $9.05_{\pm 0.84}$ & $11.55_{\pm 1.09}$ & $0.18_{\pm 0.08}$ & $0.01_{\pm 0.02}$ \\
& y-Aware & Full FT & PPMI & $8.54_{\pm 0.88}$ & $11.00_{\pm 1.33}$ & $0.32_{\pm 0.14}$ & $0.13_{\pm 0.13}$ \\
\cdashline{2-8}
& BrainNorm & No FT & UKB & $7.42_{\pm 0.91}$ & $9.68_{\pm 1.07}$ & $0.83_{\pm 0.03}$ & $0.29_{\pm 0.07}$ \\
& BrainNorm & CN-only FT & PPMI & $\underline{6.33_{\pm 0.77}}$ & $\underline{8.64_{\pm 1.07}}$ & $\mathbf{0.84_{\pm 0.03}}$ & $\underline{0.44_{\pm 0.06}}$ \\
& BrainNorm & CN-only FT & All & $\mathbf{5.44_{\pm 0.55}}$ & $\mathbf{7.56_{\pm 0.94}}$ & $\underline{0.83_{\pm 0.03}}$ & $\mathbf{0.57_{\pm 0.05}}$ \\

\hline
\end{tabular}
}
\label{tab:age_estimation_methods}
\end{table*}

\begin{table*}[h!]
\centering
\setlength{\tabcolsep}{4pt}
\caption{Zero-shot disease classification performance (AUC in \%) across binary and multi-class tasks using top-k@20. Results are reported as mean$_{\pm \mathrm{std}}$. FT (own): finetuning on the corresponding dataset CN only. FT (all CN): finetuning on pooled CN data from ADNI+AIBL+MIRIAD+NIFD+PPMI. A: Anomaly. P+A: Disease Prior restricted Anomaly. BAG: Brain Age Gap}
\label{tab:zero_shot_before_after}
\resizebox{0.9\textwidth}{!}{
\begin{tabular}{llcccc}
\toprule
\textbf{Dataset/Task} & \textbf{Method} & \textbf{Full FT} & \textbf{No FT} & \textbf{FT (own)} & \textbf{FT (all CN)} \\
\midrule

\multirow{3}{*}{\textbf{ADNI} [AD/CN]}
& \textbf{A}   &-& $54.78_{\pm 4.04}$ & $72.80_{\pm 4.39}$ & $72.37_{\pm 5.07}$ \\
& \textbf{P+A} &-& $64.61_{\pm 3.00}$ & $79.27_{\pm 3.94}$ & $79.16_{\pm 3.79}$ \\
& \textbf{BAG} &-& $82.80_{\pm 1.95}$ & $86.19_{\pm 2.62}$ & $86.02_{\pm 2.12}$ \\
\cdashline{2-6}
& BrainIAC+BAG & $46.68_{\pm3.97}$ & - & -  & - \\
& BrainMVP+BAG & $57.58_{\pm2.20}$ & - & -  & - \\
& BrainHarmony+BAG & $66.65_{\pm3.78}$ & - & -  & - \\
& RadFM+BAG & $72.93_{\pm1.50}$ & - & - & - \\
& NeuroVFM+BAG & $65.00_{\pm1.00}$ & - & - & - \\
& Neuro-JEPA+BAG & $70.42_{\pm0.75}$ & - & - & - \\
& MedicalNet+BAG & $65.75_{\pm4.39}$ & - & -  & - \\
& y-aware+BAG & $70.83_{\pm3.69}$ & - & -  & - \\
\midrule

\multirow{3}{*}{\textbf{ADNI} [sMCI/pMCI]}
& \textbf{A}   &-& $47.33_{\pm4.11}$ & $62.49_{\pm 6.14}$ & $62.79_{\pm 6.55}$ \\
& \textbf{P+A} &-& $55.28_{\pm3.04}$ & $69.44_{\pm 2.62}$ & $69.01_{\pm 3.37}$ \\
& \textbf{BAG} &-& $73.59_{\pm 5.01}$ & $74.98_{\pm 4.64}$ & $74.58_{\pm 5.03}$ \\
\cdashline{2-6}
& BrainIAC+BAG & $46.85_{\pm1.63}$ & - & -  & - \\
& BrainMVP+BAG & $53.62_{\pm1.50}$ & - & -  & - \\
& BrainHarmony+BAG & $57.18_{\pm8.79}$ & - & -  & - \\
& RadFM+BAG & $63.71_{\pm0.77}$ & - & - & - \\
& NeuroVFM+BAG & $58.61_{\pm 0.53}$ & - & - & - \\
& Neuro-JEPA+BAG & $61.60_{\pm0.36}$ & - & - & - \\
& MedicalNet+BAG & $58.86_{\pm3.99}$ & - & -  & - \\
& y-aware+BAG & $61.60_{\pm5.63}$ & - & -  & - \\
\midrule

\multirow{3}{*}{\textbf{AIBL} [AD/CN]}
& \textbf{A}   &-& $49.55_{\pm 5.98}$ & $68.09_{\pm 3.42}$ & $70.70_{\pm 2.23}$ \\
& \textbf{P+A} &-& $61.24_{\pm 6.10}$ & $79.83_{\pm 1.97}$ & $77.60_{\pm 3.60}$ \\
& \textbf{BAG} &-& $87.14_{\pm 1.19}$ & $90.56_{\pm 1.89}$ & $89.97_{\pm 2.54}$ \\
\cdashline{2-6}
& BrainIAC+BAG & $42.72_{\pm5.34}$ & - & -  & - \\
& BrainMVP+BAG & $43.38_{\pm0.17}$ & - & -  & - \\
& BrainHarmony+BAG & $45.74_{\pm6.25}$ & - & -  & - \\
& RadFM+BAG & $59.91_{\pm4.23}$ & - & - & - \\
& NeuroVFM+BAG & $45.00_{\pm1.00}$ & - & - & - \\
& Neuro-JEPA+BAG & $46.03_{\pm1.31}$ & - & - & - \\
& MedicalNet+BAG & $53.10_{\pm8.24}$ & - & -  & - \\
& y-aware+BAG & $53.50_{\pm6.05}$ & - & -  & - \\
\midrule

\multirow{3}{*}{\textbf{MIRIAD} [AD/CN]}
& \textbf{A}   &-& $71.00_{\pm 8.74}$ & $80.58_{\pm 8.03}$ & $93.51_{\pm 7.58}$ \\
& \textbf{P+A} &-& $81.89_{\pm 7.23}$ & $85.91_{\pm 7.87}$ & $96.16_{\pm 3.59}$ \\
& \textbf{BAG} &-& $98.52_{\pm 2.47}$ & $98.89_{\pm 2.32}$ & $98.56_{\pm 3.22}$ \\
\cdashline{2-6}
& BrainIAC+BAG & $52.12_{\pm8.61}$ & - & -  & - \\
& BrainMVP+BAG & $50.74_{\pm0.03}$ & - & -  & - \\
& BrainHarmony+BAG & $54.32_{\pm6.76}$ & - & -  & - \\
& RadFM+BAG & $58.81_{\pm5.63}$ & - & - & - \\
& NeuroVFM+BAG & $52.00_{\pm1.00}$ & - & - & - \\
& Neuro-JEPA+BAG & $51.36_{\pm0.45}$ & - & - & - \\
& MedicalNet+BAG & $47.22_{\pm8.27}$ & - & -  & - \\
& y-aware+BAG & $73.37_{\pm8.64}$ & - & -  & - \\
\midrule

\multirow{3}{*}{\textbf{NIFD} [FTD/CN]}
& \textbf{A}   &-& $72.69_{\pm 5.95}$ & $79.89_{\pm 7.02}$ & $84.09_{\pm 8.06}$ \\
& \textbf{P+A} &-& $75.46_{\pm 6.65}$ & $78.91_{\pm 6.13}$ & $83.05_{\pm 7.68}$ \\
& \textbf{BAG} &-& $82.38_{\pm 4.13}$ & $82.80_{\pm 4.61}$ & $85.18_{\pm 3.07}$ \\
\cdashline{2-6}
& BrainIAC+BAG & $50.52_{\pm2.15}$ & - & -  & - \\
& BrainMVP+BAG & $51.08_{\pm0.03}$ & - & -  & - \\
& BrainHarmony+BAG & $51.68_{\pm2.54}$ & - & -  & - \\
& RadFM+BAG & $54.78_{\pm3.31}$ & - & - & - \\
& NeuroVFM+BAG & $52.00_{\pm0.00}$ & - & - & - \\
& Neuro-JEPA+BAG & $51.34_{\pm0.05}$ & - & - & - \\
& MedicalNet+BAG & $52.10_{\pm3.38}$ & - & -  & - \\
& y-aware+BAG & $55.73_{\pm3.31}$ & - & -  & - \\
\midrule

\multirow{3}{*}{\textbf{PPMI} [PD/CN]}
& \textbf{A}   &-& $49.60_{\pm 2.48}$ & $50.26_{\pm 6.23}$ & $53.02_{\pm 7.96}$ \\
& \textbf{P+A} &-& $51.99_{\pm 4.53}$ & $51.65_{\pm 7.30}$ & $53.12_{\pm 5.18}$ \\
& \textbf{BAG} &-& $55.03_{\pm 4.57}$ & $55.59_{\pm 5.33}$ & $56.10_{\pm 6.14}$ \\
\cdashline{2-6}
& BrainIAC+BAG & $44.90_{\pm4.30}$ & - & -  & - \\
& BrainMVP+BAG & $45.42_{\pm0.01}$ & - & -  & - \\
& BrainHarmony+BAG & $45.69_{\pm3.95}$ & - & -  & - \\
& RadFM+BAG & $49.53_{\pm1.78}$ & - & - & - \\
& NeuroVFM+BAG & $46.00_{\pm0.00}$ & - & - & - \\
& Neuro-JEPA+BAG & $45.49_{\pm0.25}$ & - & - & - \\
& MedicalNet+BAG & $44.96_{\pm3.66}$ & - & -  & - \\
& y-aware+BAG & $46.54_{\pm3.80}$ & - & -  & - \\
\midrule

\textbf{ADNI+NIFD+PPMI} [AD/FTD/PD]
& \textbf{P+A} &-& $61.31_{\pm 1.50}$ & $67.20_{\pm 4.12}$ & $66.44_{\pm 3.59}$ \\
\midrule

\textbf{ADNI+NIFD+PPMI} [AD/FTD/PD/CN]
& \textbf{P+A} &-& $63.30_{\pm 1.01}$ & $68.12_{\pm 2.14}$ & $68.02_{\pm 1.83}$ \\
\bottomrule
\end{tabular}
}
\end{table*}

\begin{table*}[t]
\centering
\scriptsize
\setlength{\tabcolsep}{1.5pt}
\renewcommand{\arraystretch}{1.18}
\caption{Full-train linear probing performance (in \%) across datasets and binary tasks for AD and MCI classes. Prior work rows are reported with full finetuning (end-to-end). BrainNorm methods are shown under No FT, FT (own), and FT (all). Values are reported as $mean_{\pm std}$. Within each task block, the best value is in \textbf{bold} and second-best is \underline{underlined}.}
\label{tab:fulltrain_linearprobe_binary_admci_baselines_brainnorm}
\begin{tabular}{lcccccccc}
\toprule
\multirow{2}{*}{\textbf{Method}} & \multicolumn{2}{c}{\textbf{Full FT}} & \multicolumn{2}{c}{\textbf{No FT+LP}} & \multicolumn{2}{c}{\textbf{CN-only FT (own)+LP}} & \multicolumn{2}{c}{\textbf{CN-only FT (all)+LP}} \\
\cmidrule(lr){2-3}\cmidrule(lr){4-5}\cmidrule(lr){6-7}\cmidrule(lr){8-9}
 & \textbf{BACC} & \textbf{AUC} & \textbf{BACC} & \textbf{AUC} & \textbf{BACC} & \textbf{AUC} & \textbf{BACC} & \textbf{AUC} \\
\midrule
\multicolumn{9}{l}{\textbf{ADNI | AD/CN}} \\
\addlinespace[2pt]
3D-ViT & $78.06_{\pm 3.56}$ & $83.63_{\pm 2.61}$ & -- & -- & -- & -- & -- & -- \\
BrainHarmony & $66.51_{\pm 7.98}$ & $78.78_{\pm 5.70}$& -- & -- & -- & -- & -- & -- \\
BrainMVP & $71.94_{\pm 2.15}$ & $83.11_{\pm 4.65}$ & -- & -- & -- & -- & -- & -- \\
BrainIAC & $66.70_{\pm 2.56}$ & $74.66_{\pm 3.70}$ & -- & -- & -- & -- & -- & -- \\
NeuroVFM & $81.70_{\pm 2.28}$ & $90.40_{\pm 1.43}$ & -- & -- & -- & -- & -- & -- \\
Neuro-JEPA & $85.66_{\pm 1.48}$ & $92.71_{\pm 1.49}$ & -- & -- & -- & -- & -- & -- \\
RadFM & $79.33_{\pm 6.58}$ & $90.77_{\pm 2.14}$ & -- & -- & -- & -- & -- & -- \\
MedicalNet & $81.79_{\pm 2.80}$ & $88.79_{\pm 2.76}$ & -- & -- & -- & -- & -- & -- \\
y-Aware & $85.27_{\pm 1.74}$ & $93.07_{\pm 1.57}$ & -- & -- & -- & -- & -- & -- \\
\cdashline{1-9}
BrainNorm\_M1 & -- & -- & $87.07_{\pm 1.61}$ & $93.35_{\pm 2.50}$ & $86.31_{\pm 1.57}$ & $93.42_{\pm 2.18}$ & $86.97_{\pm 1.93}$ & $93.65_{\pm 2.20}$ \\
BrainNorm\_M2 & -- & -- & $86.85_{\pm 2.87}$ & $94.27_{\pm 2.22}$ & $\underline{87.67_{\pm 3.13}}$ & $\underline{94.62_{\pm 2.12}}$ & $\mathbf{87.34_{\pm 2.78}}$ & $\mathbf{94.66_{\pm 1.95}}$ \\
BrainNorm\_M3 & -- & -- & $76.48_{\pm 2.61}$ & $83.81_{\pm 2.42}$ & $78.81_{\pm 2.05}$ & $86.19_{\pm 1.77}$ & $79.27_{\pm 2.19}$ & $86.71_{\pm 1.31}$ \\
\midrule
\multicolumn{9}{l}{\textbf{ADNI | sMCI/pMCI}} \\
\addlinespace[2pt]
3D-ViT & $60.82_{\pm 2.80}$ & $64.70_{\pm 4.93}$ & -- & -- & -- & -- & -- & -- \\
BrainHarmony & $55.10_{\pm3.05}$ & $70.42_{\pm2.22}$ & -- & -- & -- & -- & -- & -- \\
BrainMVP & $51.79_{\pm 3.61}$ & $70.59_{\pm 3.28}$ & -- & -- & -- & -- & -- & -- \\
BrainIAC & $56.56_{\pm 1.87}$ & $62.17_{\pm 1.20}$ & -- & -- & -- & -- & -- & -- \\
NeuroVFM & $60.25_{\pm 4.40}$ & $72.27_{\pm 3.58}$ & -- & -- & -- & -- & -- & -- \\
Neuro-JEPA & $63.59_{\pm 5.96}$ & $67.91_{\pm 7.16}$ & -- & -- & -- & -- & -- & -- \\
RadFM & $58.03_{\pm 5.52}$ & $71.00_{\pm 2.57}$ & -- & -- & -- & -- & -- & -- \\
MedicalNet & $67.88_{\pm 3.75}$ & $72.80_{\pm 4.61}$ & -- & -- & -- & -- & -- & -- \\
y-Aware & $63.63_{\pm 7.94}$ & $78.41_{\pm 3.10}$ & -- & -- & -- & -- & -- & -- \\
\cdashline{1-9}
BrainNorm\_M1 & -- & -- & $69.88_{\pm 1.92}$ & $77.90_{\pm 2.48}$ & $71.88_{\pm1.60}$ & $78.84_{\pm2.39}$ & $72.40_{\pm 2.00}$ & $79.29_{\pm 2.15}$ \\
BrainNorm\_M2 & -- & -- & $71.96_{\pm 4.10}$ & $80.75_{\pm 3.80}$ & $\underline{73.59_{\pm2.82}}$ & $\underline{81.11_{\pm2.99}}$ & $\mathbf{74.43_{\pm 3.03}}$ & $\mathbf{81.76_{\pm 3.36}}$ \\
BrainNorm\_M3 & -- & -- & $64.53_{\pm 2.83}$ & $70.41_{\pm 3.42}$ & $69.61_{\pm2.76}$ & $76.21_{\pm4.11}$ & $68.68_{\pm 2.46}$ & $75.34_{\pm 4.65}$ \\
\midrule
\multicolumn{9}{l}{\textbf{AIBL | AD/CN}} \\
\addlinespace[2pt]
3D-ViT & $63.26_{\pm 2.58}$ & $78.26_{\pm 3.00}$ & -- & -- & -- & -- & -- & -- \\
BrainHarmony & $53.79_{\pm 1.15}$ & $72.36_{\pm 4.57}$& -- & -- & -- & -- & -- & -- \\
BrainMVP & $53.00_{\pm 4.91}$ & $72.89_{\pm 4.64}$ & -- & -- & -- & -- & -- & -- \\
BrainIAC & $54.69_{\pm 3.79}$ & $69.89_{\pm 4.94}$ & -- & -- & -- & -- & -- & -- \\
NeuroVFM & $55.82_{\pm 8.01}$ & $80.56_{\pm 7.19}$ & -- & -- & -- & -- & -- & -- \\
Neuro-JEPA & $83.43_{\pm 6.99}$ & $90.01_{\pm 3.75}$ & -- & -- & -- & -- & -- & -- \\
RadFM & $69.54_{\pm 7.41}$ & $87.36_{\pm 4.42}$ & -- & -- & -- & -- & -- & -- \\
MedicalNet & $60.62_{\pm 5.55}$ & $73.68_{\pm 3.79}$ & -- & -- & -- & -- & -- & -- \\
y-Aware & $67.67_{\pm 9.36}$ & $83.79_{\pm 8.13}$ & -- & -- & -- & -- & -- & -- \\
\cdashline{1-9}
BrainNorm\_M1 & -- & -- & $82.38_{\pm 7.91}$ & $92.82_{\pm 0.91}$ & $85.10_{\pm 4.30}$ & $92.48_{\pm 0.70}$ & $85.34_{\pm 4.36}$ & $92.92_{\pm 0.68}$ \\
BrainNorm\_M2 & -- & -- & $\underline{85.91_{\pm 5.47}}$ & $93.39_{\pm 1.71}$ & $85.82_{\pm 3.24}$ & $\underline{93.62_{\pm 1.66}}$ & $\mathbf{87.01_{\pm 4.80}}$ & $\mathbf{93.64_{\pm 1.30}}$ \\
BrainNorm\_M3 & -- & -- & $56.66_{\pm 6.26}$ & $58.70_{\pm 6.46}$ & $58.23_{\pm 4.86}$ & $62.61_{\pm 3.59}$ & $62.39_{\pm 7.90}$ & $66.74_{\pm 11.25}$ \\
\midrule
\multicolumn{9}{l}{\textbf{MIRIAD | AD/CN}} \\
\addlinespace[2pt]
3D-ViT & $82.29_{\pm 4.84}$ & $84.52_{\pm 4.29}$ & -- & -- & -- & -- & -- & -- \\
BrainHarmony & $79.82_{\pm 9.19}$ & $84.38_{\pm8.49}$ & -- & -- & -- & -- & -- & -- \\
BrainMVP & $73.73_{\pm 5.49}$ & $91.49_{\pm 4.98}$ & -- & -- & -- & -- & -- & -- \\
BrainIAC & $83.85_{\pm 5.85}$ & $91.27_{\pm 5.21}$ & -- & -- & -- & -- & -- & -- \\
NeuroVFM & $74.60_{\pm 8.29}$ & $93.11_{\pm 6.06}$ & -- & -- & -- & -- & -- & -- \\
Neuro-JEPA & $85.87_{\pm 9.85}$ & $93.45_{\pm 7.21}$ & -- & -- & -- & -- & -- & -- \\
RadFM & $80.71_{\pm 9.82}$ & $92.41_{\pm 9.80}$ & -- & -- & -- & -- & -- & -- \\
MedicalNet & $88.37_{\pm 5.37}$ & $92.22_{\pm 7.57}$ & -- & -- & -- & -- & -- & -- \\
y-Aware & $87.81_{\pm 5.65}$ & $96.74_{\pm 2.15}$ & -- & -- & -- & -- & -- & -- \\
\cdashline{1-9}
BrainNorm\_M1 & -- & -- & $87.92_{\pm 9.24}$ & $95.63_{\pm 5.10}$ & $87.87_{\pm 8.34}$ & $95.63_{\pm 5.22}$ & $86.58_{\pm 7.73}$ & $95.83_{\pm 5.74}$ \\
BrainNorm\_M2 & -- & -- & $\underline{92.72_{\pm 10.00}}$ & $\mathbf{98.26_{\pm 3.88}}$ & $92.60_{\pm 10.09}$ & $\underline{98.13_{\pm 4.10}}$ & $\mathbf{92.97_{\pm 8.97}}$ & $97.92_{\pm 4.63}$ \\
BrainNorm\_M3 & -- & -- & $86.49_{\pm 11.65}$ & $90.39_{\pm 11.27}$ & $84.63_{\pm 8.87}$ & $92.93_{\pm 8.51}$ & $90.32_{\pm 8.63}$ & $97.15_{\pm 4.73}$ \\
\bottomrule
\end{tabular}
\end{table*}

\begin{table*}[t]
\centering
\scriptsize
\setlength{\tabcolsep}{1.5pt}
\renewcommand{\arraystretch}{1.18}
\caption{Full-train linear probing performance (in \%) across datasets and binary tasks for FTD and PD classes. Prior work rows are reported with full finetuning (end-to-end). BrainNorm methods are shown under No FT, FT (own), and FT (all). Values are reported as $mean_{\pm std}$. Within each task block, the best value is in \textbf{bold} and second-best is \underline{underlined}.}
\label{tab:fulltrain_linearprobe_binary_pdftd_baselines_brainnorm}
\begin{tabular}{lcccccccc}
\toprule
\multirow{2}{*}{\textbf{Method}} & \multicolumn{2}{c}{\textbf{Full FT}} & \multicolumn{2}{c}{\textbf{No FT+LP}} & \multicolumn{2}{c}{\textbf{CN-only FT (own)+LP}} & \multicolumn{2}{c}{\textbf{CN-only FT (all)+LP}} \\
\cmidrule(lr){2-3}\cmidrule(lr){4-5}\cmidrule(lr){6-7}\cmidrule(lr){8-9}
 & \textbf{BACC} & \textbf{AUC} & \textbf{BACC} & \textbf{AUC} & \textbf{BACC} & \textbf{AUC} & \textbf{BACC} & \textbf{AUC} \\
\midrule
\multicolumn{9}{l}{\textbf{NIFD | FTD/CN}} \\
\addlinespace[2pt]
3D-ViT & $52.96_{\pm 1.98}$ & $60.68_{\pm 7.12}$ & -- & -- & -- & -- & -- & -- \\
BrainHarmony & $52.59_{\pm3.56}$ & $69.42_{\pm3.36}$& -- & -- & -- & -- & -- & -- \\
BrainMVP & $56.60_{\pm 4.34}$ & $82.26_{\pm 6.45}$ & -- & -- & -- & -- & -- & -- \\
BrainIAC & $53.86_{\pm 5.68}$ & $66.12_{\pm 5.12}$ & -- & -- & -- & -- & -- & -- \\
NeuroVFM & $61.22_{\pm 5.53}$ & $82.03_{\pm 5.88}$ & -- & -- & -- & -- & -- & -- \\
Neuro-JEPA & $80.30_{\pm 2.29}$ & $86.88_{\pm 3.83}$ & -- & -- & -- & -- & -- & -- \\
RadFM & $70.08_{\pm 6.41}$ & $81.09_{\pm 5.85}$ & -- & -- & -- & -- & -- & -- \\
MedicalNet & $78.75_{\pm 3.52}$ & $87.69_{\pm 2.76}$ & -- & -- & -- & -- & -- & -- \\
y-Aware & $77.92_{\pm 6.58}$ & $89.14_{\pm 4.01}$ & -- & -- & -- & -- & -- & -- \\
\cdashline{1-9}
BrainNorm\_M1 & -- & -- & $80.87_{\pm 3.79}$ & $88.55_{\pm 4.49}$ & $81.49_{\pm 5.36}$ & $88.71_{\pm 4.60}$ & $79.45_{\pm 2.44}$ & $88.42_{\pm 3.07}$ \\
BrainNorm\_M2 & -- & -- & $\underline{81.82_{\pm 6.92}}$ & $\underline{89.15_{\pm 4.29}}$ & $\mathbf{82.62_{\pm 7.61}}$ & $\mathbf{89.34_{\pm 4.77}}$ & $78.45_{\pm 2.86}$ & $88.22_{\pm 4.43}$ \\
BrainNorm\_M3 & -- & -- & $77.64_{\pm 2.85}$ & $82.97_{\pm 6.26}$ & $72.04_{\pm 3.47}$ & $82.26_{\pm 4.20}$ & $75.45_{\pm 4.54}$ & $81.40_{\pm 5.59}$ \\
\midrule
\multicolumn{9}{l}{\textbf{PPMI | PD/CN}} \\
\addlinespace[2pt]
3D-ViT & $50.05_{\pm 0.27}$ & $54.75_{\pm 3.05}$ & -- & -- & -- & -- & -- & -- \\
BrainHarmony & $50.20_{\pm0.96}$ & $55.90_{\pm2.76}$ & -- & -- & -- & -- & -- & -- \\
BrainMVP & $51.00_{\pm0.91}$ & $59.41_{\pm2.75}$ & -- & -- & -- & -- & -- & -- \\
BrainIAC & $50.50_{\pm 0.71}$ & $57.38_{\pm 4.45}$ & -- & -- & -- & -- & -- & -- \\
NeuroVFM & $50.00_{\pm 0.00}$ & $59.64_{\pm 2.99}$ & -- & -- & -- & -- & -- & -- \\
Neuro-JEPA & $54.51_{\pm 6.49}$ & $56.70_{\pm 7.19}$ & -- & -- & -- & -- & -- & -- \\
RadFM & $50.27_{\pm 0.61}$ & $61.18_{\pm 4.06}$ & -- & -- & -- & -- & -- & -- \\
MedicalNet & $50.67_{\pm 0.60}$ & $59.51_{\pm 4.40}$ & -- & -- & -- & -- & -- & -- \\
y-Aware & $50.10_{\pm 0.70}$ & $57.28_{\pm 2.53}$ & -- & -- & -- & -- & -- & -- \\
\cdashline{1-9}
BrainNorm\_M1 & -- & -- & $\mathbf{60.43_{\pm 2.59}}$ & $\underline{66.01_{\pm 5.30}}$ & $55.36_{\pm 4.42}$ & $61.20_{\pm 4.92}$ & $56.12_{\pm 5.79}$ & $61.27_{\pm 5.74}$ \\
BrainNorm\_M2 & -- & -- & $\underline{59.08_{\pm 3.38}}$ & $\mathbf{66.12_{\pm 5.23}}$ & $57.73_{\pm 1.73}$ & $62.03_{\pm 5.16}$ & $58.00_{\pm 4.14}$ & $61.65_{\pm 6.53}$ \\
BrainNorm\_M3 & -- & -- & $52.40_{\pm 2.47}$ & $48.49_{\pm 4.08}$ & $50.19_{\pm 2.90}$ & $49.62_{\pm 5.17}$ & $51.59_{\pm 4.72}$ & $46.92_{\pm 5.33}$ \\
\bottomrule
\end{tabular}
\end{table*}

\begin{table*}[t]
\centering
\scriptsize
\setlength{\tabcolsep}{1.5pt}
\renewcommand{\arraystretch}{1.18}
\caption{Full-train linear probing performance across datasets and multi-class tasks. Prior work rows are reported with full finetuning (end-to-end). BrainNorm methods are shown under No FT, FT (own), and FT (all). Values are reported as $mean_{\pm std}$. Within each task block, the best value is in \textbf{bold} and second-best is \underline{underlined}.}
\label{tab:fulltrain_linearprobe_multi_baselines_brainnorm}
\begin{tabular}{lcccccccc}
\toprule
\multirow{2}{*}{\textbf{Method}} & \multicolumn{2}{c}{\textbf{Full FT}} & \multicolumn{2}{c}{\textbf{No FT+LP}} & \multicolumn{2}{c}{\textbf{CN-only FT (own)+LP}} & \multicolumn{2}{c}{\textbf{CN-only FT (all)+LP}} \\
\cmidrule(lr){2-3}\cmidrule(lr){4-5}\cmidrule(lr){6-7}\cmidrule(lr){8-9}
 & \textbf{BACC} & \textbf{AUC} & \textbf{BACC} & \textbf{AUC} & \textbf{BACC} & \textbf{AUC} & \textbf{BACC} & \textbf{AUC} \\
\midrule
\multicolumn{9}{l}{\textbf{ADNI+NIFD+PPMI | AD/FTD/PD}} \\
\addlinespace[2pt]
3D-ViT & $58.86_{\pm 1.65}$ & $79.72_{\pm 1.08}$ & -- & -- & -- & -- & -- & -- \\
BrainHarmony & $74.49_{\pm 3.83}$ & $90.57_{\pm 1.85}$ & -- & -- & -- & -- & -- & -- \\
BrainMVP & $82.79_{\pm 2.63}$ & $96.78_{\pm 0.92}$ & -- & -- & -- & -- & -- & -- \\
BrainIAC & $55.65_{\pm 1.50}$ & $79.17_{\pm 1.44}$ & -- & -- & -- & -- & -- & -- \\
NeuroVFM & $62.26_{\pm 7.80}$ & $86.97_{\pm 3.55}$ & -- & -- & -- & -- & -- & -- \\
Neuro-JEPA & $81.83_{\pm 2.63}$ & $94.87_{\pm 0.96}$ & -- & -- & -- & -- & -- & -- \\
RadFM & $75.39_{\pm 3.20}$ & $92.00_{\pm 0.80}$ & -- & -- & -- & -- & -- & -- \\
MedicalNet & $82.06_{\pm 2.71}$ & $94.64_{\pm 1.42}$ & -- & -- & -- & -- & -- & -- \\
y-Aware & $83.75_{\pm 1.52}$ & $96.70_{\pm 0.82}$ & -- & -- & -- & -- & -- & -- \\
\cdashline{1-9}
BrainNorm\_M1 & -- & -- & $83.21_{\pm 4.27}$ & $96.91_{\pm 0.88}$ & $82.90_{\pm 3.75}$ & $96.89_{\pm 0.72}$ & $83.33_{\pm 3.73}$ & $96.85_{\pm 0.63}$ \\
BrainNorm\_M2 & -- & -- & $83.98_{\pm 4.48}$ & $\mathbf{97.01_{\pm 0.99}}$ & $\underline{84.22_{\pm 3.94}}$ & $96.84_{\pm 0.95}$ & $\mathbf{84.24_{\pm 3.84}}$ & $\underline{97.00_{\pm 0.83}}$ \\
BrainNorm\_M3 & -- & -- & $66.85_{\pm 2.82}$ & $86.86_{\pm 1.77}$ & $66.29_{\pm 4.13}$ & $84.84_{\pm 2.59}$ & $64.87_{\pm 3.63}$ & $83.87_{\pm 2.94}$ \\
\midrule
\multicolumn{9}{l}{\textbf{ADNI+NIFD+PPMI | AD/FTD/PD/CN}} \\
\addlinespace[2pt]
3D-ViT & $46.68_{\pm 5.41}$ & $76.54_{\pm 3.04}$ & -- & -- & -- & -- & -- & -- \\
BrainHarmony & $49.07_{\pm 3.65}$ & $81.73_{\pm 2.24}$ & -- & -- & -- & -- & -- & -- \\
BrainMVP & $49.06_{\pm 2.00}$ & $79.66_{\pm 1.15}$ & -- & -- & -- & -- & -- & -- \\
BrainIAC & $53.36_{\pm 2.00}$ & $72.24_{\pm 1.15}$ & -- & -- & -- & -- & -- & -- \\
NeuroVFM & $49.68_{\pm 8.34}$ & $83.80_{\pm 2.31}$ & -- & -- & -- & -- & -- & -- \\
Neuro-JEPA & $68.53_{\pm 3.70}$ & $88.56_{\pm 1.45}$ & -- & -- & -- & -- & -- & -- \\
RadFM & $57.44_{\pm 3.06}$ & $84.61_{\pm 1.44}$ & -- & -- & -- & -- & -- & -- \\
MedicalNet & $59.51_{\pm 3.33}$ & $84.87_{\pm 2.23}$ & -- & -- & -- & -- & -- & -- \\
y-Aware & $69.87_{\pm 1.96}$ & $90.81_{\pm 1.44}$ & -- & -- & -- & -- & -- & -- \\
\cdashline{1-9}
BrainNorm\_M1 & -- & -- & $67.10_{\pm 2.52}$ & $90.73_{\pm 0.93}$ & $67.74_{\pm 1.97}$ & $90.82_{\pm 1.32}$ & $67.55_{\pm 2.12}$ & $91.01_{\pm 1.22}$ \\
BrainNorm\_M2 & -- & -- & $\mathbf{70.32_{\pm 1.77}}$ & $\mathbf{91.91_{\pm 1.24}}$ & $\underline{70.23_{\pm 2.31}}$ & $91.78_{\pm 1.63}$ & $70.09_{\pm 1.83}$ & $\underline{91.82_{\pm 1.63}}$ \\
BrainNorm\_M3 & -- & -- & $51.31_{\pm 3.44}$ & $83.11_{\pm 1.11}$ & $44.93_{\pm 2.10}$ & $79.08_{\pm 1.90}$ & $44.40_{\pm 2.45}$ & $79.76_{\pm 1.98}$ \\
\bottomrule
\end{tabular}
\end{table*}

\begin{table*}[t]
\centering
\scriptsize
\setlength{\tabcolsep}{3.0pt}
\renewcommand{\arraystretch}{1.12}
\caption{Few-shot performance across binary tasks and datasets for BrainNorm variants. Each method is shown under three modes: No FT, CN-only FT on the corresponding dataset only, and CN-only FT on pooled CN data from ADNI+AIBL+MIRIAD+PPMI+NIFD. Values are reported as mean $\pm$ standard deviation.}
\label{tab:fewshot_brainnorm_binary}
\begin{tabular}{llccccc}
\toprule
\textbf{Method} & \textbf{Mode} & $\mathbf{k=1}$ & $\mathbf{k=2}$ & $\mathbf{k=4}$ & $\mathbf{k=8}$ & $\mathbf{k=16}$ \\
\midrule
\multicolumn{6}{l}{\textbf{ADNI | [AD/CN]}} \\
\addlinespace[1pt]
\multirow{3}{*}{BrainNorm\_M1} & No FT + LP & $55.31_{\pm 25.18}$ & $57.69_{\pm 25.01}$ & $62.54_{\pm 19.82}$ & $75.21_{\pm 7.97}$ & $80.60_{\pm 2.06}$ \\
 & CN-only FT(own) + LP & $68.86_{\pm 23.78}$ & $67.60_{\pm 22.95}$ & $59.59_{\pm 27.55}$ & $80.23_{\pm 1.77}$ & $80.84_{\pm 2.49}$ \\
 & CN-only FT (all) + LP & $69.22_{\pm 23.65}$ & $69.47_{\pm 21.01}$ & $59.60_{\pm 27.49}$ & $80.33_{\pm 2.61}$ & $81.09_{\pm 2.21}$ \\
\cline{2-7}
\addlinespace[1pt]
\multirow{3}{*}{BrainNorm\_M2} & No FT + LP & $55.30_{\pm 13.75}$ & $50.70_{\pm 11.70}$ & $70.57_{\pm 9.48}$ & $78.28_{\pm 6.23}$ & $83.16_{\pm 4.02}$ \\
 & CN-only FT(own) + LP & $79.51_{\pm 10.37}$ & $69.28_{\pm 21.02}$ & $77.13_{\pm 12.10}$ & $84.95_{\pm 1.88}$ & $86.08_{\pm 3.34}$ \\
 & CN-only FT (all) + LP & $79.09_{\pm 11.48}$ & $71.64_{\pm 16.71}$ & $78.14_{\pm 10.73}$ & $85.18_{\pm 1.24}$ & $86.20_{\pm 2.85}$ \\
\cline{2-7}
\addlinespace[1pt]
\multirow{3}{*}{BrainNorm\_M3} & No FT + LP & $58.12_{\pm 5.38}$ & $55.21_{\pm 5.64}$ & $52.04_{\pm 7.94}$ & $54.60_{\pm 6.48}$ & $62.73_{\pm 8.69}$ \\
 & CN-only FT(own) + LP & $69.24_{\pm 1.52}$ & $64.68_{\pm 5.49}$ & $64.81_{\pm 6.61}$ & $51.97_{\pm 15.69}$ & $61.45_{\pm 9.33}$ \\
 & CN-only FT (all) + LP & $67.74_{\pm 3.82}$ & $61.49_{\pm 5.65}$ & $61.68_{\pm 6.62}$ & $43.83_{\pm 12.94}$ & $64.79_{\pm 3.60}$ \\
\addlinespace[1pt]
\midrule
\multicolumn{6}{l}{\textbf{AIBL | [AD/CN]}} \\
\addlinespace[1pt]
\multirow{3}{*}{BrainNorm\_M1} & No FT + LP & $84.06_{\pm 4.36}$ & $80.07_{\pm 8.27}$ & $82.50_{\pm 5.89}$ & $84.36_{\pm 5.29}$ & $85.55_{\pm 4.04}$ \\
 & CN-only FT(own) + LP & $85.46_{\pm 3.15}$ & $78.31_{\pm 7.09}$ & $85.17_{\pm 3.81}$ & $85.49_{\pm 3.82}$ & $86.21_{\pm 3.09}$ \\
 & CN-only FT (all) + LP & $85.83_{\pm 4.26}$ & $84.11_{\pm 5.60}$ & $86.68_{\pm 4.23}$ & $85.61_{\pm 4.43}$ & $85.97_{\pm 4.06}$ \\
\cline{2-7}
\addlinespace[1pt]
\multirow{3}{*}{BrainNorm\_M2} & No FT + LP & $79.87_{\pm 2.29}$ & $78.24_{\pm 8.15}$ & $81.14_{\pm 7.51}$ & $85.26_{\pm 6.41}$ & $85.53_{\pm 5.29}$ \\
 & CN-only FT(own) + LP & $85.53_{\pm 3.17}$ & $82.40_{\pm 7.13}$ & $83.63_{\pm 7.06}$ & $87.91_{\pm 5.26}$ & $88.19_{\pm 3.81}$ \\
 & CN-only FT (all) + LP & $88.23_{\pm 2.36}$ & $85.15_{\pm 4.84}$ & $85.89_{\pm 7.04}$ & $89.24_{\pm 3.46}$ & $88.75_{\pm 3.40}$ \\
\cline{2-7}
\addlinespace[1pt]
\multirow{3}{*}{BrainNorm\_M3} & No FT + LP & $55.27_{\pm 2.65}$ & $56.55_{\pm 7.61}$ & $54.55_{\pm 5.61}$ & $51.73_{\pm 7.23}$ & $58.54_{\pm 9.07}$ \\
 & CN-only FT(own) + LP & $67.42_{\pm 4.35}$ & $49.57_{\pm 9.01}$ & $54.73_{\pm 9.72}$ & $47.20_{\pm 7.82}$ & $67.12_{\pm 8.45}$ \\
 & CN-only FT (all) + LP & $65.50_{\pm 7.56}$ & $47.37_{\pm 9.43}$ & $52.35_{\pm 9.79}$ & $55.75_{\pm 9.26}$ & $61.38_{\pm 8.51}$ \\
\addlinespace[1pt]
\midrule
\multicolumn{6}{l}{\textbf{MIRIAD | [AD/CN]}} \\
\addlinespace[1pt]
\multirow{3}{*}{BrainNorm\_M1} & No FT + LP & $90.07_{\pm 8.61}$ & $91.11_{\pm 9.15}$ & $94.79_{\pm 5.81}$ & $94.67_{\pm 5.52}$ & $94.41_{\pm 5.98}$ \\
 & CN-only FT(own) + LP & $88.08_{\pm 9.44}$ & $88.11_{\pm 9.80}$ & $94.40_{\pm 5.92}$ & $94.62_{\pm 5.77}$ & $94.42_{\pm 6.14}$ \\
 & CN-only FT (all) + LP & $81.37_{\pm 9.26}$ & $82.70_{\pm 9.34}$ & $94.77_{\pm 6.48}$ & $94.83_{\pm 6.45}$ & $94.74_{\pm 6.53}$ \\
\cline{2-7}
\addlinespace[1pt]
\multirow{3}{*}{BrainNorm\_M2} & No FT + LP & $83.50_{\pm 7.97}$ & $89.89_{\pm 9.89}$ & $93.39_{\pm 6.16}$ & $93.81_{\pm 4.24}$ & $97.19_{\pm 4.64}$ \\
 & CN-only FT(own) + LP & $88.97_{\pm 7.89}$ & $77.30_{\pm 9.43}$ & $95.30_{\pm 3.90}$ & $94.24_{\pm 5.05}$ & $96.91_{\pm 5.62}$ \\
 & CN-only FT (all) + LP & $92.50_{\pm 6.31}$ & $92.10_{\pm 9.14}$ & $97.17_{\pm 3.87}$ & $96.69_{\pm 4.35}$ & $97.66_{\pm 4.64}$ \\
\cline{2-7}
\addlinespace[1pt]
\multirow{3}{*}{BrainNorm\_M3} & No FT + LP & $73.05_{\pm 9.60}$ & $60.20_{\pm 9.30}$ & $69.40_{\pm 7.21}$ & $51.42_{\pm 9.09}$ & $76.14_{\pm 8.85}$ \\
 & CN-only FT(own) + LP & $77.72_{\pm 7.91}$ & $70.01_{\pm 7.57}$ & $74.96_{\pm 8.04}$ & $58.39_{\pm 9.93}$ & $72.24_{\pm 9.21}$ \\
 & CN-only FT (all) + LP & $83.78_{\pm 7.90}$ & $64.72_{\pm 9.94}$ & $77.91_{\pm 7.68}$ & $70.99_{\pm 9.19}$ & $84.47_{\pm 9.17}$ \\
\addlinespace[1pt]
\midrule
\multicolumn{6}{l}{\textbf{NIFD | [FTD/CN]}} \\
\addlinespace[1pt]
\multirow{3}{*}{BrainNorm\_M1} & No FT + LP & $75.92_{\pm 7.00}$ & $77.55_{\pm 3.57}$ & $77.85_{\pm 3.42}$ & $78.86_{\pm 3.28}$ & $79.76_{\pm 3.13}$ \\
 & CN-only FT(own) + LP & $75.77_{\pm 6.77}$ & $77.36_{\pm 3.82}$ & $77.79_{\pm 3.40}$ & $78.79_{\pm 3.08}$ & $78.78_{\pm 3.78}$ \\
 & CN-only FT (all) + LP & $77.15_{\pm 6.71}$ & $79.56_{\pm 3.15}$ & $80.12_{\pm 2.94}$ & $80.54_{\pm 3.07}$ & $80.44_{\pm 3.43}$ \\
\cline{2-7}
\addlinespace[1pt]
\multirow{3}{*}{BrainNorm\_M2} & No FT + LP & $76.13_{\pm 7.36}$ & $68.57_{\pm 9.56}$ & $80.34_{\pm 6.08}$ & $81.50_{\pm 6.52}$ & $81.25_{\pm 5.35}$ \\
 & CN-only FT(own) + LP & $78.08_{\pm 8.74}$ & $67.56_{\pm 8.15}$ & $81.31_{\pm 6.33}$ & $82.17_{\pm 6.67}$ & $81.50_{\pm 5.79}$ \\
 & CN-only FT (all) + LP & $82.31_{\pm 6.36}$ & $69.58_{\pm 9.81}$ & $83.51_{\pm 4.47}$ & $84.22_{\pm 5.04}$ & $84.03_{\pm 4.44}$ \\
\cline{2-7}
\addlinespace[1pt]
\multirow{3}{*}{BrainNorm\_M3} & No FT + LP & $57.96_{\pm 7.49}$ & $43.29_{\pm 8.14}$ & $40.93_{\pm 7.22}$ & $67.92_{\pm 4.26}$ & $63.60_{\pm 9.48}$ \\
 & CN-only FT(own) + LP & $59.37_{\pm 7.62}$ & $44.14_{\pm 8.86}$ & $53.96_{\pm 8.69}$ & $54.86_{\pm 8.47}$ & $59.08_{\pm 6.90}$ \\
 & CN-only FT (all) + LP & $52.81_{\pm 9.17}$ & $39.87_{\pm 9.47}$ & $43.67_{\pm 9.06}$ & $60.71_{\pm 9.85}$ & $62.56_{\pm 9.62}$ \\
\addlinespace[1pt]
\midrule
\multicolumn{6}{l}{\textbf{PPMI | [PD/CN]}} \\
\addlinespace[1pt]
\multirow{3}{*}{BrainNorm\_M1} & No FT + LP & $58.39_{\pm 5.47}$ & $57.75_{\pm 5.56}$ & $58.55_{\pm 5.24}$ & $57.85_{\pm 5.31}$ & $57.60_{\pm 4.83}$ \\
 & CN-only FT(own) + LP & $57.54_{\pm 4.87}$ & $56.23_{\pm 5.61}$ & $58.12_{\pm 5.11}$ & $57.48_{\pm 5.47}$ & $57.47_{\pm 5.15}$ \\
 & CN-only FT (all) + LP & $58.51_{\pm 5.04}$ & $56.56_{\pm 5.78}$ & $58.19_{\pm 5.25}$ & $58.09_{\pm 5.41}$ & $58.03_{\pm 5.62}$ \\
\cline{2-7}
\addlinespace[1pt]
\multirow{3}{*}{BrainNorm\_M2} & No FT + LP & $52.79_{\pm 2.76}$ & $54.52_{\pm 4.30}$ & $57.71_{\pm 6.21}$ & $53.23_{\pm 5.58}$ & $55.02_{\pm 2.60}$ \\
 & CN-only FT(own) + LP & $52.05_{\pm 2.87}$ & $52.71_{\pm 5.56}$ & $56.76_{\pm 6.38}$ & $53.18_{\pm 5.25}$ & $55.33_{\pm 3.59}$ \\
 & CN-only FT (all) + LP & $53.85_{\pm 3.35}$ & $54.85_{\pm 6.76}$ & $59.25_{\pm 6.55}$ & $55.57_{\pm 3.42}$ & $57.16_{\pm 4.66}$ \\
\cline{2-7}
\addlinespace[1pt]
\multirow{3}{*}{BrainNorm\_M3} & No FT + LP & $50.32_{\pm 2.39}$ & $49.96_{\pm 3.37}$ & $49.39_{\pm 3.19}$ & $51.02_{\pm 4.09}$ & $50.59_{\pm 4.21}$ \\
 & CN-only FT(own) + LP & $51.26_{\pm 2.31}$ & $49.83_{\pm 3.60}$ & $49.70_{\pm 3.43}$ & $49.80_{\pm 3.53}$ & $50.98_{\pm 1.78}$ \\
 & CN-only FT (all) + LP & $48.73_{\pm 3.39}$ & $46.82_{\pm 4.09}$ & $47.94_{\pm 4.54}$ & $46.90_{\pm 4.77}$ & $48.56_{\pm 2.95}$ \\
\addlinespace[1pt]
\midrule
\multicolumn{6}{l}{\textbf{ADNI | [sMCI/pMCI]}} \\
\addlinespace[1pt]
\multirow{3}{*}{BrainNorm\_M1} & No FT + LP & $47.51_{\pm 8.51}$ & $54.44_{\pm 8.45}$ & $54.86_{\pm 7.34}$ & $55.02_{\pm 7.63}$ & $63.71_{\pm 5.73}$ \\
 & CN-only FT(own) + LP & $45.83_{\pm 8.04}$ & $54.48_{\pm 9.32}$ & $53.24_{\pm 9.68}$ & $55.27_{\pm 7.51}$ & $65.14_{\pm 5.47}$ \\
 & CN-only FT (all) + LP & $45.81_{\pm 9.68}$ & $53.77_{\pm 9.69}$ & $53.10_{\pm 9.34}$ & $54.49_{\pm 8.52}$ & $66.92_{\pm 6.68}$ \\
\cline{2-7}
\addlinespace[1pt]
\multirow{3}{*}{BrainNorm\_M2} & No FT + LP & $48.93_{\pm 8.31}$ & $55.41_{\pm 4.76}$ & $61.50_{\pm 6.58}$ & $61.44_{\pm 6.34}$ & $62.36_{\pm 6.41}$ \\
 & CN-only FT(own) + LP & $48.85_{\pm 8.94}$ & $59.82_{\pm 9.02}$ & $67.08_{\pm 6.04}$ & $65.56_{\pm 7.99}$ & $69.30_{\pm 4.74}$ \\
 & CN-only FT (all) + LP & $48.59_{\pm 8.33}$ & $59.09_{\pm 6.11}$ & $67.03_{\pm 7.30}$ & $67.97_{\pm 8.38}$ & $70.60_{\pm 6.67}$ \\
\cline{2-7}
\addlinespace[1pt]
\multirow{3}{*}{BrainNorm\_M3} & No FT + LP & $50.52_{\pm 5.35}$ & $46.67_{\pm 6.03}$ & $45.99_{\pm 5.67}$ & $47.60_{\pm 8.16}$ & $47.01_{\pm 8.25}$ \\
 & CN-only FT(own) + LP & $59.41_{\pm 3.15}$ & $50.33_{\pm 7.92}$ & $53.25_{\pm 7.06}$ & $53.89_{\pm 9.06}$ & $53.85_{\pm 8.44}$ \\
 & CN-only FT (all) + LP & $57.61_{\pm 5.56}$ & $51.30_{\pm 8.49}$ & $55.07_{\pm 7.93}$ & $52.41_{\pm 6.33}$ & $51.81_{\pm 8.90}$ \\
\addlinespace[1pt]
\bottomrule
\end{tabular}
\end{table*}

\begin{table*}[t]
\centering
\scriptsize
\setlength{\tabcolsep}{3.0pt}
\renewcommand{\arraystretch}{1.12}
\caption{Few-shot performance across multi-class tasks for BrainNorm variants. Each method is shown under three modes: No FT, CN-only FT on the corresponding dataset only, and CN-only FT on pooled CN data from ADNI+AIBL+MIRIAD+PPMI+NIFD. Values are reported as mean $\pm$ standard deviation.}
\label{tab:fewshot_brainnorm_multiclass}
\begin{tabular}{llccccc}
\toprule
\textbf{Method} & \textbf{Mode} & $\mathbf{k=1}$ & $\mathbf{k=2}$ & $\mathbf{k=4}$ & $\mathbf{k=8}$ & $\mathbf{k=16}$ \\
\midrule
\multicolumn{6}{l}{\textbf{ADNI+NIFD+PPMI | [AD/FTD/PD]}} \\
\addlinespace[1pt]
\multirow{3}{*}{BrainNorm\_M1} & No FT + LP & $57.02_{\pm 9.59}$ & $64.05_{\pm 9.45}$ & $80.38_{\pm 3.50}$ & $82.35_{\pm 2.93}$ & $85.18_{\pm 3.57}$ \\
 & CN-only FT(own) + LP & $58.15_{\pm 9.92}$ & $62.82_{\pm 9.94}$ & $81.12_{\pm 3.45}$ & $83.36_{\pm 3.50}$ & $85.13_{\pm 2.73}$ \\
 & CN-only FT (all) + LP & $57.74_{\pm 9.61}$ & $62.38_{\pm 9.24}$ & $81.08_{\pm 3.52}$ & $83.39_{\pm 3.95}$ & $84.92_{\pm 3.55}$ \\
\cline{2-7}
\addlinespace[1pt]
\multirow{3}{*}{BrainNorm\_M2} & No FT + LP & $58.97_{\pm 8.17}$ & $66.91_{\pm 8.10}$ & $78.65_{\pm 5.27}$ & $82.12_{\pm 3.50}$ & $84.34_{\pm 3.06}$ \\
 & CN-only FT(own) + LP & $58.67_{\pm 8.78}$ & $67.13_{\pm 8.68}$ & $79.83_{\pm 5.17}$ & $81.17_{\pm 4.52}$ & $83.08_{\pm 1.11}$ \\
 & CN-only FT (all) + LP & $58.46_{\pm 9.95}$ & $67.38_{\pm 9.30}$ & $80.40_{\pm 4.89}$ & $81.71_{\pm 4.10}$ & $83.33_{\pm 1.88}$ \\
\cline{2-7}
\addlinespace[1pt]
\multirow{3}{*}{BrainNorm\_M3} & No FT + LP & $53.18_{\pm 6.50}$ & $50.36_{\pm 7.08}$ & $52.60_{\pm 8.68}$ & $53.73_{\pm 8.67}$ & $63.86_{\pm 8.68}$ \\
 & CN-only FT(own) + LP & $51.18_{\pm 8.62}$ & $50.39_{\pm 5.39}$ & $52.84_{\pm 8.99}$ & $53.47_{\pm 5.39}$ & $66.13_{\pm 3.71}$ \\
 & CN-only FT (all) + LP & $50.95_{\pm 9.40}$ & $54.15_{\pm 6.12}$ & $53.28_{\pm 7.56}$ & $53.04_{\pm 6.34}$ & $66.91_{\pm 4.28}$ \\
\addlinespace[1pt]
\midrule
\multicolumn{6}{l}{\textbf{ADNI+NIFD+PPMI | [AD/FTD/PD/CN]}} \\
\addlinespace[1pt]
\multirow{3}{*}{BrainNorm\_M1} & No FT + LP & $68.04_{\pm 3.85}$ & $67.86_{\pm 3.92}$ & $69.10_{\pm 5.04}$ & $73.87_{\pm 2.06}$ & $75.60_{\pm 2.76}$ \\
 & CN-only FT(own) + LP & $68.27_{\pm 3.33}$ & $69.52_{\pm 5.52}$ & $71.80_{\pm 3.52}$ & $72.81_{\pm 4.60}$ & $76.47_{\pm 3.14}$ \\
 & CN-only FT (all) + LP & $68.42_{\pm 2.91}$ & $69.51_{\pm 5.51}$ & $71.95_{\pm 3.21}$ & $72.48_{\pm 4.55}$ & $76.61_{\pm 2.51}$ \\
\cline{2-7}
\addlinespace[1pt]
\multirow{3}{*}{BrainNorm\_M2} & No FT + LP & $65.18_{\pm 7.85}$ & $69.72_{\pm 6.86}$ & $73.32_{\pm 4.96}$ & $77.32_{\pm 2.31}$ & $78.88_{\pm 2.49}$ \\
 & CN-only FT(own) + LP & $62.96_{\pm 8.59}$ & $71.88_{\pm 2.75}$ & $74.11_{\pm 4.60}$ & $78.12_{\pm 2.59}$ & $79.92_{\pm 2.56}$ \\
 & CN-only FT (all) + LP & $63.78_{\pm 8.25}$ & $72.75_{\pm 2.23}$ & $75.15_{\pm 4.24}$ & $78.76_{\pm 2.80}$ & $80.04_{\pm 2.52}$ \\
\cline{2-7}
\addlinespace[1pt]
\multirow{3}{*}{BrainNorm\_M3} & No FT + LP & $60.02_{\pm 6.14}$ & $63.33_{\pm 6.43}$ & $63.04_{\pm 4.05}$ & $67.11_{\pm 2.37}$ & $69.83_{\pm 3.60}$ \\
 & CN-only FT(own) + LP & $59.40_{\pm 6.11}$ & $62.17_{\pm 8.15}$ & $61.34_{\pm 7.88}$ & $67.60_{\pm 2.32}$ & $67.22_{\pm 2.58}$ \\
 & CN-only FT (all) + LP & $58.54_{\pm 6.03}$ & $59.74_{\pm 6.16}$ & $60.22_{\pm 7.41}$ & $64.34_{\pm 6.19}$ & $64.31_{\pm 6.45}$ \\
\addlinespace[1pt]
\bottomrule
\end{tabular}
\end{table*}

\onecolumn

\begin{center}
\scriptsize
\setlength{\tabcolsep}{3pt}
\renewcommand{\arraystretch}{1.2}

\begin{longtable}{ccccc}
\caption{\normalsize Atlas parcel index mapping used in BrainNorm.\label{tab:atlas_parcels}} \\

\toprule
\textbf{Parcel ID} & \textbf{Atlas} & \textbf{Tissue\_Class} & \textbf{Label ID} & \textbf{Region} \\
\midrule
\endfirsthead

\toprule
\textbf{Parcel ID} & \textbf{Atlas} & \textbf{Tissue\_Class} & \textbf{Label ID} & \textbf{Region} \\
\midrule
\endhead

\midrule
\multicolumn{5}{r}{Continued on next page} \\
\midrule
\endfoot

\bottomrule
\endlastfoot

0 & AAL3 & Gray\_Matter & 1 & Left Precentral Gyrus \\
1 & AAL3 & Gray\_Matter & 2 & Right Precentral Gyrus \\
2 & AAL3 & Gray\_Matter & 3 & Left Superior Frontal Gyrus \\
3 & AAL3 & Gray\_Matter & 4 & Right Superior Frontal Gyrus \\
4 & AAL3 & Gray\_Matter & 5 & Left Middle Frontal Gyrus \\
5 & AAL3 & Gray\_Matter & 6 & Right Middle Frontal Gyrus \\
6 & AAL3 & Gray\_Matter & 7 & Left Inferior Frontal Gyrus Opercular Part \\
7 & AAL3 & Gray\_Matter & 8 & Right Inferior Frontal Gyrus Opercular Part \\
8 & AAL3 & Gray\_Matter & 9 & Left Inferior Frontal Gyrus Triangular Part \\
9 & AAL3 & Gray\_Matter & 10 & Right Inferior Frontal Gyrus Triangular Part \\
10 & AAL3 & Gray\_Matter & 11 & Left Inferior Frontal Gyrus Orbital Part \\
11 & AAL3 & Gray\_Matter & 12 & Right Inferior Frontal Gyrus Orbital Part \\
12 & AAL3 & Gray\_Matter & 13 & Left Rolandic Operculum \\
13 & AAL3 & Gray\_Matter & 14 & Right Rolandic Operculum \\
14 & AAL3 & Gray\_Matter & 15 & Left Supplementary Motor Area \\
15 & AAL3 & Gray\_Matter & 16 & Right Supplementary Motor Area \\
16 & AAL3 & Gray\_Matter & 17 & Left Olfactory Cortex \\
17 & AAL3 & Gray\_Matter & 18 & Right Olfactory Cortex \\
18 & AAL3 & Gray\_Matter & 19 & Left Superior Medial Frontal Gyrus \\
19 & AAL3 & Gray\_Matter & 20 & Right Superior Medial Frontal Gyrus \\
20 & AAL3 & Gray\_Matter & 21 & Left Medial Orbitofrontal Cortex \\
21 & AAL3 & Gray\_Matter & 22 & Right Medial Orbitofrontal Cortex \\
22 & AAL3 & Gray\_Matter & 23 & Left Gyrus Rectus \\
23 & AAL3 & Gray\_Matter & 24 & Right Gyrus Rectus \\
24 & AAL3 & Gray\_Matter & 25 & Left Orbitofrontal Cortex Medial Part \\
25 & AAL3 & Gray\_Matter & 26 & Right Orbitofrontal Cortex Medial Part \\
26 & AAL3 & Gray\_Matter & 27 & Left Orbitofrontal Cortex Anterior Part \\
27 & AAL3 & Gray\_Matter & 28 & Right Orbitofrontal Cortex Anterior Part \\
28 & AAL3 & Gray\_Matter & 29 & Left Orbitofrontal Cortex Posterior Part \\
29 & AAL3 & Gray\_Matter & 30 & Right Orbitofrontal Cortex Posterior Part \\
30 & AAL3 & Gray\_Matter & 31 & Left Orbitofrontal Cortex Lateral Part \\
31 & AAL3 & Gray\_Matter & 32 & Right Orbitofrontal Cortex Lateral Part \\
32 & AAL3 & Gray\_Matter & 33 & Left Insula \\
33 & AAL3 & Gray\_Matter & 34 & Right Insula \\
34 & AAL3 & Gray\_Matter & 37 & Left Middle Cingulate and Paracingulate Gyri \\
35 & AAL3 & Gray\_Matter & 38 & Right Middle Cingulate and Paracingulate Gyri \\
36 & AAL3 & Gray\_Matter & 39 & Left Posterior Cingulate Gyrus \\
37 & AAL3 & Gray\_Matter & 40 & Right Posterior Cingulate Gyrus \\
38 & AAL3 & Gray\_Matter & 41 & Left Hippocampus \\
39 & AAL3 & Gray\_Matter & 42 & Right Hippocampus \\
40 & AAL3 & Gray\_Matter & 43 & Left Parahippocampal Gyrus \\
41 & AAL3 & Gray\_Matter & 44 & Right Parahippocampal Gyrus \\
42 & AAL3 & Gray\_Matter & 45 & Left Amygdala \\
43 & AAL3 & Gray\_Matter & 46 & Right Amygdala \\
44 & AAL3 & Gray\_Matter & 47 & Left Calcarine Fissure and Cortex \\
45 & AAL3 & Gray\_Matter & 48 & Right Calcarine Fissure and Cortex \\
46 & AAL3 & Gray\_Matter & 49 & Left Cuneus \\
47 & AAL3 & Gray\_Matter & 50 & Right Cuneus \\
48 & AAL3 & Gray\_Matter & 51 & Left Lingual Gyrus \\
49 & AAL3 & Gray\_Matter & 52 & Right Lingual Gyrus \\
50 & AAL3 & Gray\_Matter & 53 & Left Superior Occipital Gyrus \\
51 & AAL3 & Gray\_Matter & 54 & Right Superior Occipital Gyrus \\
52 & AAL3 & Gray\_Matter & 55 & Left Middle Occipital Gyrus \\
53 & AAL3 & Gray\_Matter & 56 & Right Middle Occipital Gyrus \\
54 & AAL3 & Gray\_Matter & 57 & Left Inferior Occipital Gyrus \\
55 & AAL3 & Gray\_Matter & 58 & Right Inferior Occipital Gyrus \\
56 & AAL3 & Gray\_Matter & 59 & Left Fusiform Gyrus \\
57 & AAL3 & Gray\_Matter & 60 & Right Fusiform Gyrus \\
58 & AAL3 & Gray\_Matter & 61 & Left Postcentral Gyrus \\
59 & AAL3 & Gray\_Matter & 62 & Right Postcentral Gyrus \\
60 & AAL3 & Gray\_Matter & 63 & Left Superior Parietal Gyrus \\
61 & AAL3 & Gray\_Matter & 64 & Right Superior Parietal Gyrus \\
62 & AAL3 & Gray\_Matter & 65 & Left Inferior Parietal Gyrus \\
63 & AAL3 & Gray\_Matter & 66 & Right Inferior Parietal Gyrus \\
64 & AAL3 & Gray\_Matter & 67 & Left Supramarginal Gyrus \\
65 & AAL3 & Gray\_Matter & 68 & Right Supramarginal Gyrus \\
66 & AAL3 & Gray\_Matter & 69 & Left Angular Gyrus \\
67 & AAL3 & Gray\_Matter & 70 & Right Angular Gyrus \\
68 & AAL3 & Gray\_Matter & 71 & Left Precuneus \\
69 & AAL3 & Gray\_Matter & 72 & Right Precuneus \\
70 & AAL3 & Gray\_Matter & 73 & Left Paracentral Lobule \\
71 & AAL3 & Gray\_Matter & 74 & Right Paracentral Lobule \\
72 & AAL3 & Gray\_Matter & 75 & Left Caudate Nucleus \\
73 & AAL3 & Gray\_Matter & 76 & Right Caudate Nucleus \\
74 & AAL3 & Gray\_Matter & 77 & Left Putamen \\
75 & AAL3 & Gray\_Matter & 78 & Right Putamen \\
76 & AAL3 & Gray\_Matter & 79 & Left Pallidum \\
77 & AAL3 & Gray\_Matter & 80 & Right Pallidum \\
78 & AAL3 & Gray\_Matter & 83 & Left Heschl’s Gyrus \\
79 & AAL3 & Gray\_Matter & 84 & Right Heschl’s Gyrus \\
80 & AAL3 & Gray\_Matter & 85 & Left Superior Temporal Gyrus \\
81 & AAL3 & Gray\_Matter & 86 & Right Superior Temporal Gyrus \\
82 & AAL3 & Gray\_Matter & 87 & Left Temporal Pole: Superior Temporal Gyrus \\
83 & AAL3 & Gray\_Matter & 88 & Right Temporal Pole: Superior Temporal Gyrus \\
84 & AAL3 & Gray\_Matter & 89 & Left Middle Temporal Gyrus \\
85 & AAL3 & Gray\_Matter & 90 & Right Middle Temporal Gyrus \\
86 & AAL3 & Gray\_Matter & 91 & Left Temporal Pole: Middle Temporal Gyrus \\
87 & AAL3 & Gray\_Matter & 92 & Right Temporal Pole: Middle Temporal Gyrus \\
88 & AAL3 & Gray\_Matter & 93 & Left Inferior Temporal Gyrus \\
89 & AAL3 & Gray\_Matter & 94 & Right Inferior Temporal Gyrus \\
90 & AAL3 & Gray\_Matter & 95 & Left Crus I of Cerebellar Hemisphere \\
91 & AAL3 & Gray\_Matter & 96 & Right Crus I of Cerebellar Hemisphere \\
92 & AAL3 & Gray\_Matter & 97 & Left Crus II of Cerebellar Hemisphere \\
93 & AAL3 & Gray\_Matter & 98 & Right Crus II of Cerebellar Hemisphere \\
94 & AAL3 & Gray\_Matter & 99 & Left Lobule III of Cerebellar Hemisphere \\
95 & AAL3 & Gray\_Matter & 100 & Right Lobule III of Cerebellar Hemisphere \\
96 & AAL3 & Gray\_Matter & 101 & Left Lobule IV-V of Cerebellar Hemisphere \\
97 & AAL3 & Gray\_Matter & 102 & Right Lobule IV-V of Cerebellar Hemisphere \\
98 & AAL3 & Gray\_Matter & 103 & Left Lobule VI of Cerebellar Hemisphere \\
99 & AAL3 & Gray\_Matter & 104 & Right Lobule VI of Cerebellar Hemisphere \\
100 & AAL3 & Gray\_Matter & 105 & Left Lobule VIIB of Cerebellar Hemisphere \\
101 & AAL3 & Gray\_Matter & 106 & Right Lobule VIIB of Cerebellar Hemisphere \\
102 & AAL3 & Gray\_Matter & 107 & Left Lobule VIII of Cerebellar Hemisphere \\
103 & AAL3 & Gray\_Matter & 108 & Right Lobule VIII of Cerebellar Hemisphere \\
104 & AAL3 & Gray\_Matter & 109 & Left Lobule IX of Cerebellar Hemisphere \\
105 & AAL3 & Gray\_Matter & 110 & Right Lobule IX of Cerebellar Hemisphere \\
106 & AAL3 & Gray\_Matter & 111 & Left Lobule X of Cerebellar Hemisphere \\
107 & AAL3 & Gray\_Matter & 112 & Right Lobule X of Cerebellar Hemisphere \\
108 & AAL3 & Gray\_Matter & 113 & Lobule I-II of Vermis \\
109 & AAL3 & Gray\_Matter & 114 & Lobule III of Vermis \\
110 & AAL3 & Gray\_Matter & 115 & Lobule IV-V of Vermis \\
111 & AAL3 & Gray\_Matter & 116 & Lobule VI of Vermis \\
112 & AAL3 & Gray\_Matter & 117 & Lobule VII of Vermis \\
113 & AAL3 & Gray\_Matter & 118 & Lobule VIII of Vermis \\
114 & AAL3 & Gray\_Matter & 119 & Lobule IX of Vermis \\
115 & AAL3 & Gray\_Matter & 120 & Lobule X of Vermis \\
116 & AAL3 & Gray\_Matter & 121 & Left Anteroventral Thalamic Nucleus \\
117 & AAL3 & Gray\_Matter & 122 & Right Anteroventral Thalamic Nucleus \\
118 & AAL3 & Gray\_Matter & 123 & Left Lateral Posterior Thalamic Nucleus \\
119 & AAL3 & Gray\_Matter & 124 & Right Lateral Posterior Thalamic Nucleus \\
120 & AAL3 & Gray\_Matter & 125 & Left Ventral Anterior Thalamic Nucleus \\
121 & AAL3 & Gray\_Matter & 126 & Right Ventral Anterior Thalamic Nucleus \\
122 & AAL3 & Gray\_Matter & 127 & Left Ventral Lateral Thalamic Nucleus \\
123 & AAL3 & Gray\_Matter & 128 & Right Ventral Lateral Thalamic Nucleus \\
124 & AAL3 & Gray\_Matter & 129 & Left Ventral Posterolateral Thalamic Nucleus \\
125 & AAL3 & Gray\_Matter & 130 & Right Ventral Posterolateral Thalamic Nucleus \\
126 & AAL3 & Gray\_Matter & 131 & Left Intralaminar Thalamic Nuclei \\
127 & AAL3 & Gray\_Matter & 132 & Right Intralaminar Thalamic Nuclei \\
128 & AAL3 & Gray\_Matter & 133 & Left Reuniens Thalamic Nucleus \\
129 & AAL3 & Gray\_Matter & 134 & Right Reuniens Thalamic Nucleus \\
130 & AAL3 & Gray\_Matter & 135 & Left Mediodorsal Medial Magnocellular Thalamic Nucleus \\
131 & AAL3 & Gray\_Matter & 136 & Right Mediodorsal Medial Magnocellular Thalamic Nucleus \\
132 & AAL3 & Gray\_Matter & 137 & Left Mediodorsal Lateral Parvocellular Thalamic Nucleus \\
133 & AAL3 & Gray\_Matter & 138 & Right Mediodorsal Lateral Parvocellular Thalamic Nucleus \\
134 & AAL3 & Gray\_Matter & 139 & Left Lateral Geniculate Thalamic Nucleus \\
135 & AAL3 & Gray\_Matter & 140 & Right Lateral Geniculate Thalamic Nucleus \\
136 & AAL3 & Gray\_Matter & 141 & Left Medial Geniculate Thalamic Nucleus \\
137 & AAL3 & Gray\_Matter & 142 & Right Medial Geniculate Thalamic Nucleus \\
138 & AAL3 & Gray\_Matter & 143 & Left Pulvinar Inferior Thalamic Nucleus \\
139 & AAL3 & Gray\_Matter & 144 & Right Pulvinar Inferior Thalamic Nucleus \\
140 & AAL3 & Gray\_Matter & 145 & Left Pulvinar Medial Thalamic Nucleus \\
141 & AAL3 & Gray\_Matter & 146 & Right Pulvinar Medial Thalamic Nucleus \\
142 & AAL3 & Gray\_Matter & 147 & Left Pulvinar Anterior Thalamic Nucleus \\
143 & AAL3 & Gray\_Matter & 148 & Right Pulvinar Anterior Thalamic Nucleus \\
144 & AAL3 & Gray\_Matter & 149 & Left Pulvinar Lateral Thalamic Nucleus \\
145 & AAL3 & Gray\_Matter & 150 & Right Pulvinar Lateral Thalamic Nucleus \\
146 & AAL3 & Gray\_Matter & 151 & Left Anterior Cingulate Cortex Subgenual Part \\
147 & AAL3 & Gray\_Matter & 152 & Right Anterior Cingulate Cortex Subgenual Part \\
148 & AAL3 & Gray\_Matter & 153 & Left Anterior Cingulate Cortex Pregenual Part \\
149 & AAL3 & Gray\_Matter & 154 & Right Anterior Cingulate Cortex Pregenual Part \\
150 & AAL3 & Gray\_Matter & 155 & Left Anterior Cingulate Cortex Supracallosal Part \\
151 & AAL3 & Gray\_Matter & 156 & Right Anterior Cingulate Cortex Supracallosal Part \\
152 & AAL3 & Gray\_Matter & 157 & Left Nucleus Accumbens \\
153 & AAL3 & Gray\_Matter & 158 & Right Nucleus Accumbens \\
154 & AAL3 & Gray\_Matter & 159 & Left Ventral Tegmental Area \\
155 & AAL3 & Gray\_Matter & 160 & Right Ventral Tegmental Area \\
156 & AAL3 & Gray\_Matter & 161 & Left Substantia Nigra Pars Compacta \\
157 & AAL3 & Gray\_Matter & 162 & Right Substantia Nigra Pars Compacta \\
158 & AAL3 & Gray\_Matter & 163 & Left Substantia Nigra Pars Reticulata \\
159 & AAL3 & Gray\_Matter & 164 & Right Substantia Nigra Pars Reticulata \\
160 & AAL3 & Gray\_Matter & 165 & Left Red Nucleus \\
161 & AAL3 & Gray\_Matter & 166 & Right Red Nucleus \\
162 & AAL3 & Gray\_Matter & 167 & Left Locus Coeruleus \\
163 & AAL3 & Gray\_Matter & 168 & Right Locus Coeruleus \\
164 & AAL3 & Gray\_Matter & 169 & Dorsal Raphe Nucleus \\
165 & AAL3 & Gray\_Matter & 170 & Median Raphe Nucleus \\
166 & JHU & White\_Matter & 1 & Middle Cerebellar Peduncle \\
167 & JHU & White\_Matter & 2 & Pontine Crossing Tract \\
168 & JHU & White\_Matter & 3 & Genu Of Corpus Callosum \\
169 & JHU & White\_Matter & 4 & Body Of Corpus Callosum \\
170 & JHU & White\_Matter & 5 & Splenium Of Corpus Callosum \\
171 & JHU & White\_Matter & 6 & Fornix \\
172 & JHU & White\_Matter & 7 & Right Corticospinal Tract \\
173 & JHU & White\_Matter & 8 & Left Corticospinal Tract \\
174 & JHU & White\_Matter & 9 & Right Medial Lemniscus \\
175 & JHU & White\_Matter & 10 & Left Medial Lemniscus \\
176 & JHU & White\_Matter & 11 & Right Inferior Cerebellar Peduncle \\
177 & JHU & White\_Matter & 12 & Left Inferior Cerebellar Peduncle \\
178 & JHU & White\_Matter & 13 & Right Superior Cerebellar Peduncle \\
179 & JHU & White\_Matter & 14 & Left Superior Cerebellar Peduncle \\
180 & JHU & White\_Matter & 15 & Right Cerebral Peduncle \\
181 & JHU & White\_Matter & 16 & Left Cerebral Peduncle \\
182 & JHU & White\_Matter & 17 & Right Anterior Limb Of Internal Capsule \\
183 & JHU & White\_Matter & 18 & Left Anterior Limb Of Internal Capsule \\
184 & JHU & White\_Matter & 19 & Right Posterior Limb Of Internal Capsule \\
185 & JHU & White\_Matter & 20 & Left Posterior Limb Of Internal Capsule \\
186 & JHU & White\_Matter & 21 & Right Retrolenticular Part Of Internal Capsule \\
187 & JHU & White\_Matter & 22 & Left Retrolenticular Part Of Internal Capsule \\
188 & JHU & White\_Matter & 23 & Right Anterior Corona Radiata \\
189 & JHU & White\_Matter & 24 & Left Anterior Corona Radiata \\
190 & JHU & White\_Matter & 25 & Right Superior Corona Radiata \\
191 & JHU & White\_Matter & 26 & Left Superior Corona Radiata \\
192 & JHU & White\_Matter & 27 & Right Posterior Corona Radiata \\
193 & JHU & White\_Matter & 28 & Left Posterior Corona Radiata \\
194 & JHU & White\_Matter & 29 & Right Posterior Thalamic Radiation \\
195 & JHU & White\_Matter & 30 & Left Posterior Thalamic Radiation \\
196 & JHU & White\_Matter & 31 & Right Sagittal Stratum \\
197 & JHU & White\_Matter & 32 & Left Sagittal Stratum \\
198 & JHU & White\_Matter & 33 & Right External Capsule \\
199 & JHU & White\_Matter & 34 & Left External Capsule \\
200 & JHU & White\_Matter & 35 & Right Cingulum (cingulate Gyrus) \\
201 & JHU & White\_Matter & 36 & Left Cingulum (cingulate Gyrus) \\
202 & JHU & White\_Matter & 37 & Right Cingulum (hippocampus) \\
203 & JHU & White\_Matter & 38 & Left Cingulum (hippocampus) \\
204 & JHU & White\_Matter & 39 & Right Fornix / Stria Terminalis \\
205 & JHU & White\_Matter & 40 & Left Fornix / Stria Terminalis \\
206 & JHU & White\_Matter & 41 & Right Superior Longitudinal Fasciculus \\
207 & JHU & White\_Matter & 42 & Left Superior Longitudinal Fasciculus \\
208 & JHU & White\_Matter & 43 & Right Superior Fronto-occipital Fasciculus \\
209 & JHU & White\_Matter & 44 & Left Superior Fronto-occipital Fasciculus \\
210 & JHU & White\_Matter & 45 & Right Inferior Fronto-occipital Fasciculus \\
211 & JHU & White\_Matter & 46 & Left Inferior Fronto-occipital Fasciculus \\
212 & JHU & White\_Matter & 47 & Right Uncinate Fasciculus \\
213 & JHU & White\_Matter & 48 & Left Uncinate Fasciculus \\
214 & JHU & White\_Matter & 49 & Right Tapetum \\
215 & JHU & White\_Matter & 50 & Left Tapetum \\
216 & HO & Cerebrospinal\_Fluid & 2 & Left Lateral Ventricle \\
217 & HO & Cerebrospinal\_Fluid & 13 & Right Lateral Ventricle \\

\end{longtable}
\end{center}





\end{document}